\documentclass[10pt,twocolumn,letterpaper]{article}

\newif\ifarxiv
\arxivtrue
\ifarxiv
    \usepackage[pagenumbers]{cvpr}
\else
    \usepackage[review]{cvpr}
\fi

\usepackage{amsmath,amsfonts,bm}

\def\1{\bm{1}}

\DeclareMathAlphabet{\mathsfit}{\encodingdefault}{\sfdefault}{m}{sl}
\SetMathAlphabet{\mathsfit}{bold}{\encodingdefault}{\sfdefault}{bx}{n}

\usepackage{float}
\usepackage{booktabs}
\usepackage{graphicx}
\usepackage{pifont}
\usepackage{multirow}
\usepackage{multicol}
\usepackage{makecell}
\usepackage{adjustbox}
\usepackage{tabularx}
\usepackage{ragged2e}
\usepackage{array}

\usepackage{microtype}
\usepackage{amsmath,amssymb,amsfonts,mathtools}
\usepackage{xcolor}
\usepackage{url}
\usepackage{xurl}
\usepackage{enumitem}
\usepackage{tikz}
\usetikzlibrary{arrows.meta,positioning,fit,calc,shapes.geometric}
\usepackage{subcaption}
\usepackage{wrapfig}
\usepackage{needspace}
\usepackage{placeins}
\usepackage{algorithm}
\usepackage[noend]{algpseudocode}

\newcommand{\rev}[1]{#1}
\newif\ifrevisionhighlights
\revisionhighlightsfalse
\DeclareRobustCommand{\revb}[1]{%
  \ifrevisionhighlights{\color{red}#1}\else#1\fi
}

\definecolor{draftred}{RGB}{160,20,20}
\definecolor{methodblue}{RGB}{35,83,135}
\definecolor{softgray}{RGB}{245,245,245}

\newcommand{\MethodName}{PhyScale}

\newcommand{\revcitep}[1]{\citep{#1}}

\newcommand{\arxivwording}[2]{\ifarxiv#1\else#2\fi}

\definecolor{cvprblue}{rgb}{0.21,0.49,0.74}
\usepackage[pagebackref,breaklinks,colorlinks,allcolors=cvprblue]{hyperref}

\def\confName{CVPR}
\def\confYear{2027}

\title{\MethodName: Physically Grounded Object Size Correction \\in Text-to-Image Generation}

\ifarxiv
    \author{Lingxiao Li\textsuperscript{1}\qquad
            Ledell Wu\textsuperscript{2}\qquad
            Boqing Gong\textsuperscript{1}\\[2pt]
            {\normalsize \textsuperscript{1}Boston University\qquad
             \textsuperscript{2}Creatify AI}\\[2pt]
            {\small\url{https://lingxiao-li.github.io/physcale/}}}
    \hypersetup{
        pdftitle={PhyScale: Physically Grounded Object Size Correction in Text-to-Image Generation},
        pdfauthor={Lingxiao Li, Ledell Wu, Boqing Gong}
    }
\fi

\begin{document}

\twocolumn[{%
\renewcommand\twocolumn[1][]{#1}%
\maketitle
\begin{center}
    \centering
    \captionsetup{type=figure}
    \includegraphics[width=\textwidth]{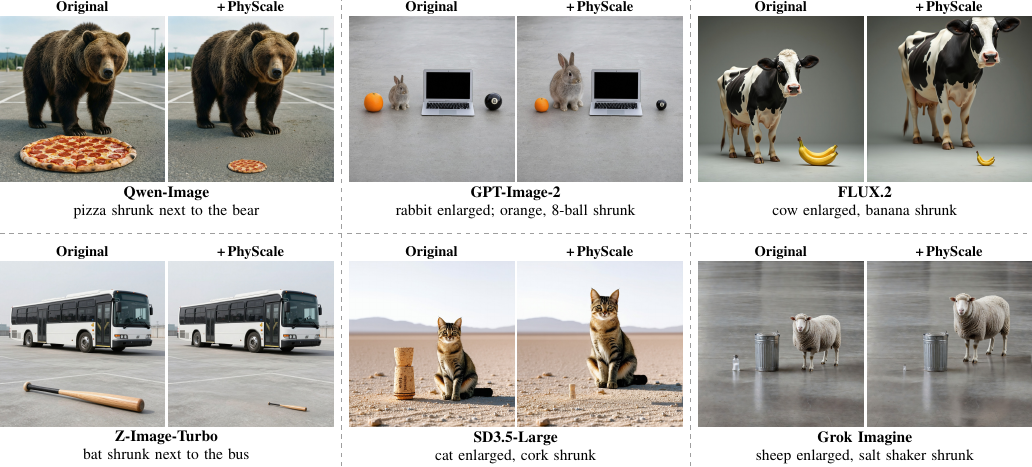}
    \captionof{figure}{
    \rev{\textbf{Relative-scale failures from modern image generators and corrections by \MethodName{}.} Each pair shows the original output of the named generator (left) and our result (right) with identical framing; \MethodName{} resizes only the objects listed below each pair, preserving object identity and the rest of the scene.}
    }
    \label{fig:teaser}
\end{center}%
}]

\begin{abstract}
Generated images can appear photorealistic while depicting objects at implausible relative physical sizes. Correcting these errors requires distinguishing physical size from its depth- and pose-dependent appearance. We introduce \MethodName{}, a framework that formulates size correction as geometric inference and planning, separating physical-size changes from their image-space realization. Using typical real-world object dimensions from a knowledge base and estimated scene geometry, it computes size, depth, and placement commands for an image editor controlled through spatial rotary positional embeddings. To learn the associated appearance changes, we adapt the editor on only 2,468 paired renders from controlled 3D object transformations. \rev{The resulting editor outperforms prior geometry-aware editors in target fidelity and identity preservation on real-image edits. Combined with our geometric inference and planning front end, it matches a state-of-the-art VLM-based correction pipeline on GenScale at one-twentieth of the inference cost, and human raters prefer its corrections.}
\end{abstract}
\section{Introduction}
\begin{figure*}[t]
    \centering
    \includegraphics[width=0.95\textwidth]{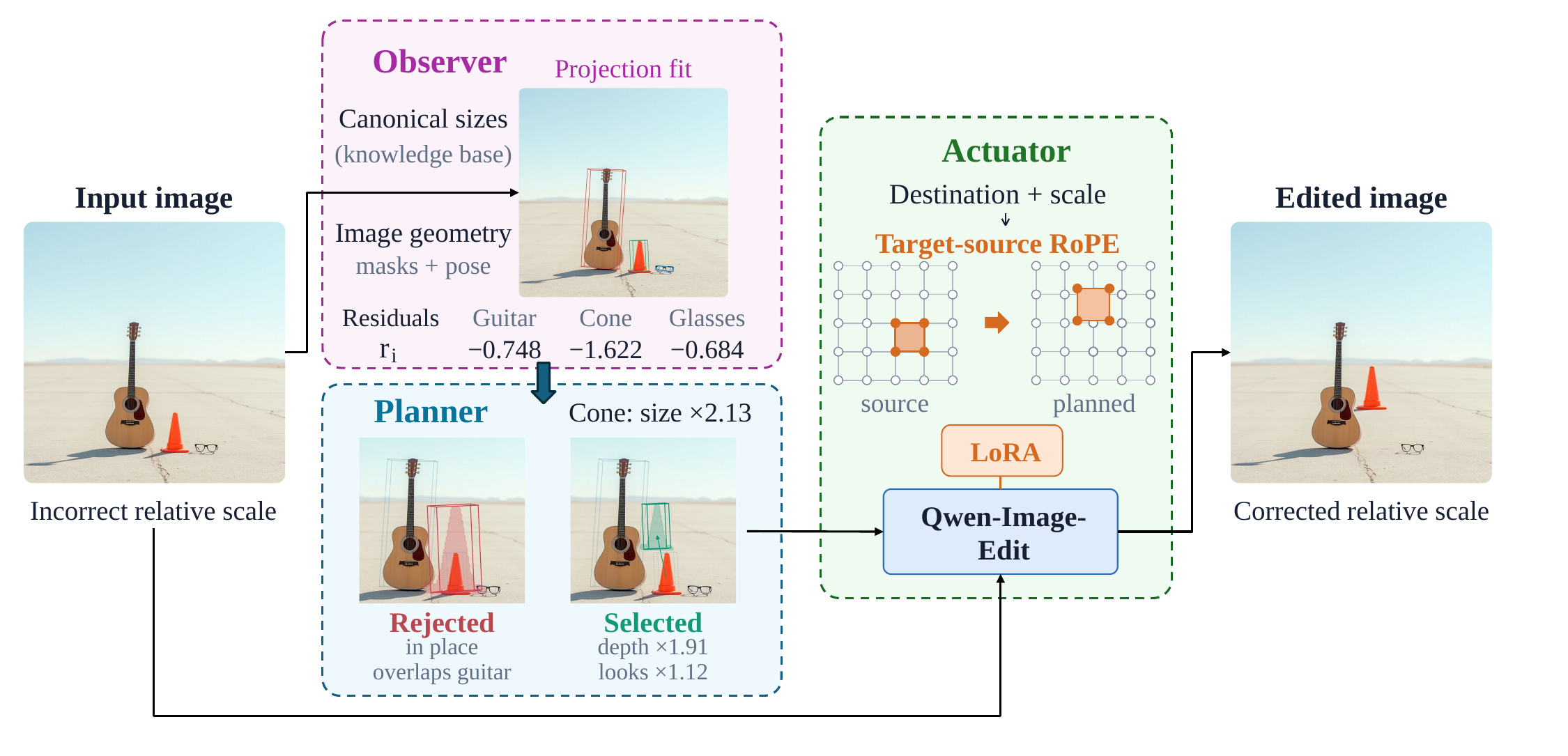}
    \vspace{-2ex}
    \caption{\textbf{Overview of \MethodName{}.} The observer combines instance masks, estimated scene geometry and pose, and knowledge-base dimensions to estimate physical-size residuals. The planner allocates size corrections and searches over depth and placement under scene constraints and the editor's operating limits. The actuator executes the resulting plan by modifying the edited object's source-token RoPE coordinates in an adapted Qwen-Image-Edit model.}
    \label{fig:pipeline}
    \vspace{-2ex}
\end{figure*}

Modern text-to-image generators can synthesize photorealistic scenes~\citep{rombach2022highresolution,esser2024scaling}, yet render familiar objects at implausible relative physical sizes (Fig.~\ref{fig:teaser}). This failure is systematic rather than anecdotal: GenScale~\citep{li2026genscale} documents a mean-regression pattern in which physically small objects are rendered too large and large objects too small by state-of-the-art image generators~\citep{openai2026gptimage2,google2026nanobanana2,xai2026grokimagine,wu2025qwenimage,blackforestlabs2025flux2}. We aim to correct these disproportions in existing images while preserving object identities and scene context.

Correcting the object sizes in generated images, however, requires distinguishing physical size from apparent size. Under perspective, a small nearby object and a larger distant object can have the same projected extent~\citep{hartley2004multiple}, while pose further affects it through foreshortening. Size errors must therefore be estimated in the context of scene geometry. \rev{Applying a correction also requires a feasible placement: enlarging an object in place can cause collisions or move it outside the frame, while moving it farther away reduces its projected footprint.} The physical change and its feasible image-space realization are thus distinct but coupled decisions.

Existing spatial controllers~\citep{zhang2023controlnet,li2023gligen} and geometry-aware editors, including ObjectMover~\citep{yu2025objectmover}, PhyEdit~\citep{xu2026phyedit}, and RoPEMover~\citep{oztas2026ropemover}, provide mechanisms for prescribed spatial edits, but rely on user-specified layouts, regions, or transformations rather than inferring the required physical correction. ReScale~\citep{li2026genscale} addresses this decision stage with a vision-language model (VLM) that diagnoses size errors and plans local resizing while considering depth and scene constraints. \rev{Geometric computation offers inspectable decisions and substantially faster, cheaper inference than VLM-based diagnosis and planning.} This raises two questions: \textbf{(1) Can physical-size corrections be computed explicitly from estimated scene geometry? (2) Can an image editor faithfully execute the resulting correction commands?}

\rev{We introduce \MethodName{}}, a framework connecting size measurement, planning, and editing through explicit geometric commands (Fig.~\ref{fig:pipeline}). Given \rev{typical real-world} object dimensions and estimated scene geometry, \rev{a pose-aware observer estimates size errors by fitting scaled object geometry to observed projections. A geometric planner allocates corrections across objects and searches depth and lateral placement} under scene constraints and the editor's calibrated operating limits. \rev{This process requires} no user-provided layouts, boxes, or resize factors.

To execute the resulting commands, we encode the planned apparent size and destination by modifying spatial rotary positional embeddings (RoPE) in Qwen-Image-Edit~\citep{wu2025qwenimage}\revb{, following RoPEMover~\citep{oztas2026ropemover}}. This control specifies the projection, but the editor must also learn the \rev{associated changes in visibility, contact, and shadows. We adapt it} with LoRA~\citep{hu2022lora} on only 2,468 paired renders from controlled 3D object transformations using Kubric~\citep{greff2022kubric}. Only the image editor is adapted, without real-image editing pairs; the geometric estimation and planning stages require no task-specific training.

\rev{\textbf{Our contributions are three-fold:}}
\begin{itemize}[leftmargin=1.5em, labelsep=0.5em, itemsep=4pt, topsep=2pt, parsep=0pt, partopsep=0pt]
\item \rev{We formulate \textbf{relative physical-size correction} as geometric inference and planning that distributes size changes across objects and selects depth and placement under scene and editing constraints.}

\item \revb{We introduce \textbf{full-scene 3D supervision for RoPE-controlled editing}: from only 2,468 rendered pairs, the editor learns size-depth edits and their associated changes in visibility, contact, and shadows.}

\item \rev{We show that the editor \textbf{outperforms state-of-the-art geometry-aware editors} on real-image edits, and that the complete system matches ReScale on GenScale~\citep{li2026genscale} at one-twentieth of its inference cost with higher human preference.}
\end{itemize}

\section{Related Work}
\paragraph{Geometry-controlled image generation and editing.}
\arxivwording{Image generation can be controlled through semantic attributes~\citep{li2023hae,li2025hypdae}, 2D maps}{Spatial controllers guide synthesis with 2D maps} or regions~\citep{zhang2023controlnet,li2023gligen,zhang2025eligen} or rendered 3D layouts~\citep{agrawal2026seethrough3d}, while language-mediated systems infer layouts or refine outputs~\arxivwording{\citep{lian2024llmgrounded,Li24bifrost,wu2024selfcorrecting}}{\citep{lian2024llmgrounded,wu2024selfcorrecting}}. Image editors use lifted features or reconstructed geometry~\citep{pandey2024diffusionhandles,yenphraphai2024imagesculpting}, video priors~\citep{yu2025objectmover,wu2026chronoedit}, visual 3D controls~\citep{wang2026boxctrl,xu2026phyedit}, or positional encodings~\citep{bai2026pefield,sella2026looserope}. \arxivwording{Training data include synthetic images of object states~\citep{chen2025states} as well as rendered}{To learn scene effects, prior work uses rendered}~\citep{michel2023object3dit,zhang2026geoedit} or captured counterfactual pairs~\citep{winter2024objectdrop}. \rev{\revb{RoPEMover~\citep{oztas2026ropemover} relocates and rescales objects by warping RoPE coordinates, and GIVE~\citep{chang2026give} uses depth and orientation boxes. Our editor adopts the RoPE warp and learns planned size-depth transformations and their appearance changes from rendered pairs alone.}}

\paragraph{Physical size reasoning and correction.}
Choosing these edits for size correction requires distinguishing physical dimensions from apparent extent. Object-size knowledge supports commonsense reasoning~\citep{bagherinezhad2016elephants} and monocular depth estimation~\citep{wu2018sizetodepth}\rev{, and single-view metrology~\revcitep{criminisi2000single,hoiem2008putting} relates image measurements to physical size}. In generation, ASQL regularizes summed entity attention using LLM-derived size orderings~\citep{jayasekara2026allinone}. For existing images, ReScale~\citep{li2026genscale} uses a multimodal agent to diagnose size errors and plan contact-preserving resizing while considering depth and scene constraints. \MethodName{} instead estimates size deviations through pose-aware projection fitting and plans size changes, depth, and placement under scene and actuator constraints. This provides an inspectable geometric alternative to VLM decisions while keeping the actuator interchangeable.

\section{Method}
\label{sec:method}
\begin{figure*}[t]
    \centering
    \includegraphics[width=1.0\textwidth]{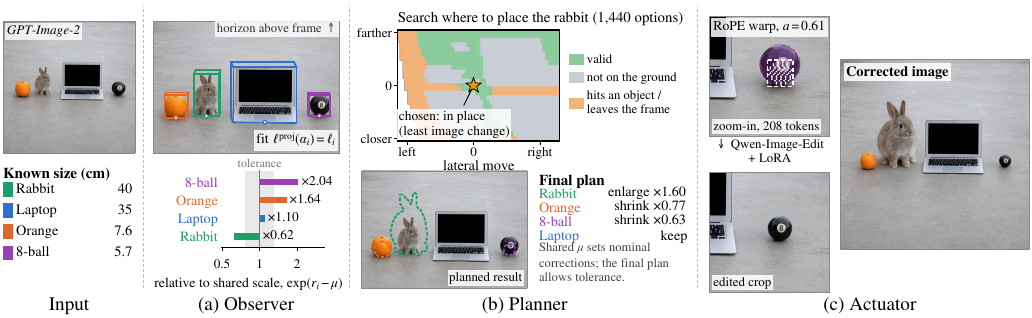}
    \caption{\textbf{Running example for the three method stages.} (a) The observer fits each object's knowledge-base cuboid to its image projection and estimates its size residual. (b) The planner allocates size corrections and searches depth-lateral placements under prioritized constraints; the colored map shows candidate placements for the \rev{rabbit}. (c) The actuator converts the selected projection into a RoPE warp and edits an enlarged local crop when the target is small.}
    \label{fig:method_example}
    \vspace{-2ex}
\end{figure*}

\paragraph{Approach overview.}
\MethodName{} separates geometric correction decisions from learned image editing through the three-stage pipeline in Fig.~\ref{fig:pipeline}. We first formulate relative physical-size correction and define size-depth edit commands (Sec.~\ref{sec:problem_formulation}). The \textbf{observer} estimates size errors by fitting knowledge-base object dimensions to image measurements under estimated scene geometry and pose (Sec.~\ref{sec:observer}). The \textbf{planner} uses these estimates to choose size corrections, depth, and lateral placement under prioritized scene and editor constraints (Sec.~\ref{sec:planner}). The \textbf{actuator} executes the resulting image-space transformation through spatial rotary positional embeddings (Sec.~\ref{sec:actuator}). The geometric decisions are explicitly computed and inspectable; only the actuator receives task-specific training.

\subsection{Problem Formulation \& Preliminaries}
\label{sec:problem_formulation}
\paragraph{Relative physical size.}
Given a generated image $I$, target and reference object names $T$ and $R$, and a knowledge base of typical real-world object dimensions, we seek a corrected image $I'$ with plausible relative physical sizes while preserving object identities and scene context. The system binds the names to image instances and infers geometric changes without user-provided destination boxes or resize factors. \rev{Following GenScale~\citep{li2026genscale}, whose object categories were selected for stable real-world dimensions, the knowledge base supplies a default size per category; a user-specified size can replace it without changing the algorithm.}

Let $S_i$ denote a representative linear physical dimension of object $i$ (e.g., its height), $d_i$ its camera-frame depth, and $\ell_i$ its projected linear extent in pixels. Under local weak perspective~\citep{hartley2004multiple},
\begin{equation}
\ell_i \approx \kappa_i\frac{fS_i}{d_i},
\label{eq:projection_model}
\end{equation}
where $f$ is the focal length in pixels and $\kappa_i$ captures shape- and pose-dependent foreshortening. For two objects $T$ and $R$ in the same image, the focal length cancels:
\begin{equation}
\log\frac{\ell_T}{\ell_R}
\approx
\log\frac{S_T}{S_R}
-
\log\frac{d_T}{d_R}
+
\log\frac{\kappa_T}{\kappa_R}.
\label{eq:projected_relative_scale}
\end{equation}
Projected size ratios therefore confound physical size with depth, shape, and pose. We use these relations to parameterize image edits.

\paragraph{Size-depth edits.}
For intended size $S_i'$ and depth $d_i'$, we define the size-depth command
\begin{equation}
\mathbf{c}_i=(c_{s,i},c_{d,i}),
\qquad
c_{s,i}=\log\frac{S_i'}{S_i},
\qquad
c_{d,i}=\log\frac{d_i'}{d_i}.
\label{eq:action}
\end{equation}
Positive $c_{s,i}$ enlarges the object; positive $c_{d,i}$ moves it farther away. For a fixed camera and an identity- and pose-preserving edit with approximately unchanged foreshortening, Eq.~\ref{eq:projection_model} gives the predicted apparent scaling factor
\begin{equation}
a_i=\exp(c_{s,i}-c_{d,i})\approx\frac{\ell_i'}{\ell_i}.
\label{eq:apparent_action}
\end{equation}
Thus $(c_{s,i},0)$ changes size at fixed depth, whereas $(0,c_{d,i})$ changes depth at fixed size. Equal commands $c_{s,i}=c_{d,i}$ give $a_i=1$, approximately preserving projected extent while allowing changes in position, occlusion, and contact. This freedom allows a physical-size correction without forcing an in-place resize\rev{: in Fig.~\ref{fig:pipeline}, the cone is enlarged $2.13\times$ and moved $1.91\times$ farther, so it appears only $1.12\times$ larger}. Appendix~A gives the full derivation.

\subsection{Pose-Aware Physical Observer}
\label{sec:observer}
\paragraph{Physical size from inverse projection.}
The observer estimates size by fitting a uniform scale to a cuboid, rather than reconstructing detailed 3D shape. For instance $i$, let $M_i$ be its mask, $\mathbf{D}_i=(L_i,W_i,H_i)$ its knowledge-base length, width, and height, and $Q_i$ the eight corners of the corresponding ground-anchored cuboid. Flat and elongated objects use orientation $R_i$ from a frozen pose estimator~\citep{you2026unipose9d}; compact objects use an upright orientation. \rev{The pose estimator supplies orientation only: we discard its predicted metric size and translation and use its axis-length ratios solely to match predicted axes to $(L_i,W_i,H_i)$.}

For ground-supported objects, we approximate the contact point by the horizontal center of the mask's bounding box at its lowest occupied row. We use fixed camera intrinsics and camera height, with tilt set by the estimated horizon; the horizon is fit from MoGe-2 depth predictions on segmented ground pixels~\citep{wang2025moge2}. Intersecting the camera ray through the contact point with the ground plane gives position $g_i$. Appendix~D.1 gives the camera parameters, horizon fitting, and validity checks.

Let $\Pi$ be the camera projector and $\operatorname{extent}(\cdot)$ return the longest side of the axis-aligned 2D bounding box enclosing a mask or projected points. For a trial scale $\alpha>0$ applied uniformly to the cuboid, we measure
\begin{equation}
\begin{aligned}
\ell_i &= \operatorname{extent}(M_i),\\
\ell_i^{\mathrm{proj}}(\alpha) &= \operatorname{extent}\!\bigl(\Pi(g_i+\alpha R_iQ_i)\bigr).
\end{aligned}
\label{eq:observer_extent_proj}
\end{equation}
With the camera, orientation, and ground position held fixed during fitting, the only unknown is $\alpha$. We obtain $\alpha_i$ from
\begin{equation}
\ell_i^{\mathrm{proj}}(\alpha_i)=\ell_i,
\label{eq:observer_alpha}
\end{equation}
using bisection over a scale interval where projected extent increases with $\alpha$ and brackets the observed extent. If no such interval exists, the observer abstains.

Let $S_i^{\mathrm{kb}}$ be the knowledge-base value of the representative dimension $S_i$ defined in Sec.~\ref{sec:problem_formulation} (e.g., $H_i$ when $S_i$ is height). Uniform scaling gives $\widehat S_i=\alpha_iS_i^{\mathrm{kb}}$, so the log-size residual is
\begin{equation}
r_i=\log\frac{\widehat S_i}{S_i^{\mathrm{kb}}}=\log\alpha_i.
\label{eq:observer_residual}
\end{equation}
Here $r_i=0$ agrees with the knowledge-base size; positive and negative values indicate larger and smaller estimates. Relative size is correct when objects have the same residual. In Fig.~\ref{fig:method_example}(a), the fitted scales show the \rev{8-ball and orange} as relatively large, the \rev{rabbit} as relatively small, and the \rev{laptop} close to the shared scale used by the planner.

\begin{figure*}[t]
    \centering
    \includegraphics[width=1.0\textwidth]{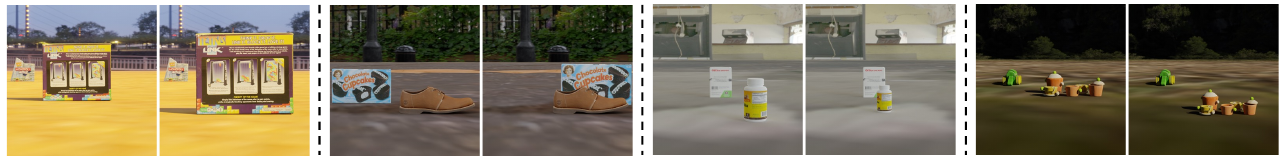}
    \caption{\textbf{Geometrically validated training pairs from 3D scenes.} Each pair shows a source image (left) and the destination re-rendered after the prescribed object transformation (right), with the camera, reference object, and environment fixed.}
    \label{fig:counterfactual_pairs}
\end{figure*}

\begin{figure*}[t]
\centering
\includegraphics[width=1.0\linewidth]{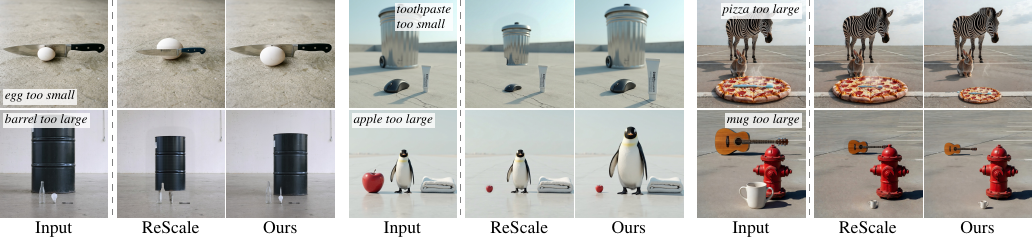}
\vspace{-2ex}
\caption{\textbf{End-to-end size correction.} Input / ReScale / \MethodName{} triplets; annotations mark size errors.}
\label{fig:exp_genscale_qual}
\vspace{-2ex}
\end{figure*}

\subsection{Size and Placement Planner}
\label{sec:planner}
The observer estimates relative-size errors; the planner decides how to distribute the correction and where edited objects should be placed.

\paragraph{Allocating size corrections.}
After resizing object $i$ by $c_{s,i}$, its residual becomes $r_i+c_{s,i}$. We parameterize the correction by a shared post-edit residual $\mu$\rev{, the size offset common to all objects after correction}:
\begin{equation}
c_{s,i}=\mu-r_i.
\label{eq:gauge_command}
\end{equation}
Thus the nominal residuals agree, recovering the knowledge-base size ratios. Setting $\mu=r_R$ keeps reference $R$ unchanged; varying $\mu$ distributes the correction across objects. A final tolerance refinement may reduce these commands. In Fig.~\ref{fig:method_example}(b), the resulting plan keeps the \rev{laptop} unchanged, shrinks the \rev{orange and 8-ball}, and enlarges the \rev{rabbit}.

\paragraph{Searching for a valid placement.}
For each $\mu$, Eq.~\ref{eq:gauge_command} fixes $c_{s,i}$; we then search each edited object's depth command $c_{d,i}$ and horizontal displacement $\Delta x_i$, while vertical placement follows the ground plane. The projector predicts the destination footprint and apparent scale $a_i=\exp(c_{s,i}-c_{d,i})$.

Candidates are compared by prioritized constraints: support and collision first; then framing, overlap/depth ordering, and \rev{target resolution in the editor's token grid (Sec.~\ref{sec:actuator})}; then the actuator's calibrated scaling range. Ties are broken by image change, motion, and the number of edited objects, so lower-priority improvements cannot override higher-priority violations. \rev{These priorities are explicit rules and can be reordered without retraining: ranking spatial-relation preservation directly after physical validity keeps 99.8\% of left-right and 99.7\% of front-back object orders (Appendix~G.7).}

We search discrete $\mu$ values on a coarse depth-lateral grid and refine the selected solution on a finer grid; multi-object placement proceeds greedily from larger to smaller source areas. Figure~\ref{fig:method_example}(b) visualizes the \rev{rabbit search: invalid candidates violate support or collision/frame constraints,} while the selected in-place candidate yields the least image change among valid choices. A final refinement removes unnecessary resizing within the relative-size tolerance. Appendix~D.2 gives the grids, tolerances, tie-breaking, and fallback rules.

\begin{table*}[t]
\centering
\captionsetup{skip=2pt,singlelinecheck=false}
\caption{\rev{\textbf{GenScale \revb{Task~1} correction.} Mean relation error $|s-3|$ ($\downarrow$) on 2,800 images (8,001 relations); $s$ is the five-vote Gemini score (3: plausible size), aggregated by unique mode or rounded median. Unjudgeable relations receive error 2 \revb{(omitted in~\citep{li2026genscale})}.}}
\label{tab:exp_genscale}
\small
\renewcommand{\arraystretch}{0.96}
\setlength{\tabcolsep}{2.5pt}
\resizebox{0.9\linewidth}{!}{%
\begin{tabular*}{\linewidth}{@{\extracolsep{\fill}}lrrrrrrrr@{}}
\toprule
& \makecell{Nano\\Banana 2} & \makecell{GPT-\\Image-2} & \makecell{Z-Image\\Turbo} & \makecell{Grok\\Imagine} & \makecell{Qwen-Image\\2512} & FLUX.2 & \makecell{SD3.5\\Large} & \makecell{All\\seven} \\
\midrule
Source & 0.620 & 0.707 & 0.898 & 0.787 & 0.992 & 1.188 & 1.174 & 0.910 \\
ReScale & \textbf{0.459} & 0.557 & 0.682 & \textbf{0.492} & \textbf{0.581} & 0.824 & 0.948 & 0.649 \\
Ours & 0.472 & \textbf{0.491} & \textbf{0.673} & 0.514 & 0.664 & \textbf{0.816} & \textbf{0.851} & \textbf{0.640} \\
\bottomrule
\end{tabular*}
}
\end{table*}

\subsection{Physical RoPE Actuator}
\label{sec:actuator}
\paragraph{Encoding the planned edit.}
We encode the target's planned projection in Qwen-Image-Edit~\citep{wu2025qwenimage} \revb{with the spatial RoPE warp of RoPEMover~\citep{oztas2026ropemover}, which translates and scales the target's token coordinates}. For one target, we omit the object index and use apparent scale $a=\exp(c_s-c_d)$ and the planned destination center.

Let $\mathbf{m}_{\mathrm{src}}$ and $\mathbf{m}_{\mathrm{dst}}$ be the source and destination centers on the latent grid, and let $\mathcal{J}_T$ contain source-image tokens selected by target mask $M_T$. We replace each source-token coordinate $\mathbf{p}_j$ by
\begin{equation}
\mathbf{p}'_j=
\begin{cases}
\mathbf{m}_{\mathrm{dst}}+a(\mathbf{p}_j-\mathbf{m}_{\mathrm{src}}), & j\in\mathcal{J}_T,\\
\mathbf{p}_j, & j\notin\mathcal{J}_T.
\end{cases}
\label{eq:rope_warp}
\end{equation}
We recompute spatial RoPE only in the source-image stream; other coordinates and token features remain unchanged. Figure~\ref{fig:method_example}(c) shows the corresponding \rev{$a=0.61$ warp that shrinks the 8-ball and a zoomed crop providing 208 target tokens}. The resulting positional field $\rho$ is the sole geometric condition. \rev{For ground-supported objects it identifies the command: the destination's contact point fixes the depth change and $a$ then fixes the size change, so distinct size-depth commands yield distinct fields, including iso-apparent edits (Appendix~A).} Because the interface depends only on the projected transform, it can also accept requests from another front end.

\paragraph{Supervision from 3D scene edits.}
RoPE specifies the requested projection but not the associated changes in visibility, contact, or shadows. We therefore scale and reposition targets in 3D scenes and re-render them with Kubric/Blender~\citep{greff2022kubric}, keeping the camera, reference object, and environment fixed. Our 2,468 validated pairs from 624 scanned assets cover pure and joint size-depth edits, approximately iso-apparent edits, and identity pairs (Fig.~\ref{fig:counterfactual_pairs}); no real-image editing pairs are used.

During training, the centers in Eq.~\ref{eq:rope_warp} come from rendered masks, and the measured destination-to-source extent ratio $a^{\mathrm{render}}$ replaces analytic $a$. Partially occluded destinations use amodal masks. Appendix~C gives data construction details.

\paragraph{Parameter-efficient adaptation.}
We freeze the pretrained editor and train only rank-16 LoRA parameters $\theta$~\citep{hu2022lora}. With source latent $z_{\mathrm{src}}$, noisy destination latent $x_t$, and flow-matching target $v_t$, we minimize the base objective~\citep{lipman2023flow}:
\begin{equation}
\mathcal{L}_{\mathrm{act}}
=\mathbb{E}\!\left[\left\|v_{\theta}(x_t,t;z_{\mathrm{src}},\rho)-v_t\right\|_2^2\right],
\label{eq:actuator_loss}
\end{equation}
with no geometric, perceptual, or region-specific auxiliary loss. Identity pairs use unchanged images and positional fields, providing preservation supervision. Appendix~B gives architecture and optimization details.

\paragraph{Inference for small targets.}
At inference, $\rho$ uses the planner's destination and analytic scale $a$. Targets with insufficient source-token support are edited in an enlarged crop containing the source, destination, and surrounding scene, then blended back into the image. The system abstains if no crop satisfies the geometry and token-support requirements\rev{, and a final check restores any edited object missing from the output (Appendix~B.6)}.

\Needspace{4\baselineskip}
\section{Experiments}
\label{sec:experiments}
\begin{figure}[t]
\centering
\includegraphics[width=0.86\linewidth]{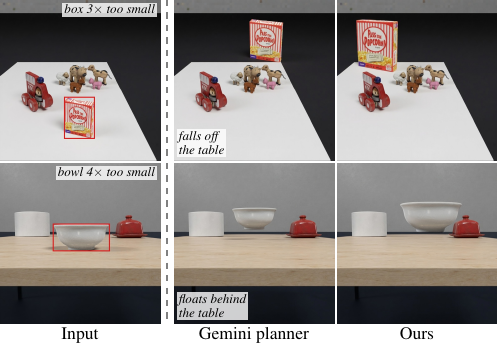}
\captionsetup{skip=2pt,singlelinecheck=false}
\caption{\textbf{Placement after resizing.} Gemini produces off-table or floating destinations.}
\label{fig:exp_planner_qual}
\vspace{-2ex}
\end{figure}

\subsection{Experimental Setup}
\label{sec:exp_setup}
\rev{We evaluate end-to-end correction on GenScale, size estimation and placement in scenes with known geometry, and prescribed editing on real before-and-after photographs. Matched training controls isolate positional commands and 3D supervision; human judgments assess correction quality, placement, and scene effects (Appendix~E).}

We adapt Qwen-Image-Edit-2511~\citep{wu2025qwenimage,qwen2025edit2511} with rank-16 LoRA on 2,468 rendered pairs for ten epochs at $768\times768$. Inference uses 40 steps, guidance 1, and RoPE control for the first 20 steps (Appendix~B).

\subsection{End-to-End Relative-Size Correction}
\label{sec:exp_genscale}
\rev{On GenScale~\citep{li2026genscale} \revb{Task~1}, \MethodName{} reduces relative-size error on all seven generators and matches ReScale overall (Table~\ref{tab:exp_genscale}; Fig.~\ref{fig:exp_genscale_qual}). Blinded human raters prefer its corrections in relative size, content preservation, and naturalness (Table~\ref{tab:exp_human_e2e}).}

\label{sec:exp_efficiency}
On 30 shared GenScale inputs, \MethodName{} \rev{is faster and substantially cheaper than} ReScale (Table~\ref{tab:exp_efficiency}). Savings mainly come from replacing VLM diagnosis and planning with geometric computation. Times sum observation, planning, editing, and restoration (Appendix~E.8).

\rev{Non-target objects not touching edited targets retain their size (median change 0.05\%), and \MethodName{} alters less of the remaining background than ReScale. With text instructions, GPT-Image-2~\revcitep{openai2026gptimage2} and the base Qwen-Image-Edit~\revcitep{qwen2025edit2511} reach errors of 0.78 and 0.88 on a shared subset, versus 0.57-0.58 for \MethodName{} (Appendix~G.4).}

\begin{table}[t]
\centering
\captionsetup{skip=2pt,singlelinecheck=false}
\caption{\rev{\textbf{Human preference on GenScale.} Five blinded raters, 105 pairs sampled where outputs differ; positive preference ($[-1,1]$) favors \MethodName{}.}}
\label{tab:exp_human_e2e}
\small
\renewcommand{\arraystretch}{0.96}
\setlength{\tabcolsep}{3pt}
\resizebox{0.9\linewidth}{!}{%
\begin{tabular*}{\linewidth}{@{\extracolsep{\fill}}lrrrr@{}}
\toprule
& \rev{Ours} & \rev{ReScale} & \rev{Same} & \rev{Pref.} \\
\midrule
\rev{Relative size} & \rev{\textbf{195}} & \rev{78} & \rev{252} & \rev{$+0.22$} \\
\rev{Preservation} & \rev{\textbf{235}} & \rev{30} & \rev{260} & \rev{$+0.39$} \\
\rev{Naturalness} & \rev{\textbf{323}} & \rev{44} & \rev{158} & \rev{$+0.53$} \\
\bottomrule
\end{tabular*}
}
\end{table}

\begin{table}[t]
\centering
\captionsetup{skip=2pt,singlelinecheck=false}
\caption{\textbf{Pipeline efficiency.} Mean summed stage time and estimated cost per image.}
\label{tab:exp_efficiency}
\footnotesize
\renewcommand{\arraystretch}{0.96}
\setlength{\tabcolsep}{3pt}
\resizebox{0.55\linewidth}{!}{%
\begin{tabular}{lrr}
\toprule
Method & Time (s)$\downarrow$ & Cost (\$)$\downarrow$ \\
\midrule
ReScale & 98.0 & 0.696 \\
\MethodName{} & \textbf{52.8} & \textbf{0.037} \\
\bottomrule
\end{tabular}
}
\end{table}

\begin{table}[t]
\centering
\captionsetup{skip=2pt,singlelinecheck=false}
\caption{\textbf{Observer accuracy (\%).} Given true dimensions, identify the wrong-size object and predict its correction within 20\%. Photoreal.: generated images.}
\label{tab:exp_observer}
\small
\renewcommand{\arraystretch}{0.96}
\setlength{\tabcolsep}{3pt}
\resizebox{0.9\linewidth}{!}{%
\begin{tabular*}{\linewidth}{@{\extracolsep{\fill}}lrr@{}}
\toprule
Method & Blender & Photoreal. \\
\midrule
Gemini 3.1 Pro & 44.8 & 38.5 \\
\rev{\MethodName{} (reconstruction)} & \rev{45.7} & \rev{\textbf{47.7}} \\
\MethodName{} & \textbf{56.9} & 44.0 \\
\bottomrule
\end{tabular*}
}
\end{table}

\begin{table}[t]
\centering
\captionsetup{skip=2pt,singlelinecheck=false}
\caption{\textbf{Component swap.} Size error ($\downarrow$) \revb{under three judges} and estimated cost per image; 196 sources. }
\label{tab:exp_swap}
\small
\renewcommand{\arraystretch}{0.96}
\setlength{\tabcolsep}{3pt}
\resizebox{0.9\linewidth}{!}{%
\begin{tabular*}{\linewidth}{@{\extracolsep{\fill}}lcrrrr@{}}
\toprule
 & & \multicolumn{3}{c}{Error by judge} & \\
Front end & Editor & Gemini & GPT & \revb{Claude} & Cost (\$) \\
\midrule
Gemini 3.1 Pro & ReScale & 0.701 & 0.799 & \revb{0.704} & 0.731 \\
\MethodName{} & ReScale & 0.826 & 0.801 & \revb{0.795} & 0.314 \\
Gemini 3.1 Pro & \MethodName{} & \textbf{0.570} & \underline{0.713} & \revb{\textbf{0.602}} & 0.499 \\
\MethodName{} & \MethodName{} & \underline{0.654} & \textbf{0.710} & \revb{\underline{0.669}} & \textbf{0.057} \\
\bottomrule
\end{tabular*}
}
\end{table}

\begin{table}[t]
\centering
\captionsetup{skip=2pt,singlelinecheck=false}
\caption{\textbf{Planner stress tests.} \rev{Automatic pass rate on rendered scenes (\%, $\uparrow$).} Support: full tabletop support.}
\label{tab:exp_planner}
\small
\renewcommand{\arraystretch}{0.96}
\setlength{\tabcolsep}{3pt}
\resizebox{0.9\linewidth}{!}{%
\begin{tabular*}{\linewidth}{@{\extracolsep{\fill}}lrr@{}}
\toprule
& Ours & Gemini \\
\midrule
\rev{Pass rate, rendered} & \textbf{87.9} & 80.2 \\
\quad + support & \textbf{75.9} & 69.0 \\
\bottomrule
\end{tabular*}
}
\vspace{-2ex}
\end{table}

\begin{figure*}[t]
\centering
\includegraphics[width=0.94\linewidth]{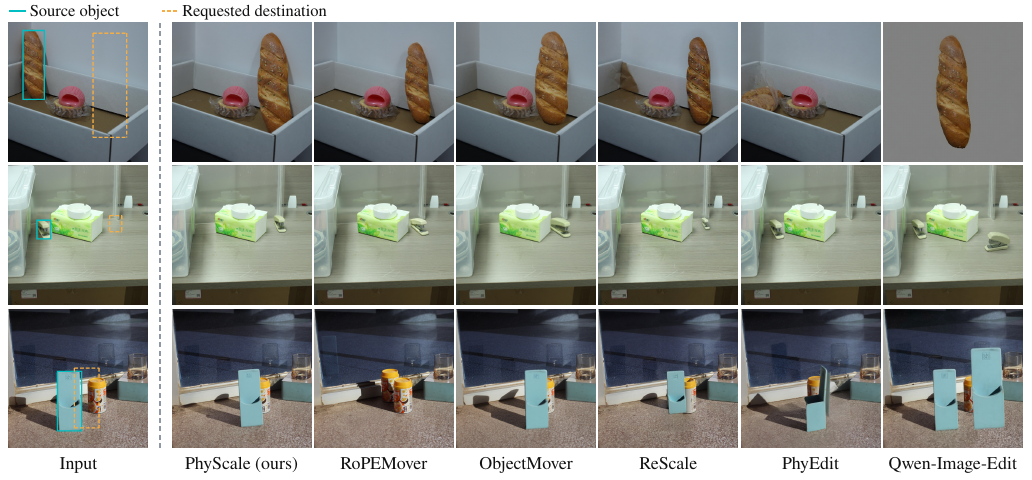}
\captionsetup{skip=2pt,singlelinecheck=false}
\caption{\textbf{Prescribed real-image edits.} Shared source/destination boxes; ReScale uses only its editor\rev{; RoPEMover uses its released drag parameters}.}
\label{fig:exp_objectmover_qual}
\vspace{-2ex}
\end{figure*}

\begin{table*}[t]
\centering
\captionsetup{skip=2pt,singlelinecheck=false,width=\linewidth}
\caption{\textbf{Image editing on ObjectMover.} Text-box and RoPE controls start from the same Qwen-Image-Edit-2511 checkpoint and use identical training pairs and numbers of updates.}
\label{tab:exp_actuator}
\footnotesize
\renewcommand{\arraystretch}{0.96}
\setlength{\tabcolsep}{1.5pt}
\resizebox{0.8\linewidth}{!}{%
\begin{tabular}{@{}lcccccccccc@{}}
\toprule
& \multicolumn{3}{c}{\makecell{Qwen-Image-Edit-2511\\\scriptsize\citep{qwen2025edit2511}}}
& \makecell{PhyEdit\\\scriptsize\citep{xu2026phyedit}}
& \makecell{ReScale\\\scriptsize\citep{li2026genscale}}
& \makecell{ObjectMover\\\scriptsize\citep{yu2025objectmover}}
& \rev{\makecell{RoPEMover\\\scriptsize\citep{oztas2026ropemover}}}
& \makecell{\MethodName{}\\\rev{(ours)}}
& \multicolumn{2}{c}{\makecell{Retrained\\controls}} \\
\cmidrule(lr){2-4}\cmidrule(lr){10-11}
Metric & \makecell{Crop +\\text} & \makecell{RoPE\\only} & Structured & & & & & & \makecell{Text\\boxes} & RoPE \\
\midrule
Target DINO $\uparrow$ & 0.410 & 0.634 & 0.696 & 0.791 & 0.814 & \underline{0.867} & \rev{0.861} & \textbf{0.899} & 0.439 & 0.895 \\
DreamSim $\downarrow$ & 0.622 & 0.399 & 0.326 & 0.216 & 0.177 & \rev{0.140} & \rev{\underline{0.139}} & \textbf{0.105} & 0.620 & 0.108 \\
Identity DINO $\uparrow$ & 0.379 & 0.561 & 0.653 & 0.738 & 0.786 & \underline{0.813} & \rev{0.802} & \textbf{0.856} & 0.365 & 0.856 \\
\bottomrule
\end{tabular}
}
\end{table*}

To separate decision quality from editing, we pair our geometric and ReScale's Gemini front ends with both editors on a shared 196-image GenScale subset, randomly sampled from all seven generators (Table~\ref{tab:exp_swap}). \revb{With either front end and under all three judges, replacing ReScale's editor with ours lowers the error by 0.09--0.17. Our geometric front end makes no VLM call and costs \$0.006 per image, versus \$0.454 for the Gemini front end (Appendix~E.3).}

\rev{ReScale takes its diagnoses from the Gemini judge's own scores of the source image, which match the judge's verdicts at evaluation on 88.5\% of relations, versus 74.9\% for GPT~\revcitep{openai2026gpt61terra} and 78.0\% for Claude~\revcitep{anthropic2026sonnet55} (Appendix~G.2). Under these independent judges, \MethodName{} has the lower error on 700 sources (GPT: 0.680 vs.\ 0.733; Claude: 0.583 vs.\ 0.607), also with the size table withheld from the judges (Appendix~G.1).}

\subsection{Observation and Planning}
\label{sec:exp_observation_planning}
\label{sec:exp_components}

We construct 116 three-object tabletop scenes in Blender with ground-truth geometry, covering cases where in-place size correction causes truncation, occlusion, or undersized targets. Gemini-3.1-Flash-Lite-Image~\citep{google2026flashliteimage} generates matched photorealistic counterparts, of which 109 pass consistency checks (Appendix~E.5).

\paragraph{Size diagnosis.}
\rev{Our observer achieves higher accuracy than Gemini~\revcitep{google2026gemini31pro} in both domains without any VLM call, as does a SAM~3D-based variant (Table~\ref{tab:exp_observer}; Appendix~E.4). For correctly sized objects moved only in depth, a 2D apparent-size reading reports a size error in all 57 controlled cases, our observer in 7 (Appendix~G.10). The correction is also robust to the size prior: substituting LLM-generated sizes reverses fewer than 1\% of correction directions (Appendix~G.9).}

\paragraph{Feasible placement.}
On this set, our planner and Gemini 3.1 Pro receive identical observations, a prescribed resize, \rev{and the same placement requirements}, choosing only depth and lateral position. Both use Blender to execute rendered-scene plans and our frozen editor for photorealistic scenes. Automatic checks test framing, target visibility and size, and neighbor preservation; five blinded raters compare placements. \rev{Our planner passes more automatic checks on rendered scenes (Table~\ref{tab:exp_planner}), and raters prefer its placements in both domains ($+0.122$ rendered, $+0.172$ photorealistic on a $[-1,1]$ scale; Fig.~\ref{fig:exp_planner_qual}).}

\subsection{Controlled Editing and Actuator Adaptation}
\label{sec:exp_actuator}
\begin{table}[t]
\centering
\captionsetup{skip=2pt,singlelinecheck=false}
\caption{\textbf{User study on editing quality.} \rev{Identity} and naturalness scores (0--4) and completed requests (\%). ReScale$^{\dagger}$ uses a separate aggregate.}
\label{tab:exp_human}
\small
\renewcommand{\arraystretch}{0.96}
\setlength{\tabcolsep}{3pt}
\resizebox{1.0\linewidth}{!}{%
\begin{tabular}{@{}lrrrrr@{}}
\toprule
& Qwen & PhyEdit & ReScale$^{\dagger}$ & ObjectMover & \makecell[t]{\MethodName{}\\\rev{(ours)}} \\
\midrule
Identity $\uparrow$ & 3.233 & 2.200 & 3.600 & \underline{3.767} & \textbf{3.861} \\
Naturalness $\uparrow$ & 1.627 & 2.417 & 3.100 & \underline{3.633} & \textbf{3.728} \\
Completed (\%) $\uparrow$ & 2.0 & 31.2 & \underline{78.0} & 74.0 & \textbf{89.9} \\
\bottomrule
\end{tabular}
}
\end{table}
ObjectMover~\citep{yu2025objectmover} provides real before-and-after photographs for prescribed object movement and resizing. All editors receive the same 188 requests; ReScale uses only its editor. Target DINO~\citep{caron2021dino} measures feature cosine similarity between edited and ground-truth destination object crops; DreamSim~\citep{fu2023dreamsim} measures their perceptual distance. Identity DINO uses the same similarity against source-object crops as an identity-preservation proxy. Our editor leads all three metrics (Table~\ref{tab:exp_actuator}; Fig.~\ref{fig:exp_objectmover_qual}).

For real-image edits, raters compare outputs for the same request and assess identity, naturalness, and overall preference. \rev{Our editor matches ObjectMover and exceeds the others in identity and naturalness, and completes the most requests} (Table~\ref{tab:exp_human}).

\paragraph{\texorpdfstring{\rev{Comparison with RoPEMover.}}{Comparison with RoPEMover.}}
\rev{\revb{Both editors use the same backbone and RoPE warp. Executing} the same \MethodName{} plans on GenScale images, ours realizes 77\% of the planned size changes and RoPEMover 17\%; on 150 held-out size-depth renders, ours reaches 0.921 target-box DINO versus 0.768 (Appendix~G.5).}

\paragraph{Positional control.}
To test whether positional commands help beyond fine-tuning, we train two models with matched backbones, data, budgets, and \rev{identical} visual inputs. One uses RoPE; the other receives source and destination boxes in text. \rev{RoPE improves command following on both ObjectMover and held-out renders} (Table~\ref{tab:exp_actuator}).

\begin{figure}[t]
\centering
\includegraphics[width=1.0\linewidth]{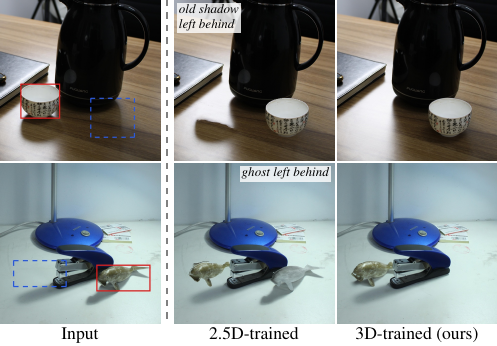}
\caption{\textbf{Scene effects.} Real-image edits show old shadows or object remnants after 2.5D training.}
\label{fig:exp_supervision_qual}
\vspace{-1em}
\end{figure}

\paragraph{Supervision and scene effects.}
\rev{Size-depth training improves resizing: replacing these commands with translation-only commands on the same scenes lowers held-out silhouette IoU from 0.771 to 0.629 (Appendix~G.6).} \rev{To test scene effects, we compare} full-scene 3D re-rendering with source-object warping and compositing (2.5D), using the same editor, source scenes, and training budget. \rev{3D supervision yields higher target fidelity on both real photographs and held-out renders, and three blinded raters prefer its physical plausibility (contact, shadows, and occlusion).} Figure~\ref{fig:exp_supervision_qual} shows real-image examples.

\Needspace{6\baselineskip}
\section{Conclusion}
\label{sec:conclusion}
We presented \MethodName{}, a framework that separates physical-size inference and planning from learned image editing. Its observer estimates size errors from knowledge-base dimensions and scene geometry, its planner selects size, depth, and placement under prioritized constraints, and its RoPE-controlled editor executes the resulting projection using limited training pairs. Across GenScale and controlled editing, \MethodName{} \rev{matches ReScale in correction performance while providing stronger prescribed-edit fidelity, higher human preference, and substantially lower inference cost.} These results show that explicit geometric computation and data-efficient learned execution provide a practical alternative to VLM-based correction decisions.
\paragraph{Limitations and Future Work.}
\rev{\MethodName{} relies on pretrained vision modules, category-level size priors, and ground-supported objects, and corrects images after generation; future work should incorporate metric size priors directly into generative-model training.}

{
    \small
    \bibliographystyle{ieeenat_fullname}
    \bibliography{main}
}

\clearpage
\ifarxiv\else
    \setcounter{page}{1}
\fi
\maketitlesupplementary
\ifarxiv
    \pdfbookmark[0]{Supplementary Material}{supplementary}
\fi

\appendix
\renewcommand{\FloatBarrier}{}
\setcounter{dbltopnumber}{3}
\subsection*{Reproducibility Statement}
Appendix~A provides the derivation of the size-depth parameterization. Appendices~B-D document the actuator architecture, training and inference settings, construction and validation of the rendered training pairs, and observer and planner implementations, including planner pseudocode. Appendix~E describes the evaluation datasets and protocols, baseline configurations, scoring and statistical procedures, ablations, human studies, and runtime measurements and cost-estimation assumptions. \rev{Appendix~F provides additional qualitative comparisons, and Appendix~G adds evaluations with independent judges, human raters, alternative editors, matched retraining, and planner and observer analyses.} \revb{We will publicly release all code, training data, and model weights upon publication.}
\setlength{\textfloatsep}{10pt plus 2pt minus 2pt}
\setlength{\floatsep}{10pt plus 2pt minus 2pt}
\setlength{\intextsep}{9pt plus 2pt minus 2pt}
\makeatletter
\setlength{\@fptop}{0pt}
\setlength{\@fpsep}{12pt}
\setlength{\@fpbot}{0pt plus 1fil}
\makeatother

\begin{center}
{\LARGE\bfseries Appendix}\par
\end{center}
\pdfbookmark[0]{Appendix}{appendix-overview}
\noindent This appendix provides derivations, implementation and data details, and experimental protocols with additional results. Sections~\ref{app:scale_depth_derivation}--\ref{app:observer-planner} supplement the method; Section~\ref{app:experiments} supplements the experiments. Section~\ref{app:additional_visualizations} provides additional visual comparisons\rev{, and Section~\ref{app:rev} adds evaluations with independent judges, human raters, preservation measurements, alternative editors, matched retraining, planner and observer analyses, and size-prior robustness}.

\medskip
\needspace{8\baselineskip}
\noindent\textbf{Contents}\par\smallskip
\begingroup
\footnotesize
\setlength{\tabcolsep}{3pt}
\renewcommand{\arraystretch}{1.00}
\noindent\begin{tabularx}{0.94\columnwidth}{@{}Xr@{}}
\hspace*{0em}\hyperref[app:scale_depth_derivation]{\textbf{\ref*{app:scale_depth_derivation}\quad Derivation of the Size--Depth Parameterization}} & \pageref{app:scale_depth_derivation} \\
\hspace*{0em}\hyperref[app:actuator_impl]{\textbf{\ref*{app:actuator_impl}\quad Actuator Implementation Details}} & \pageref{app:actuator_impl} \\
\hspace*{1em}\hyperref[app:actuator_backbone]{\ref*{app:actuator_backbone}\quad Backbone and Parameter-Efficient Adaptation} & \pageref{app:actuator_backbone} \\
\hspace*{1em}\hyperref[app:rope_impl]{\ref*{app:rope_impl}\quad RoPE Command Injection} & \pageref{app:rope_impl} \\
\hspace*{1em}\hyperref[app:actuator_training]{\ref*{app:actuator_training}\quad Training Objective and Optimization} & \pageref{app:actuator_training} \\
\hspace*{1em}\hyperref[app:actuator_inference]{\ref*{app:actuator_inference}\quad Inference} & \pageref{app:actuator_inference} \\
\hspace*{1em}\hyperref[app:object_centric]{\ref*{app:object_centric}\quad Object-Centric Resampling} & \pageref{app:object_centric} \\
\hspace*{1em}\hyperref[app:restoration]{\ref*{app:restoration}\quad Restoration of Lost Targets} & \pageref{app:restoration} \\
\hspace*{0em}\hyperref[app:actuator_data]{\textbf{\ref*{app:actuator_data}\quad Training Pairs from 3D Scene Edits}} & \pageref{app:actuator_data} \\
\hspace*{1em}\hyperref[app:data_rendering]{\ref*{app:data_rendering}\quad Motivation and Rendering Setup} & \pageref{app:data_rendering} \\
\hspace*{1em}\hyperref[app:data_distribution]{\ref*{app:data_distribution}\quad Scene and Command Distribution} & \pageref{app:data_distribution} \\
\hspace*{1em}\hyperref[app:data_construction]{\ref*{app:data_construction}\quad Constraint-Driven Row Construction} & \pageref{app:data_construction} \\
\hspace*{1em}\hyperref[app:data_tokens]{\ref*{app:data_tokens}\quad Token-Support Constraint} & \pageref{app:data_tokens} \\
\hspace*{1em}\hyperref[app:data_materialization]{\ref*{app:data_materialization}\quad Rendered Geometry and Materialization} & \pageref{app:data_materialization} \\
\hspace*{1em}\hyperref[app:data_certification]{\ref*{app:data_certification}\quad Physical Certification} & \pageref{app:data_certification} \\
\hspace*{0em}\hyperref[app:observer-planner]{\textbf{\ref*{app:observer-planner}\quad Observer and Planner Details}} & \pageref{app:observer-planner} \\
\hspace*{1em}\hyperref[app:observer]{\ref*{app:observer}\quad Physical Observer} & \pageref{app:observer} \\
\hspace*{1em}\hyperref[app:planner]{\ref*{app:planner}\quad Size and Placement Planner Details} & \pageref{app:planner} \\
\hspace*{2em}\hyperref[app:planner_abstention]{\ref*{app:planner_abstention}\quad \rev{Planner Configuration for the Placement Stress Tests}} & \pageref{app:planner_abstention} \\
\hspace*{2em}\hyperref[app:planner_pseudocode]{\ref*{app:planner_pseudocode}\quad Planner Pseudocode} & \pageref{app:planner_pseudocode} \\
\hspace*{0em}\hyperref[app:experiments]{\textbf{\ref*{app:experiments}\quad Experimental Protocols and Additional Results}} & \pageref{app:experiments} \\
\hspace*{1em}\hyperref[app:exp_setup]{\ref*{app:exp_setup}\quad Implementation and Statistical Setup} & \pageref{app:exp_setup} \\
\hspace*{1em}\hyperref[app:exp_genscale]{\ref*{app:exp_genscale}\quad GenScale Scoring} & \pageref{app:exp_genscale} \\
\hspace*{1em}\hyperref[app:exp_swap]{\ref*{app:exp_swap}\quad Front-End--Actuator Swap} & \pageref{app:exp_swap} \\
\hspace*{1em}\hyperref[app:exp_observer]{\ref*{app:exp_observer}\quad Observer Evaluation} & \pageref{app:exp_observer} \\
\hspace*{2em}\hyperref[app:exp_observer_reconstruction]{\ref*{app:exp_observer_reconstruction}\quad Reconstruction-Based Observer Variant} & \pageref{app:exp_observer_reconstruction} \\
\hspace*{1em}\hyperref[app:exp_planner]{\ref*{app:exp_planner}\quad Planner Stress Tests} & \pageref{app:exp_planner} \\
\hspace*{2em}\hyperref[app:stress_construction]{\ref*{app:stress_construction}\quad Construction and Automatic Evaluation} & \pageref{app:stress_construction} \\
\hspace*{2em}\hyperref[app:stress_photorealistic]{\ref*{app:stress_photorealistic}\quad Photorealistic Transfer} & \pageref{app:stress_photorealistic} \\
\hspace*{2em}\hyperref[app:stress_examples]{\ref*{app:stress_examples}\quad Complete Stress-Test Galleries} & \pageref{app:stress_examples} \\
\hspace*{1em}\hyperref[app:exp_actuator]{\ref*{app:exp_actuator}\quad Controlled Editing and Training Ablations} & \pageref{app:exp_actuator} \\
\hspace*{2em}\hyperref[app:exp_interface]{\ref*{app:exp_interface}\quad Textual versus Positional Commands} & \pageref{app:exp_interface} \\
\hspace*{2em}\hyperref[app:exp_supervision]{\ref*{app:exp_supervision}\quad 3D versus 2.5D Supervision} & \pageref{app:exp_supervision} \\
\hspace*{2em}\hyperref[app:exp_training_curves]{\ref*{app:exp_training_curves}\quad Training Dynamics} & \pageref{app:exp_training_curves} \\
\hspace*{2em}\hyperref[app:exp_command_diagnostics]{\ref*{app:exp_command_diagnostics}\quad Command and Geometric Diagnostics} & \pageref{app:exp_command_diagnostics} \\
\hspace*{1em}\hyperref[app:exp_human]{\ref*{app:exp_human}\quad User Study on Editing Quality} & \pageref{app:exp_human} \\
\hspace*{1em}\hyperref[app:exp_efficiency]{\ref*{app:exp_efficiency}\quad Pipeline Time and Cost} & \pageref{app:exp_efficiency} \\
\hspace*{0em}\hyperref[app:additional_visualizations]{\textbf{\ref*{app:additional_visualizations}\quad Additional Visualizations}} & \pageref{app:additional_visualizations} \\
\hspace*{1em}\hyperref[app:qual_genscale]{\ref*{app:qual_genscale}\quad GenScale Relative-Size Correction} & \pageref{app:qual_genscale} \\
\hspace*{1em}\hyperref[app:qual_objectmover]{\ref*{app:qual_objectmover}\quad Prescribed Object Movement and Resizing} & \pageref{app:qual_objectmover} \\
\hspace*{1em}\hyperref[app:qual_planner]{\ref*{app:qual_planner}\quad Photorealistic Planner Comparisons} & \pageref{app:qual_planner} \\
\hspace*{1em}\hyperref[app:qual_supervision]{\ref*{app:qual_supervision}\quad 3D versus 2.5D Supervision} & \pageref{app:qual_supervision} \\
\hspace*{0em}\hyperref[app:rev]{\textbf{\ref*{app:rev}\quad Additional Evaluations}} & \pageref{app:rev} \\
\hspace*{1em}\hyperref[app:rev_judges]{\ref*{app:rev_judges}\quad Independent Judges and Human Evaluation on GenScale} & \pageref{app:rev_judges} \\
\hspace*{1em}\hyperref[app:rev_coupling]{\ref*{app:rev_coupling}\quad \rev{Agreement Between ReScale's Diagnoses and the Judges}} & \pageref{app:rev_coupling} \\
\hspace*{1em}\hyperref[app:rev_heldout]{\ref*{app:rev_heldout}\quad Sources Not Used During Development} & \pageref{app:rev_heldout} \\
\hspace*{1em}\hyperref[app:rev_preservation]{\ref*{app:rev_preservation}\quad Preservation and General-Purpose Editor Baselines} & \pageref{app:rev_preservation} \\
\hspace*{1em}\hyperref[app:rev_ropemover]{\ref*{app:rev_ropemover}\quad RoPEMover as the Pipeline Editor} & \pageref{app:rev_ropemover} \\
\hspace*{1em}\hyperref[app:rev_retrain]{\ref*{app:rev_retrain}\quad Matched Retraining} & \pageref{app:rev_retrain} \\
\hspace*{1em}\hyperref[app:rev_relation]{\ref*{app:rev_relation}\quad Reordering Planner Priorities for Spatial Relations} & \pageref{app:rev_relation} \\
\hspace*{1em}\hyperref[app:rev_inplace]{\ref*{app:rev_inplace}\quad \rev{In-Place Resizing versus Full Placement}} & \pageref{app:rev_inplace} \\
\hspace*{1em}\hyperref[app:rev_sizeprior]{\ref*{app:rev_sizeprior}\quad Robustness to the Size Prior} & \pageref{app:rev_sizeprior} \\
\hspace*{1em}\hyperref[app:rev_2d]{\ref*{app:rev_2d}\quad \rev{Apparent-Size Baseline}} & \pageref{app:rev_2d} \\
\hspace*{1em}\hyperref[app:rev_camera]{\ref*{app:rev_camera}\quad Sensitivity to the Assumed Camera} & \pageref{app:rev_camera} \\
\hspace*{1em}\hyperref[app:rev_heldout_swap]{\ref*{app:rev_heldout_swap}\quad Editor Comparison on Sources Outside Development} & \pageref{app:rev_heldout_swap} \\
\end{tabularx}
\endgroup
\clearpage

\section{Derivation of the Size--Depth Parameterization}
\label{app:scale_depth_derivation}
This section derives the command convention in Sec.~\ref{sec:problem_formulation}. Physical size is denoted by $S$, camera-frame depth by $d$, projected linear extent by $\ell$, and the apparent scaling factor by $a$. VAE latents retain the symbol $z$. \rev{As in the main text, positive $c_d$ moves an object farther from the camera.}

\paragraph{Perspective projection.}
When an object's depth variation is small relative to its camera distance, the local weak-perspective approximation~\citep{hartley2004multiple} gives
\begin{equation}
\Delta x\simeq\frac{f}{d_i}\Delta X,\qquad \Delta y\simeq\frac{f}{d_i}\Delta Y,
\label{eq:app_perspective_projection}
\end{equation}
and hence
\begin{equation}
\ell_i\approx\kappa_i\frac{fS_i}{d_i},
\label{eq:app_projected_extent}
\end{equation}
where $\kappa_i$ captures the shape, pose, viewpoint, and chosen extent measurement. Taking logarithms yields
\begin{equation}
\log\ell_i\approx\log S_i-\log d_i+\log f+\log\kappa_i.
\label{eq:app_log_extent}
\end{equation}
For two objects viewed by the same camera, the focal length cancels:
\begin{equation}
\log\frac{\ell_T}{\ell_R}\approx\log\frac{S_T}{S_R}-\log\frac{d_T}{d_R}+\log\frac{\kappa_T}{\kappa_R}.
\label{eq:app_pair_extent}
\end{equation}
This relation exposes the ambiguity: the same apparent ratio can arise from different size, depth, and pose configurations. It does not itself estimate physical size. The observer instead fits knowledge-base geometry under its estimated camera and orientation (Sec.~\ref{sec:observer}).

\paragraph{Physical commands.}
For an edited object, with the index omitted, define
\begin{equation}
\mathbf{c}=(c_s,c_d),\qquad c_s=\log\frac{S'}{S},\qquad c_d=\log\frac{d'}{d}.
\label{eq:app_action}
\end{equation}
The physical update is therefore
\begin{equation}
S'=S\exp(c_s),\qquad d'=d\exp(c_d).
\label{eq:app_depth_update}
\end{equation}
Subtracting Eq.~\eqref{eq:app_log_extent} between destination and source gives
\begin{equation}
\log\frac{\ell'}{\ell}\approx c_s-c_d+\log\frac{\kappa'}{\kappa}.
\label{eq:app_extent_change}
\end{equation}
Under approximately unchanged pose and foreshortening, $\kappa'\approx\kappa$, this reduces to
\begin{equation}
\log\frac{\ell'}{\ell}\approx c_s-c_d,\qquad a\triangleq\exp(c_s-c_d)\approx\frac{\ell'}{\ell}.
\label{eq:app_scale}
\end{equation}
Thus $(c_s,0)$ changes size at fixed depth, while $(0,c_d)$ changes depth at fixed size. Equal size and depth commands, $c_d=c_s$, preserve the predicted projected extent. For $c_s\ne0$, the compensation ratio is
\begin{equation}
\beta\triangleq\frac{c_d}{c_s},
\label{eq:app_beta}
\end{equation}
with $\beta=0$ for a pure size change and $\beta=1$ for an iso-apparent edit. Preserving projected extent need not preserve position, visibility, contact, or the full image. During training, the actuator uses scale and destination measurements from rendered masks rather than assuming this local approximation is exact.

\rev{\paragraph{Commands and the positional field.}
The actuator is conditioned on the apparent scale $a$ and the destination center $\mathbf m_{\mathrm{dst}}$ rather than on $(c_s,c_d)$ directly. For a ground-supported object this loses no information. Let $y_c$ be the image row of its ground contact and $y_h$ the horizon row. For a point on the ground plane, depth is a strictly decreasing function of its row offset below the horizon; for a camera at height $h_c$ with small tilt, $d\approx f h_c/(y_c-y_h)$. A depth command therefore moves the contact row to
\begin{equation}
y_c'-y_h\approx(y_c-y_h)\exp(-c_d),
\label{eq:app_contact_row}
\end{equation}
and the lateral offset sets its column. Conversely, the field supplies $a$ and $\mathbf m_{\mathrm{dst}}$, which place the destination contact row at the row of $\mathbf m_{\mathrm{dst}}$ plus $a$ times the source offset between contact and center; Eq.~\eqref{eq:app_contact_row} then gives $c_d$, and $c_s=\log a+c_d$. Two different size--depth commands for the same ground-supported object therefore never share a positional field. In particular, an iso-apparent edit ($a=1$, $c_d=c_s\neq0$) differs from the identity through its displaced destination. A shared field can arise only for objects that leave the support plane, which the planner does not produce.}

\FloatBarrier
\section{Actuator Implementation Details}
\label{app:actuator_impl}

\subsection{Backbone and Parameter-Efficient Adaptation}
\label{app:actuator_backbone}

We instantiate the actuator with Qwen-Image-Edit-2511~\revcitep{wu2025qwenimage,qwen2025edit2511}. The pretrained transformer, text encoder, and VAE are frozen throughout actuator training. The source image is provided through the edit-image stream with its conditioning timestep set to zero (\texttt{zero\_cond\_t=True}).

\arxivwording{For parameter-efficient transformer adaptation~\citep{tan2024adapter}, we use LoRA~\citep{hu2022lora}}{We adapt the transformer with LoRA \citep{hu2022lora}} of rank $r_{\mathrm{LoRA}}=16$ and scaling $\alpha_{\mathrm{LoRA}}=16$. For a pretrained linear layer $W$, the adapted layer is
\begin{equation}
    Wx + \frac{\alpha_{\mathrm{LoRA}}}{r_{\mathrm{LoRA}}}BAx ,
\end{equation}
where $A$ is initialized from $\mathcal{N}(0,0.02^2)$ and $B$ is initialized to zero. Consequently, the initial actuator is exactly equivalent to the pretrained editor before optimization.

LoRA modules are inserted into 12 linear projections in each transformer block:
\begin{equation}
\begin{split}
    &\{\texttt{to\_q},\texttt{to\_k},\texttt{to\_v},
      \texttt{to\_out.0}\},\\
    &\{\texttt{add\_q\_proj},\texttt{add\_k\_proj},\\
    &\phantom{\{}\texttt{add\_v\_proj},\texttt{to\_add\_out}\},\\
    &\{\texttt{img\_mlp.net.2},\texttt{txt\_mlp.net.2},\\
    &\phantom{\{}\texttt{img\_mod.1},\texttt{txt\_mod.1}\}.
\end{split}
\end{equation}
This yields 117.97M trainable parameters while leaving the $20$B-parameter backbone frozen. Backbone weights are stored in bfloat16; LoRA parameters and LoRA computation use FP32.

\subsection{RoPE Command Injection}
\label{app:rope_impl}

Qwen-Image represents image position using three-dimensional rotary positional embeddings. The RoPE head dimensions are divided across the frame and spatial axes $(t,h,w)$ as
\begin{equation}
    (n_t,n_h,n_w)=(16,56,56),
\end{equation}
with RoPE base $b_{\mathrm{RoPE}}=10{,}000$. This base is distinct from the learned LoRA parameters $\theta$ in Sec.~\ref{sec:actuator}. For coordinate value $v$ and an axis of dimension $n$, we construct the corresponding rotary frequencies in the standard manner,
\begin{equation}
    \omega_k(v)
    =
    v\,b_{\mathrm{RoPE}}^{-2k/n},
    \qquad
    k=0,\ldots,n/2-1.
\end{equation}
Spatial grid coordinates are centered before conversion to frequencies. The frame coordinate distinguishes the two image streams: the noisy output stream uses frame index $0$, while the source edit stream uses frame index $1$. We modify only the spatial $(h,w)$ coordinates of target tokens and keep the frame coordinate unchanged.

\paragraph{Target-token support.}
Let $M_T$ be the source target mask and let the image resolution be $R$. Qwen-Image uses a spatial stride of $p=16$, yielding a $G\times G$ image-token grid with $G=R/16$. We average-pool $M_T$ over each $16\times16$ patch and select
\begin{equation}
    \mathcal{J}_T
    =
    \left\{
        j:
        \operatorname{AvgPool}_{16\times16}(M_T)_j > 0.25
    \right\}.
    \label{eq:app_token_selection}
\end{equation}
The same selection rule is used during training and inference. The effective number of command-carrying tokens is therefore
\begin{equation}
    N_T = |\mathcal{J}_T|
    \approx \frac{|M_T|}{16^2}.
    \label{eq:app_token_count}
\end{equation}

\paragraph{Spatial warp.}
For each selected token, let $\mathbf{p}_j=(h_j,w_j)$ denote its source-grid coordinate, $\mathbf{m}_{\mathrm{src}}$ the source target center, and $\mathbf{m}_{\mathrm{dst}}$ the destination center, all on the latent grid. At inference the planner supplies the destination; during training both centers come from rendered masks. From Sec.~\ref{sec:problem_formulation}, the required apparent linear scale is
\begin{equation}
    a=\exp(c_s-c_d).
\end{equation}
We assign the target token the transformed coordinate
\begin{equation}
    \mathbf p'_j
    =
    \mathbf{m}_{\mathrm{dst}}+a(\mathbf{p}_j-\mathbf{m}_{\mathrm{src}}),
    \qquad j\in\mathcal J_T ,
    \label{eq:app_rope_map}
\end{equation}
and recompute its RoPE frequencies after the implementation's valid-grid clamp. Equation~\eqref{eq:app_rope_map} is the affine field in Eq.~\ref{eq:rope_warp} before this boundary safeguard. All non-target source tokens and all noisy-output tokens retain their original coordinates. During training, $a$ in this map is replaced by the measured rendered extent ratio $a^{\mathrm{render}}$, as specified in Sec.~\ref{sec:actuator}; at inference it remains $\exp(c_s-c_d)$.

Because the source image occupies the second image stream, the modified frequencies are written only to the corresponding source-stream token positions. For a resolution $R$,
\begin{equation}
    N_{\mathrm{img}}=(R/16)^2,
\end{equation}
and the source-stream target token $j$ is located at offset $N_{\mathrm{img}}+j$. We explicitly verify that the RoPE map remains inside the source stream and that the positional-embedding tensor contains exactly the expected two image streams.

The resulting conditioning retains the target's visual content but associates its tokens with the destination geometry requested by the planner. The learned editor therefore resolves the disagreement between source appearance and modified positional structure through generative re-rendering rather than pixel-space transformation.

\subsection{Training Objective and Optimization}
\label{app:actuator_training}

For each certified pair, let $I_{\mathrm{src}}$ and $I_{\mathrm{dst}}$ denote the source and rendered destination images, and let $z_{\mathrm{src}}$ and $z_{\mathrm{dst}}$ be their VAE latents. We sample a training timestep uniformly from the base model's 1,000 diffusion timesteps, construct the noisy destination latent $x_t$, and use the same flow-matching target $v_t$ as the pretrained editor. The only task-specific conditioning is the source image together with the warped positional field:
\begin{equation}
    \mathcal{L}_{\mathrm{act}}
    =
    \mathbb{E}_{I_{\mathrm{src}},I_{\mathrm{dst}},t,\epsilon}
    \left[
        \left\|
        v_\theta
        \left(
            x_t,t;
            z_{\mathrm{src}},
            \rho
        \right)
        -
        v_t
        \right\|_2^2
    \right].
    \label{eq:app_actuator_loss}
\end{equation}
We use no segmentation, geometric-regression, perceptual, ranking, or region-weighted auxiliary loss.

The field $\rho$ depends on the source mask, source and destination centers, and apparent scale, not on the two physical commands alone. \rev{For ground-supported objects, different size--depth commands always produce different fields (Appendix~\ref{app:scale_depth_derivation}).} Identity examples use $I_{\mathrm{dst}}=I_{\mathrm{src}}$ exactly, $\mathbf{c}=(0,0)$, and no lateral displacement. For these examples no RoPE transformation is applied. The 364 identity rows (14.75\% of the \rev{training set}) provide direct preservation supervision for the target, reference, and background.

We optimize the LoRA parameters with AdamW~\citep{loshchilov2019adamw} using \[ \mathrm{lr}=10^{-4},\quad (\beta_1,\beta_2)=(0.9,0.95),\quad \mathrm{weight\ decay}=0, \] and clip the global gradient norm to $1.0$. The final model is trained at $768\times768$ resolution. Each of the 2,468 certified training rows is seen exactly ten times, for 24,680 sample exposures and 4,936 optimizer updates at global batch five. Within each epoch, rows are deterministically shuffled using a hash of the row identifier, epoch, and global seed, so every training row has identical exposure.

Our implementation uses functional DiffSynth~\revcitep{modelscope2026diffsynth} forward calls rather than the standard \texttt{DDP.forward} path. We therefore explicitly average LoRA gradients across workers after backpropagation. Synchronization is checked both after the first update and at the final checkpoint by verifying cross-worker gradient and parameter agreement.

\paragraph{\rev{Conditioning prompts.}}
Training uses global batch five, deterministic row shuffling with seed zero, a constant learning rate without warmup, and the prompt ``Move the \{target\} to a different position on the floor, adjusting its apparent size for perspective. Keep the \{reference\}, the floor and the lighting exactly unchanged.'' \rev{The correction pipeline} instead uses ``Move or resize the \{target\} as indicated. Keep the \{reference\} and the background exactly unchanged.''

\paragraph{Software and version records.}
The transformer is \path{Qwen/Qwen-Image-Edit-2511}; the text encoder and VAE are from \path{Qwen/Qwen-Image}, with the \path{Qwen/Qwen-Image-Edit} processor. The framework is DiffSynth-Studio at commit \path{744bf641d9ea02adba77b7407da039a053f2b74d}. \rev{Evaluation uses revision prefixes} \texttt{6f3ccc0b56e4}, \texttt{75e0b4be04f6}, and \texttt{ac7f9318f633}, respectively.

\subsection{Inference}
\label{app:actuator_inference}

Unless otherwise stated, we run the Qwen-Image editor for 40 denoising steps with classifier-free guidance scale $1.0$. The spatial RoPE modification is active during the first 20 denoising steps and disabled thereafter. The source image is supplied without automatic resizing and with \texttt{zero\_cond\_t=True}. GenScale uses deterministic source--command seeds; the ObjectMover and stress-test benchmarks use seed 42.

At inference time, the actuator receives the source mask, the planned destination, and $(c_s,c_d)$\rev{; in the GenScale pipeline, the size and depth commands are the planner's commands scaled by the execution factor $\eta=1.08$, and the destination follows from these scaled commands and the planned lateral offset (Appendix~\ref{app:planner_search})}. It rasterizes the source target onto the latent grid using Eq.~\eqref{eq:app_token_selection}, constructs the transformed coordinates using Eq.~\eqref{eq:app_rope_map}, and applies the resulting RoPE field to the source edit stream.

\subsection{Object-Centric Resampling}
\label{app:object_centric}
RoPE control is carried only by the source target's tokens. For small objects, \rev{inference} therefore selects the largest local crop covering both source and planned destination while providing at least 16 token cells for each where feasible. The entire crop is resized to $768\times768$, edited, resized back with Lanczos interpolation, and blended into the source with a six-pixel feather (Gaussian $\sigma=3$). This is a local scene edit, not an isolated object cutout. Planned targets that disappear are restored \rev{as described in Appendix~\ref{app:restoration}}. \rev{This 16-token threshold applies at inference;} the 20-token floor in Appendix~\ref{app:actuator_data} is a training-data certification threshold.

\subsection{\texorpdfstring{\rev{Restoration of Lost Targets}}{Restoration of Lost Targets}}
\label{app:restoration}
\rev{Before delivery, a final check verifies that every object the editor actually edited is present at its planned destination. Grounding DINO~\revcitep{liu2023groundingdino} and SAM~\revcitep{kirillov2023sam} detect the object by name in the delivered image; a detection counts only if it overlaps the planned destination mask by more than 10\%. If none does, detection is repeated on an enlarged crop around the destination, and if that also fails, the destination patch is compared with the source object; only a dissimilar patch marks the target as lost, while an uncertain case keeps the delivered image. For a lost target, the source object's pixels, with the masks of all other objects removed, are warped by the planned similarity transform with mask-premultiplied color at the native $1024\times1024$ resolution and composited at the destination, except where an object whose ground contact is closer to the camera occludes it. The restored image is accepted only if the re-check then finds the target at its destination; otherwise the unrestored image is kept. Restoration runs no diffusion and uses no VLM; \rev{on the 2,800 GenScale sources} it re-inserts 105 objects in 101 sources.}

\FloatBarrier
\section{Training Pairs from 3D Scene Edits}
\label{app:actuator_data}

\subsection{Motivation and Rendering Setup}
\label{app:data_rendering}

A source--target pair suitable for physical actuation must distinguish physical size from depth even when their effects on apparent extent cancel. We therefore construct the final actuator supervision by modifying explicit 3D scene geometry and re-rendering the complete scene using Kubric/Blender \citep{greff2022kubric}. Source and destination images share scene identity, camera, environment, reference object, and rendering configuration; only the prescribed target transformation and associated scene interaction change.

For each target, we apply a world-space scale transformation corresponding to $c_s$ and a camera-relative depth transformation corresponding to $c_d$. In addition, an independent lateral translation can be introduced along the camera-right direction. Because this direction is orthogonal to the viewing direction and camera up vector, it changes image-space horizontal placement without changing the requested depth variable. All destination images are then re-rendered, allowing perspective, occlusion, exposed surfaces, contact, shadows, and indirect illumination to respond to the transformation.

\subsection{Scene and Command Distribution}
\label{app:data_distribution}

The final \rev{training set} is rendered at $768\times768$ resolution and contains 2,468 certified source--destination pairs. The candidate pool contains 802 Google Scanned Objects assets, after holding out 231 of 1,033 available assets; the certified rows use 624 assets in 500 asset pairs. The construction limits each asset to at most two pair assignments.

To avoid tying the actuator to a single synthetic camera, we sample from a factorized camera grid comprising
\begin{itemize}
    \item focal lengths
    $\{60,75,89.06,110,135\}$ mm;
    \item camera heights
    $\{0.15,0.22,0.32,0.45,0.62\}$ m;
    \item camera distances
    $\{1.6,2.2,3.0,4.0\}$ m;
    \item horizontal camera offsets
    $\{-1,-0.5,0,0.5,1\}$ m; and
    \item horizon locations
    $\{0.36,0.42,0.50\}H$.
\end{itemize}
The Cartesian product contains 1,500 camera configurations, of which 835 occur in certified rows. Camera tilt is restricted to at most $5^\circ$. We use 381 indoor and outdoor HDR environments after excluding pure-sky environments and limiting night scenes.

Objects are rendered at yaw rotations $\{0^\circ,90^\circ,180^\circ,270^\circ\}$. Lateral destination motion is stratified into approximately $0$--$2$, $2$--$8$, and $8$--$40$ latent-token bins; the realized maximum displacement is approximately 39 token cells.

Physical commands cover
\begin{equation}
    c_s\in[-0.6,0.8],
    \qquad
    c_d\in[-0.6,0.6],
\end{equation}
with five size bins and four depth bins. The dataset includes pure-size, pure-depth, joint, compensation, identity/no-op, and translation-only command families. The design allocates approximately 15\% to exact identity examples; 364 survive certification. To stress instance binding, a same-category bystander is included with probability $0.15$. We also enforce coverage of the practically important $c_s>0,c_d>0$ regime, where physical enlargement is partially compensated by moving the object farther from the camera.

The certified family counts are 1,664 joint, 364 identity, 171 compensation, 150 size-only, 112 depth-only, and seven translation-only rows. Occlusion counts are 1,862 unoccluded, 331 target-in-front, and 275 target-behind rows. The same-category bystander flag occurs in 30.3\% of retained rows; the 0.15 probability above is a per-pair construction draw, not the realized row proportion. Physical enlargement with motion away from the camera ($c_s>0,c_d>0$) occurs in 26.4\% of rows.

\subsection{Constraint-Driven Row Construction}
\label{app:data_construction}

Naively sampling 3D placements and rejecting invalid scenes is inefficient because token support, image-space displacement, frame validity, and physical placement are strongly coupled. We instead solve backward from the desired training cell.

For a fixed asset, camera, and command, projected support is monotonic in camera-relative depth. We therefore use a one-dimensional bisection search to find a depth satisfying the requested target-token band. Given this depth, the projected horizontal centroid varies affinely with translation along the camera-right axis, allowing the required screen-space displacement to be solved directly. This produces a feasible 3D proposal before expensive rendering.

We initially construct 2,500 candidate rows. After rendering and certification, 2,468 are retained. This procedure covers 332 of the 360 joint combinations of command, token-support, and displacement strata considered during data construction.

\subsection{Token-Support Constraint}
\label{app:data_tokens}

Because the actuator command is written only on source target tokens, we explicitly control this quantity during data construction. A projected bounding hull alone is insufficient for flat, hollow, or highly non-convex assets, for which the hull can substantially overestimate actual visible support. We therefore use three stages.

First, the proposal solver requires at least 35 projected hull tokens. Second, we refine this estimate using per-asset surface area to approximate projected contour support. Finally, after rendering, the actual source segmentation mask is rasterized onto the latent grid and the row is retained only if
\begin{equation}
    N_T \geq 20 .
    \label{eq:app_token_floor}
\end{equation}
\rev{In the final training set}, source targets contain a median of 89.8 tokens; the minimum is 20.2 tokens and the 90th percentile is approximately 315 tokens.

\subsection{Rendered Geometry and Materialization}
\label{app:data_materialization}

The renderer provides explicit source and destination object masks and world-space geometry. Training-time destination centers and apparent scaling are obtained directly from the rendered masks. For partially occluded destinations, the corresponding amodal render is used. Analytic projections are retained separately as consistency checks.

For each retained row we materialize:
\begin{itemize}
    \item the source RGB image and target RGB image;
    \item target and reference segmentation masks;
    \item source and destination target centers on the latent grid;
    \item measured source and destination target extents;
    \item $(c_s,c_d)$ and $a=\exp(c_s-c_d)$;
    \item camera and scene metadata; and
    \item identifiers required for exact source--target pairing.
\end{itemize}
Writing the measured mask extents as $\ell_{\mathrm{src}}^{\mathrm{render}}$ and $\ell_{\mathrm{dst}}^{\mathrm{render}}$, the training warp uses $a^{\mathrm{render}}=\ell_{\mathrm{dst}}^{\mathrm{render}}/\ell_{\mathrm{src}}^{\mathrm{render}}$ and centers $\mathbf{m}_{\mathrm{src}},\mathbf{m}_{\mathrm{dst}}$ in Eq.~\eqref{eq:app_rope_map}. This measured ratio is kept distinct from the analytic $a=\exp(c_s-c_d)$.

\subsection{Physical Certification}
\label{app:data_certification}

Every rendered row is subjected to deterministic geometric and physical checks before entering training. The certification protocol includes:

\begin{enumerate}
    \item \textbf{Command consistency.}
    Ground-truth source and destination geometry must reproduce the prescribed
    $\Delta\log S=c_s$, $\Delta\log d=c_d$, and
    $\log a=c_s-c_d$ command relations to numerical tolerance. Here $a$ is the analytic command factor; the observed rendered extent ratio is measured separately.

    \item \textbf{Source integrity.}
    Source identity and scene metadata must correspond exactly to the intended
    source render.

    \item \textbf{Token support.}
    The measured source target must satisfy
    Eq.~\eqref{eq:app_token_floor}.

    \item \textbf{Instance identity.}
    Target and reference identities must be unambiguous, with exactly one
    intended destination target instance.

    \item \textbf{Projection consistency.}
    The analytic world-to-image projector must agree with rendered object
    geometry.

    \item \textbf{Frame validity.}
    Training targets may not be truncated by the image boundary.

    \item \textbf{Physical validity.}
    Target transformations may not introduce mesh penetration or invalid
    support relationships with other objects.

    \item \textbf{Occlusion validity.}
    Occlusion strata are evaluated using amodal object support rather than
    intersection of already-occluded visible masks.

    \item \textbf{Interaction locality.}
    The radius of image changes caused by the target transformation must
    converge; examples whose lighting or indirect-interaction changes continue
    to expand across the frame are rejected.
\end{enumerate}

Of 2,500 rendered candidates, 32 fail at least one final certification criterion, leaving 2,468 training pairs. Thirteen are rejected because the measured source target falls below the 20-token floor, 13 because the influence radius does not converge, and seven because the physical transformation produces scene-wide changes outside the certified influence regions; categories can overlap for multiply failing rows.

As an independent geometry check, the renderer and analytic projector agree on world-space object AABBs to a maximum error of $2.73\times10^{-7}$ m in the final certified set. Planned versus rendered lateral displacement has a median error of approximately $0.8$ pixels and a 90th-percentile error of $6.4$ pixels. These quantities are used only for data certification; actuator training itself uses the rendered source--target pairs and the standard flow-matching loss described above.

\FloatBarrier
\section{Observer and Planner Details}
\label{app:observer-planner}
\subsection{Physical Observer}
\label{app:observer}
\subsubsection{Open-vocabulary instance binding}
The observer uses \path{IDEA-Research/grounding-dino-base}~\revcitep{liu2023groundingdino} with one joint prompt containing the queried nouns. A noun's score is the maximum sigmoid score over its token span; the box threshold is 0.15 and at most 24 proposals are retained. \arxivwording{For object mask extraction~\citep{kirillov2023sam,chen2023sampseudo}, we convert each box into a mask with \path{facebook/sam-vit-base}}{\path{facebook/sam-vit-base}~\revcitep{kirillov2023sam} converts each box into a mask}, keeping the mask with the highest predicted IoU and discarding masks smaller than 50 pixels.

Same-noun masks with IoU at least 0.70 are merged by retaining the larger mask; nested instances are not removed solely for containment. A Hungarian one-to-one assignment binds nouns to distinct primary instances. Primary scores below 0.15 remain unmatched, and extra instances attach only to nouns with a primary match. These detection-stage rules differ from the overlap rules used later for restoring missing objects. Measurement uses a 512-coordinate grid; the native image grid is 1024 and actuator patches have stride 16.

\subsubsection{Ground contact and camera geometry}
The contact point uses the mask's lowest row and horizontal box center. Its ray intersects the ground plane to give $g_i$. Camera intrinsics are fixed at $f/W=2.4740$ (89.06\,mm on a 36\,mm sensor), with camera height 0.3237 scene units; camera tilt aligns the projected horizon with the estimated horizon. Observation and planning share these camera and support conventions rather than using pose-predicted translation.

Depth and normals come from \path{Ruicheng/moge-2-vitl-normal}; ground segmentation uses \path{nvidia/segformer-b4-finetuned-ade-512-512}, selecting ADE classes from a \rev{fixed} ground-word list. For the GenScale horizon, candidate MoGe ground points lie below row $0.45H$ and outside the target and reference. When more than 500 candidates have normal component $n_y<-0.80$, only those normal-filtered points are used. Otherwise the lower-image fallback is retained. The fit regresses inverse depth on row, removes residuals above the 90th percentile, and refits; if $1/d=A_{\mathrm{inv}}y+B_{\mathrm{inv}}$, the horizon is $y_h=-B_{\mathrm{inv}}/A_{\mathrm{inv}}$. Acceptance requires at least 300 ground pixels and a finite horizon, not a threshold on fit quality.

\rev{For the rendered scenes of the controlled observer and planner evaluations (Appendices~\ref{app:exp_observer} and~\ref{app:exp_planner}), the image-estimated horizon is instead fit with} the offset-robust model
\begin{equation}
d(y)=\frac{A_h}{y-y_h}+B_h
\label{eq:app_horizon_fit}
\end{equation}
with soft-$L_1$ nonlinear least squares. This additive-depth-offset model is distinct from the GenScale inverse-depth fit.

\subsubsection{Knowledge-base dimensions and projection fitting}
\rev{The knowledge base combines a table of category aspect ratios (109 categories) with a table of typical lengths (163 categories). The length table covers all 100 GenScale categories, and for 99 of them the typical length equals the value in GenScale's size table.} Towel and watch are excluded from the observer. Canonical dimensions $\mathbf{D}_i=(L_i,W_i,H_i)$ combine the shape and length records.

All categories use the same projection measurement as Sec.~\ref{sec:observer}. Let $Q_i$ contain the eight corners of the cuboid with knowledge-base dimensions $\mathbf{D}_i=(L_i,W_i,H_i)$, $R_i$ its orientation, $g_i$ its ground position, and $\Pi$ the shared camera projector. Writing $b_i(\alpha)$ for the projected 2D box gives
\begin{align}
b_i(\alpha)&=\operatorname{BBox}\!\left(\Pi(g_i+\alpha R_iQ_i)\right),\label{eq:app_projected_box}\\
\ell_i^{\mathrm{proj}}(\alpha)&=\max\!\left(w(b_i(\alpha)),h(b_i(\alpha))\right),\label{eq:app_observer_projected_extent}\\
\ell_i&=\max\!\left(w(\operatorname{BBox}(M_i)),h(\operatorname{BBox}(M_i))\right).\label{eq:app_observer_observed_extent}
\end{align}
Thus $\operatorname{extent}(\cdot)$ in the main text is the longest side of this box. The scalar solve finds $\alpha_i$ such that $\ell_i^{\mathrm{proj}}(\alpha_i)=\ell_i$, then returns $r_i=\log\alpha_i$. Bisection starts on $\log\alpha\in[-3,3]$ and uses 50 iterations; an invalid upper projection bound is repeatedly reduced. Failure to obtain a valid interval with endpoint projections on opposite sides of the observed extent causes abstention. The camera, orientation, and ground position remain fixed during fitting; they are not assumed to be ground truth.

For the representative physical dimension used in Sec.~\ref{sec:problem_formulation}, let $S_i^{\mathrm{kb}}$ be its knowledge-base value and $\widehat S_i=\alpha_iS_i^{\mathrm{kb}}$. Equation~\eqref{eq:observer_residual} then implies
\begin{equation}
r_i-r_j=\log\frac{\widehat S_i}{\widehat S_j}-\log\frac{S_i^{\mathrm{kb}}}{S_j^{\mathrm{kb}}}.
\label{eq:app_observer_pairwise}
\end{equation}
Residual differences therefore measure deviations from knowledge-base size ratios. Equal residuals recover those ratios without requiring every object to have its category's typical absolute size.

\subsubsection{Pose routing and canonical-axis assignment}
For the canonical extents sorted as $L_{(1)}\ge L_{(2)}\ge L_{(3)}$, the routing rule is
\begin{equation}
\begin{split}
&\operatorname{family}(L_{(1)},L_{(2)},L_{(3)})=\\
&\quad\begin{cases}
\text{flat-planar},&L_{(3)}/L_{(1)}<0.30,\\
\text{elongated},&L_{(3)}/L_{(1)}\ge0.30\text{ and }\\
&L_{(2)}/L_{(1)}<0.55,\\
\text{compact},&\text{otherwise}.
\end{cases}
\end{split}
\end{equation}
Flat and elongated objects use \rev{the released UniPose9D checkpoint}; compact objects use an upright identity rotation. The family changes only the orientation supplied to the common projection solver, not its size measurement.

Predicted metric size and translation are not consumed as measurements. For predicted axis extents $\mathbf{s}=(s_x,s_y,s_z)$ and canonical dimensions $\mathbf{D}=(L,W,H)$, normalize each by its geometric mean and choose the axis permutation
\begin{equation}
\begin{gathered}
\widetilde s_k=\frac{s_k}{(s_xs_ys_z)^{1/3}},\quad
\widetilde D_k=\frac{D_k}{(LWH)^{1/3}},\\
\pi^\star=\operatorname*{arg\,min}_{\pi\in\mathcal{S}_3}\sum_{k=1}^3\left(\log\widetilde s_k-\log\widetilde D_{\pi(k)}\right)^2.
\end{gathered}
\end{equation}
Only the axis assignment and rotation reach the scale solver. This normalization removes any global multiplicative scale in the pose model's predicted dimensions. Rotations are converted from the pose model's object-to-camera convention to the projector's world basis; handedness, axis-permutation, and scale-invariance checks accompany the implementation.

\subsubsection{\rev{Abstention}}
An object abstains when its canonical entry is missing or excluded, its foot is at most eight pixels below the horizon, its ray misses the ground, a required flat/elongated pose is missing, or no valid scale bracket exists. Unresolved instance binding prevents a corresponding observation. An unobservable target or reference voids the source; an unobservable bystander is omitted. The observer \rev{has no task-specific training} and receives no actuator-training gradient.

\subsection{Size and Placement Planner Details}
\label{app:planner}
\subsubsection{Nominal candidate geometry}
\label{app:planner_geometry}
A candidate scene offset assigns nominal size commands
\begin{equation}
c_{s,i}=\mu-r_i.
\label{eq:app_gauge}
\end{equation}
The remaining realization variables are $\mathbf{y}_i=(c_{d,i},\Delta x_i)$. The analytic projector predicts the destination footprint
\begin{equation}
M_i^{\mathrm{dst}}=F_i(M_i,c_{s,i},c_{d,i},\Delta x_i),\qquad a_i=\exp(c_{s,i}-c_{d,i}).
\label{eq:app_forward_footprint}
\end{equation}
\rev{The command dead zone is 0.26 in log-size units: smaller size commands are set to zero.} The exact common-offset relation defines the nominal allocation; the later pair-safe and minimum-edit stages described below can relax equality while \rev{satisfying the pairwise tolerance conditions}.

\subsubsection{Prioritized costs}
\label{app:planner_costs}
The planner compares a finite set of candidates in priority order, rather than solving a continuous constrained optimization to global optimality. Support and collision precede image-space feasibility, which precedes the editor's apparent-scaling range. The image-change and motion costs are considered only after these constraint tiers tie. The same ordering is used for individual placements and the completed scene.

Candidates are ranked by non-negative violations. The physical tier measures invalid support or penetration. The image-space tier \rev{($\mathrm{img}$)} combines overlap/depth ordering with frame truncation and insufficient target-token support. The range tier penalizes excessive apparent warp. Each tier is quantized by
\begin{equation}
\bar V_i^k=\left\lfloor V_i^k/0.02\right\rfloor,\qquad k\in\{\mathrm{phys},\mathrm{img},\mathrm{range}\}.
\label{eq:app_quantized_violation}
\end{equation}
The overlap dead zone is 0.15. In front of another object the overlap term scales its covered fraction by 0.6; behind another object it measures the target's covered fraction. Covering the reference doubles the term. Framing takes the maximum of the out-of-frame fraction and starvation, where starvation is $(16-N)/16$ when fewer than 16 destination-crop tokens are available. The range penalty is zero for $|c_{s,i}-c_{d,i}|\le1.2$ and otherwise increases linearly, capped at one.

Once higher-priority violations are indistinguishable, the planner favors a smaller image change. On the 512 grid it uses
\begin{equation}
\begin{aligned}
C_i^{\mathrm{change}}&=\frac{|M_i\triangle M_i^{\mathrm{dst}}|+A_i^{\mathrm{contact}}}{512^2},\\
A_i^{\mathrm{contact}}&=2n_i^{\mathrm{contact}}\cdot10.
\end{aligned}
\label{eq:app_pixel_change}
\end{equation}
Effort averages clipped depth and lateral deviations, normalized by 0.6 and 256 pixels, respectively. For fixed size commands, individual placements are compared by the ordered tuple $(\bar V_i^{\mathrm{phys}},\bar V_i^{\mathrm{img}},\bar V_i^{\mathrm{range}},C_i^{\mathrm{change}}/0.10,C_i^{\mathrm{motion}})$. The scene key aggregates worst-object violations and total costs:
\begin{equation}
\begin{split}
\mathbf{J}_{\mathrm{scene}}=\bigl(&\max_i\bar V_i^{\mathrm{phys}},\ \max_i\bar V_i^{\mathrm{img}},\ \max_i\bar V_i^{\mathrm{range}},\\
&\textstyle\sum_i C_i^{\mathrm{change}}/0.10,\ \sum_i C_i^{\mathrm{motion}},\ N_{\mathrm{edit}}\bigr).
\end{split}
\label{eq:app_scene_key}
\end{equation}
We compare these keys lexicographically: at the first component that differs, the smaller value wins. Thus lower image-change or motion cost cannot compensate for a worse higher-priority violation. The later tolerance refinements need not preserve exact common-offset equality.

\subsubsection{Multi-object search and \rev{command} refinements}
\label{app:planner_search}
For each offset hypothesis, objects are processed greedily from largest source area to smallest. Earlier placements become obstacles for later ones. The coarse pass uses three depth values and nine lateral positions, with a 64-pixel lateral step. The fine base grid uses 25 depth values and 65 lateral positions, with an eight-pixel lateral step over $\pm256$ pixels:
\begin{equation}
K_{\mathrm{coarse}}=3\times9=27,\qquad K_{\mathrm{fine}}=25\times65=1625.
\end{equation}
Depth commands lie in $[-0.6,0.6]$, restricted to the nominal direction; the nominal depth and parent lateral position are also included. The common-offset stage considers 21 grid points over $[\max_i r_i-\Lambda,\min_i r_i+\Lambda]$, with $\Lambda=1.2$, plus each object's own residual. The chosen offset is replanned on the fine grid, then propagated to additional instances. Greedy placement and coarse offset selection are approximations rather than global joint optimization. \revb{Candidates whose target would lie entirely outside the frame are skipped; in Fig.~\ref{fig:method_example}(b), 1,440 of the 1,625 fine-grid candidates remain.}

\rev{When a plan is executed, each size and depth command is multiplied by a fixed execution factor $\eta=1.08$ before it is passed to the actuator; lateral offsets are not scaled. This factor was fixed before the GenScale evaluation.} \rev{Two further refinements follow.} Pair-safe re-selection accepts a different shared offset only if no pair's excess outside $b=\log1.2$ increases, the mean excess strictly decreases, and the first three scene-key tiers are no worse than the base plan. Pair errors are computed with \rev{the executed size commands (Eq.~\eqref{eq:app_pair_excess} below Algorithm~\ref{alg:planner_main})}. The accepted size commands are then held fixed for fine-grid placement and propagated to additional instances of the same object category. The base plan is retained when no alternative passes these conditions.

A subsequent minimum-edit linear program reduces the sum of absolute size commands while preserving their signs, not exceeding their parent magnitudes, and not worsening any parent's pairwise band excess. Active magnitudes remain between 0.26 and 1.2; unchanged objects remain unchanged. These refinements, the command dead zone, and the execution \rev{factor} distinguish the delivered commands from the exact common-offset allocation in Eq.~\eqref{eq:gauge_command}. In particular, final residuals need not all be equal.

\subsubsection{Complexity and exact lazy token checks}
\label{app:planner_complexity}
For $N$ objects and $n_\mu$ offset hypotheses, the base search evaluates
\begin{equation}
O(n_\mu N K_{\mathrm{coarse}}+N K_{\mathrm{fine}})
\label{eq:planner_complexity}
\end{equation}
candidates, ignoring fixed-resolution raster work. Accounting for interaction with all other objects gives the conservative bound $O(n_\mu N^2K_{\mathrm{coarse}}+N^2K_{\mathrm{fine}})$. Bounding-box rejection avoids many pairwise tests, and lateral candidates at a fixed depth can reuse their projected footprint.

Crop-token evaluation is more expensive than the remaining key terms and can only worsen a candidate's key. The unchecked key is therefore an optimistic lower bound. The accelerated implementation sorts candidates by this key and a deterministic index, checks them in order, and stops once the next optimistic key is strictly worse than the best fully checked key. Ties are still inspected under the same index rule. This preserves exhaustive-search selection rather than approximating token support.

\subsubsection{\rev{Planner configuration for the placement stress tests}}
\label{app:planner_abstention}
\rev{The planner stress tests (Appendix~\ref{app:exp_planner}) pose a placement task with explicit requirements. The target's physical size change is prescribed, and a planner chooses only the target's depth command and lateral offset. After the change, the target's longest side must be 128--224 pixels; the target must lie fully in frame, at least 12 pixels from every image border; its entire base must rest on the support surface; at least 90\% of it must remain visible; and every neighbor must remain at least 90\% visible and lose at most 5\% of its visible area. Gemini's prompt states these requirements with the same numbers, together with the same observations, forward formulas, command bounds (depth in $[-1.9,1.9]$, lateral offset within $\pm256$ pixels, no apparent-scale limit), and acceptance tiers, so both planners solve the same specified task.}

\rev{Our planner encodes these requirements in its search, and because stress-test plans are executed in Blender rather than by the diffusion editor, it searches a wider depth range. Since the size change is prescribed and used as given, the shared-scale search, size-command dead zone and limit, and command refinements of Algorithm~\ref{alg:planner_main} do not apply; the single target is placed directly on a $25\times65$ depth--lateral grid. The candidate geometry and cost terms are those of Appendix~\ref{app:planner_costs}, with the following differences from the configuration used for GenScale correction:}
\begin{itemize}[leftmargin=1.2em,itemsep=1pt,topsep=2pt]
\item \rev{\textbf{Depth range.} Depth commands span $[-1.9,1.9]$ (25 values) instead of $[-0.6,0.6]$; the lateral grid ($\pm256$ pixels, 8-pixel step) is unchanged.}
\item \rev{\textbf{Editor range.} The apparent-scaling range penalty is not applied, because Blender renders any apparent scale; for GenScale correction it is zero up to $|c_s-c_d|=1.2$ and grows linearly beyond.}
\item \rev{\textbf{Requirement terms in the ranking.} Four terms are added: side overhang (the predicted target box at its contact row extends past the left or right edge of the support), rear overhang (the base, with depth estimated as half its width, extends past the far edge of the support), an image margin of 16 pixels, and a preferred longest-side band of 140--212 pixels, which lies inside the required 128--224 pixels. Candidates are compared lexicographically by side overhang, physical validity, overlap/framing/token support, rear overhang, margin, size band, and then the image-change and motion costs.}
\item \rev{\textbf{Overlap dead zone.} 0.03 instead of 0.15.}
\item \rev{\textbf{Small-target penalty.} Target tokens are counted over the full frame (one token per $16\times16$ patch at native resolution) instead of in the editing crop; the 16-token threshold is unchanged.}
\item \rev{\textbf{Support observation.} Support-surface columns hidden behind the target are filled by interpolating the far edge from the visible columns on either side, and the far edge is relaxed by 12 pixels when testing whether the contact point lies on the support.}
\item \rev{\textbf{Acceptance tiers.} The search is first restricted to candidates whose predicted longest side lies within the required 128--224 pixels and that stay fully in frame; if none exists, the restriction is relaxed to 96--288 pixels with at most 5\% of the target outside the frame, and then removed. Gemini receives the same three tiers.}
\end{itemize}
\rev{Both planners use the same image-estimated horizon, fit with Eq.~\eqref{eq:app_horizon_fit}. Appendix~\ref{app:exp_planner} reports the stress-test results.}

\subsubsection{Planner Pseudocode}
\label{app:planner_pseudocode}
Algorithm~\ref{alg:planner_main} summarizes shared-scale search and refinement in \rev{our planner}. Algorithm~\ref{alg:place_objects} places objects for fixed size commands using the geometry and keys in Eqs.~\eqref{eq:app_forward_footprint}--\eqref{eq:app_scene_key}. Pair-safe re-selection can change the nominal allocation, and minimum-edit refinement can shorten its commands; neither requires the final residuals to share one exact value. Grid construction and lazy token checking are specified in Sections~\ref{app:planner_search} and~\ref{app:planner_complexity}. The stress-test configuration is described separately in Section~\ref{app:planner_abstention}.

\begin{algorithm}[!htbp]
\caption{Shared-scale planning and command refinement}
\label{alg:planner_main}
\small
\begin{algorithmic}[1]
\Require Observed scene $\mathcal S$ with primary instances, masks $M_i$, residuals $r_i$, camera and support geometry
\Ensure Size--depth commands and lateral placements for the edited instances
\State Set $\Lambda=1.2$, command dead zone $\tau=0.26$, pair band $b=\log1.2$, and \rev{execution factor} $\eta=1.08$.
\State $\mathcal U\gets\operatorname{linspace}(\max_i r_i-\Lambda,\min_i r_i+\Lambda,21)\cup\{r_i\}_i$.
\State Initialize the candidate-plan set $\mathcal C\gets\varnothing$.
\For{$\mu\in\mathcal U$}
  \State Assign nominal size commands $c_{s,i}\gets\mu-r_i$ \hfill (Eq.~\eqref{eq:gauge_command}).
  \State Set commands inside the dead zone $\tau$ to zero.
  \State $(P_\mu,\mathbf J_\mu)\gets\Call{PlaceObjects}{\mathcal S,\{c_{s,i}\},\mathcal G_{\mathrm{coarse}}}$; add $(P_\mu,\mathbf J_\mu)$ to $\mathcal C$.
\EndFor
\State Choose base plan $P_0$ with lexicographically smallest $\mathbf J_{\mathrm{scene}}$ in $\mathcal C$.
\State Retain alternatives whose constraint-tier key is no worse than $P_0$ and whose $e_{jk}$ never increases.
\State Pair-safe re-select only if mean $e_{jk}$ strictly decreases; otherwise retain $P_0$.
\State Hold the accepted size commands fixed and replan with $\mathcal G_{\mathrm{fine}}$.
\State Propagate the selected correction to additional instances of the same category.
\State Apply minimum-edit LP refinement under the sign, magnitude, and pair-band restrictions in Section~\ref{app:planner_search}.
\State \Return the refined commands and placements.
\end{algorithmic}
\medskip
\noindent\footnotesize For a plan $P$, the predicted excess of pair $(j,k)$ \rev{under the executed commands $\eta c_{s}$} is
\begin{equation}
\begin{split}
e_{jk}(P)=\max\!\bigl\{0,&\bigl|(r_j+\eta c_{s,j}(P))\\
&-(r_k+\eta c_{s,k}(P))\bigr|-\log1.2\bigr\}.
\end{split}
\label{eq:app_pair_excess}
\end{equation}
\end{algorithm}

\begin{algorithm}[!htbp]
\caption{\textsc{PlaceObjects}: greedy placement with fixed size commands}
\label{alg:place_objects}
\small
\begin{algorithmic}[1]
\Require Scene $\mathcal S$, fixed size commands $\{c_{s,i}\}$, and a depth--lateral grid $\mathcal G$
\Ensure Candidate plan $P$ and its scene key $\mathbf J_{\mathrm{scene}}$
\State Sort edited primary instances by decreasing source-mask area $|M_i|$.
\State Initialize $P$ with unchanged instances and the shared support geometry.
\For{each edited instance $i$ in this order}
  \State Initialize its selected placement as unset.
  \For{each allowed $(c_{d,i},\Delta x_i)\in\mathcal G_i$}
    \State Project its destination mask and apparent scale using Eq.~\eqref{eq:app_forward_footprint}.
    \State Measure support and collision violations against the current scene.
    \State Measure overlap/depth-order, frame, and target-token violations.
    \State Measure the apparent-scaling-range violation.
    \State Quantize the three tiers using Eq.~\eqref{eq:app_quantized_violation}.
    \State Compute image-change and motion costs as in Section~\ref{app:planner_costs}.
    \State Retain the candidate with the lexicographically smallest placement key.
  \EndFor
  \State Commit its placement to $P$; it becomes an obstacle for subsequent objects.
\EndFor
\State Aggregate the worst-object tiers and total costs into $\mathbf J_{\mathrm{scene}}$ (Eq.~\eqref{eq:app_scene_key}).
\State \Return $(P,\mathbf J_{\mathrm{scene}})$.
\end{algorithmic}
\medskip
\noindent\footnotesize The base grids contain $3\times9$ coarse and $25\times65$ fine depth--lateral samples before the direction restrictions and extra candidates in Section~\ref{app:planner_search}. Exact lazy token checking (Section~\ref{app:planner_complexity}) avoids unnecessary crop evaluations without changing candidate selection.
\end{algorithm}

\FloatBarrier
\section{Experimental Protocols and Additional Results}
\label{app:experiments}
\FloatBarrier
\subsection{Implementation and Statistical Setup}
\label{app:exp_setup}
The observer and planner configurations are specified in Appendix~\ref{app:observer-planner}, actuator architecture and optimization in Appendix~\ref{app:actuator_impl}, and 3D training pairs in Appendix~\ref{app:actuator_data}. Only the actuator receives task-specific training. Gemini-3.1-Pro-preview~\revcitep{google2026gemini31pro} serves as the comparison observer or planner and as an offline judge, not as a correction-decision component of \MethodName{}. \rev{Each of the following subsections states its own evaluation population and configuration.} \revb{Appendix~\ref{app:rev} reports further evaluations, including the GenScale human study (Appendix~\ref{app:rev_judges}) and matched retraining (Appendix~\ref{app:rev_retrain}). Throughout, \emph{editor} and \emph{actuator} denote the same component, a \emph{source} is one generated image, and Gemini~3.1~Pro denotes the Gemini-3.1-Pro-preview model.}

\subsubsection{Statistics}
We report 95\% bootstrap confidence intervals (CIs) using each experiment's analysis unit: source images for source-level comparisons, scene groups for ObjectMover, and product families for planner stress tests. Most experiments use 10,000 draws. \revb{For GenScale, generator-specific intervals use 4,000 draws that resample sources within each generator; the overall interval and the intervals of Appendix~\ref{app:rev} resample prompts, so that images sharing a prompt stay together.} Observer size-identification comparisons use paired scene resampling. Holm adjustments are applied \rev{within the contrast family specified for each experiment}. Human ratings, offline model ratings, and geometric gate outcomes are reported separately. Main-text tables show point estimates only; bold and underline mark the best and second-best values \revb{among the compared methods (the retrained controls of Table~\ref{tab:exp_actuator} are not ranked)}, not statistical significance.

\FloatBarrier
\subsection{GenScale Scoring}
\label{app:exp_genscale}
\begin{figure*}[p]
\section{Additional Visualizations}
\label{app:additional_visualizations}
\begin{multicols}{2}
The following panels expand the main-text visual comparisons; aggregate results are in Appendix~\ref{app:experiments}.
\end{multicols}

\subsection{GenScale Relative-Size Correction}
\label{app:qual_genscale}
\begin{multicols}{2}
Figure~\ref{fig:app_genscale_more} expands Fig.~\ref{fig:exp_genscale_qual} with a wider range of object pairs and size errors. The examples include undersized and oversized objects, as well as scenes with more than one incorrectly sized object. Comparing each result with its source makes both the relative-size change and preservation of the surrounding scene visible.
\end{multicols}
\centering
\captionsetup{font=small,skip=4pt,singlelinecheck=false}
\includegraphics[width=\linewidth,height=0.78\textheight,keepaspectratio]{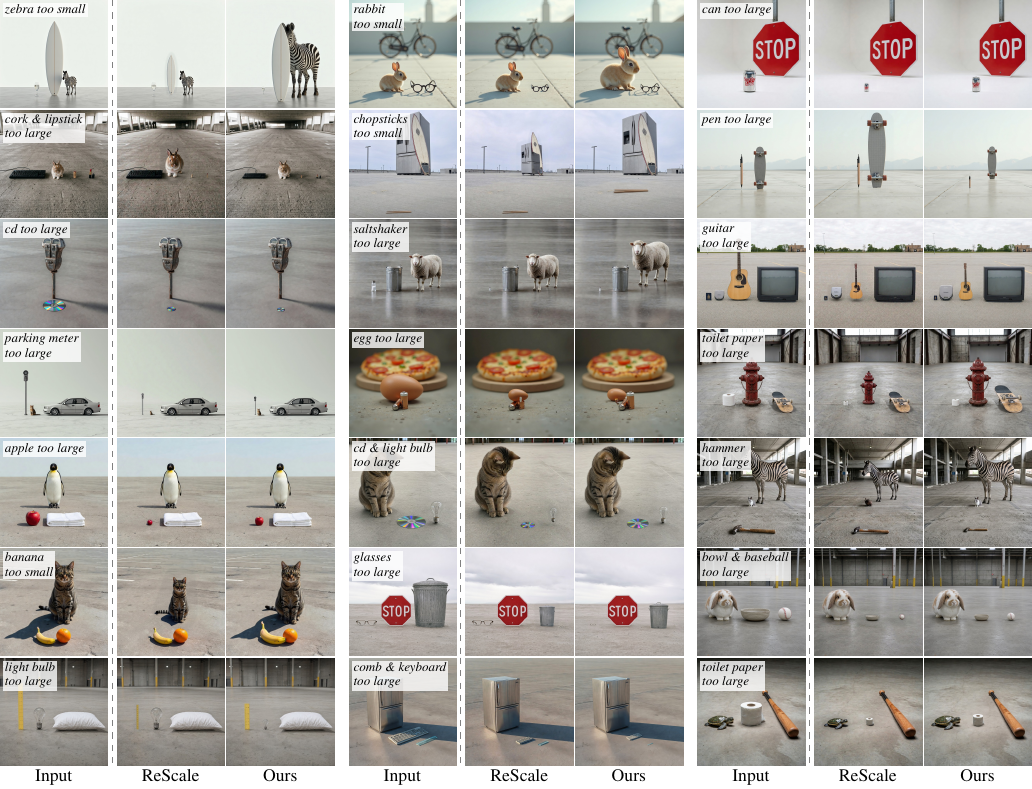}
\caption{\textbf{Additional GenScale correction examples.} Twenty-one examples are arranged in three groups of input, ReScale, and \MethodName{} columns. Annotations on the inputs identify the original relative-size errors. Each triplet compares corrections of the same source; the aggregate comparison and scoring \rev{details} are reported in Table~\ref{tab:exp_genscale} and Appendix~\ref{app:exp_genscale}.}
\label{fig:app_genscale_more}
\end{figure*}

GenScale Task 1 contains 400 sources from each of seven generators and 8,001 designed object relations. The evaluated generators are Nano Banana 2~\citep{google2026nanobanana2}, GPT-Image-2~\citep{openai2026gptimage2}, Z-Image-Turbo~\citep{zimage2025}, Grok Imagine~\citep{xai2026grokimagine}, Qwen-Image 2512~\citep{qwen2025image2512}, FLUX.2~\citep{blackforestlabs2025flux2}, and SD3.5-Large~\citep{stabilityai2024sd35}. The offline judge is Gemini-3.1-Pro-preview, with temperature 0.35 for size scoring and 0.1 for scene prefiltering. Five votes are aggregated by a unique mode or, when needed, a rounded median. The human-calibrated rubric assigns score three to the valid size band, with relation error $|s-3|$. \rev{The main protocol keeps all designed relations and assigns the maximum error of two to any relation that the judge or its scene prefilter cannot score, for example because an object is missing or no longer recognizable: an edit that removes or destroys an object has failed to correct its size.} \revb{Such relations number 549 in the source images, 663 after \MethodName{}, and 837 after ReScale.} \revb{Under the original protocol, which omits these relations, the errors are 0.829 (source), 0.491 (ReScale), and 0.517 (\MethodName{}); the two methods again differ within sampling uncertainty (difference $+0.026$, 95\% CI $[-0.010,+0.063]$).} Additional visual comparisons with ReScale appear in Fig.~\ref{fig:app_genscale_more}.

Table~\ref{tab:app_genscale_intervals} gives generator-specific paired differences\rev{; over all seven generators, ReScale minus \MethodName{} is $+0.009$ with 95\% CI $[-0.030,+0.048]$ (bootstrap clustered by prompt). Appendices~\ref{app:rev_judges} and~\ref{app:rev_heldout} report the comparison under two independent judges and on sources not used during development.}

\begin{table}[!htbp]
\captionsetup{font=small,skip=4pt}
\centering
\small
\setlength{\tabcolsep}{5pt}
\caption{\textbf{Generator-level paired differences.} ReScale error minus \MethodName{} error under the main \rev{protocol}. Per-generator intervals use 4,000 source-bootstrap resamples within each generator; \rev{the aggregate uses a bootstrap clustered by prompt.}}
\label{tab:app_genscale_intervals}
\resizebox{0.9\linewidth}{!}{%
\begin{tabular}{@{}lrr@{}}
\toprule
Generator & Difference & 95\% CI \\
\midrule
Nano Banana 2 & $-0.013123$ & $[-0.091384, 0.065094]$ \\
GPT-Image-2 & $+0.066492$ & $[-0.012142, 0.142364]$ \\
Z-Image-Turbo & $+0.009624$ & $[-0.072899, 0.090592]$ \\
Grok Imagine & $-0.021872$ & $[-0.103791, 0.058618]$ \\
Qwen-Image 2512 & $-0.083115$ & $[-0.166515, 0.000850]$ \\
FLUX.2 & $+0.007874$ & $[-0.078179, 0.096507]$ \\
SD3.5-Large & $+0.096238$ & $[-0.008475, 0.196850]$ \\
\midrule
All seven & \rev{$+0.009$} & \rev{$[-0.030, +0.048]$} \\
\bottomrule
\end{tabular}
}
\end{table}

\FloatBarrier
\subsection{Front-End--Actuator Swap}
\label{app:exp_swap}
\rev{This subsection shows that our actuator improves correction with either front end, and that our geometric front end delivers correction close to that of a Gemini front end at about one ninth of the cost.} The factorial comparison uses 196 GenScale sources with 679 relations\revb{: 28 prompts selected pseudo-randomly by a fixed hash order, each with all seven generators. Of these sources, 187 are outside the 230 used to tune planner thresholds (Appendix~\ref{app:rev_heldout})}. Front ends are the \MethodName{} observer/planner and ReScale's Gemini front end; back ends are the \MethodName{} actuator and ReScale's actuator. Each judge provides one vote per relation, with missing objects assigned error two. \rev{Besides the main Gemini judge, we re-score all four compositions with GPT-6.1 Terra~\revcitep{openai2026gpt61terra} and Claude Sonnet~5.5~\revcitep{anthropic2026sonnet55} (Appendix~\ref{app:rev_judges}; Table~\ref{tab:app_swap_contrasts}).}

\rev{\textbf{Actuator.} Replacing ReScale's actuator with ours lowers the error with either front end. With our front end, the reduction is 0.171 (95\% CI $[0.074,0.270]$, Holm-adjusted $p=0.0035$) under Gemini and 0.127 (95\% CI $[0.040,0.217]$) under Claude; with the Gemini front end, it is 0.137 (95\% CI $[0.035,0.256]$) under Gemini\revb{, paired over the 673 relations scored in both compositions}.}

\rev{\textbf{Front end.} Under the Gemini judge, the Gemini front end has a lower error (0.570 vs.\ 0.654), but that front end takes its diagnoses from the Gemini judge's own scores of the source image (Appendix~\ref{app:rev_coupling}). Under GPT, the two front ends paired with our actuator are tied (0.713 vs.\ 0.710), and under Claude the gap shrinks to 0.066. Our complete system reduces the source error by 0.21, 0.22, and 0.19 under Gemini, GPT, and Claude, with every 95\% CI excluding zero. Our observer is also more accurate on the controlled diagnosis of Table~\ref{tab:exp_observer} and calls no VLM.}

\begin{table}[!htbp]
\centering\small
\captionsetup{font=small,skip=4pt}
\setlength{\tabcolsep}{4pt}
\caption{\rev{\textbf{Component swap under three judges.} Relation error ($\downarrow$) on 196 sources and 679 relations, one vote per relation. P: \MethodName{}; G: Gemini front end; R: ReScale actuator.}}
\label{tab:app_swap_contrasts}
\rev{\resizebox{0.9\linewidth}{!}{%
\begin{tabular}{@{}llrrr@{}}
\toprule
Front end & Actuator & Gemini & \makecell{GPT-6.1\\Terra} & \makecell{Claude\\Sonnet 5.5} \\
\midrule
\multicolumn{2}{@{}l}{Source image} & 0.865 & 0.925 & 0.857 \\
P & P & 0.654 & \textbf{0.710} & 0.669 \\
G & P & \textbf{0.570} & 0.713 & \textbf{0.602} \\
P & R & 0.826 & 0.801 & 0.795 \\
G & R & 0.701 & 0.799 & 0.704 \\
\bottomrule
\end{tabular}
}}
\end{table}

\paragraph{Quality--cost trade-off.}
To place these configurations on a common estimated-cost scale, we use stage unit costs from the 30-source timing study (Appendix~\ref{app:exp_efficiency}) and each front end's actual edit count on the 196-source subset. We separate Gemini planning calls from execution calls such as background cleanup. The \rev{estimated front-end costs} are \$0.454 per source for Gemini and \$0.006 for \MethodName{}; per-edit estimates are \$0.217 for ReScale and \$0.0356 for our actuator, including restoration. The front ends issue 1.28 and 1.42 edits per source, respectively. We estimate total cost as front-end cost plus edits per source times actuator cost per edit. Table~\ref{tab:app_swap_cost} lists the totals; independently rounded entries need not sum exactly.

\begin{table*}[!t]
\centering\small
\captionsetup{font=small,skip=4pt,singlelinecheck=false}
\setlength{\tabcolsep}{4pt}
\caption{\textbf{Quality and estimated cost of the four compositions.} P: \MethodName{}; G: Gemini front end; R: ReScale actuator. Costs are projected to the 196-source swap population from measured stage unit costs, not invoices or a new end-to-end timing run. VLM decisions refers to the correction front end, not background-cleanup calls.}
\label{tab:app_swap_cost}
\resizebox{0.59\linewidth}{!}{%
\begin{tabular}{@{}lcrrrrc@{}}
\toprule
Front end & Actuator & Error $\downarrow$ & Front (\$) & Edits (\$) & Total (\$) & VLM decisions \\
\midrule
P & P & 0.654 & 0.006 & 0.051 & 0.057 & No \\
G & P & 0.570 & 0.454 & 0.046 & 0.499 & Yes \\
P & R & 0.826 & 0.006 & 0.308 & 0.314 & No \\
G & R & 0.701 & 0.454 & 0.277 & 0.731 & Yes \\
\bottomrule
\end{tabular}
}
\end{table*}

\rev{The complete \MethodName{} system costs about one ninth of the Gemini-front-end composition (\$0.057 vs.\ \$0.499 per source). These projected costs compare front-end and editor combinations on the 196-source swap subset; the one-twentieth cost ratio in the main text instead compares the complete \MethodName{} and ReScale pipelines measured on 30 shared sources (\$0.037 vs.\ \$0.696 per image; Appendix~\ref{app:exp_efficiency}).}

\FloatBarrier
\subsection{Observer Evaluation}
\label{app:exp_observer}
\arxivwording{We evaluate visual spatial reasoning~\citep{li2026visreason,wang2025site,tan2026camera} in scenes with three anonymously labeled objects, exactly one at an incorrect physical size.}{Each scene contains three anonymously labeled objects, exactly one at an incorrect physical size.} Methods receive the source image, object boxes, and correct mesh dimensions\rev{; our observer uses its own image-estimated horizon (Appendix~\ref{app:observer}), whereas Gemini and the reconstruction variant infer scene geometry on their own}. The observer identifies the most mutually consistent object pair and infers the remaining object's correction; pose-sensitive objects use predicted rather than true rotations. Success requires the correct object and a linear correction-factor error no greater than 20\%. Gemini-3.1-Pro-preview uses temperature 0.1 and structured output. The comparison uses the 116 Blender scenes and 109 accepted photorealistic counterparts described in Section~\ref{app:exp_planner}.

\subsubsection{\rev{Reconstruction-Based Observer Variant}}
\label{app:exp_observer_reconstruction}
\rev{\MethodName{} (reconstruction) keeps our observer's decision rule but replaces the cuboid projection fit \arxivwording{with single-view 3D reconstruction~\citep{wang2024slice3d}. SAM 3D Objects}{with 3D reconstruction. SAM 3D Objects}~\citep{sam3dteam2025sam3d} reconstructs each object's mesh and orientation from the image and the object mask, MoGe-3~\citep{kong2026moge3} estimates the camera and scene depth, and fitting each reconstructed mesh into this estimated scene geometry gives the object's length; the variant uses neither our horizon estimate nor our pose estimator.} Table~\ref{tab:app_observer_reconstruction} compares this \rev{variant} with our observer and Gemini under the same evaluation protocol. \rev{Our observer is the most accurate on Blender scenes and runs about 500 times faster than the reconstruction variant (0.34\,s vs.\ 176\,s per image) at a small fraction of its cost. Times are medians excluding model loading; our observer runs on an NVIDIA L40S GPU, while the variant runs MoGe-3 on an L40S, SAM 3D on an H200, and the mesh fit on CPU.}

\begin{table}[!htbp]
\centering\small
\captionsetup{font=small,skip=4pt}
\caption{\textbf{\rev{Observers on the controlled evaluation sets.}} Accuracy (\%) uses the same wrong-object and correction-factor criterion as Table~\ref{tab:exp_observer}. Time and estimated cost are \rev{median} per-image observer measurements, not pipeline totals.}
\label{tab:app_observer_reconstruction}
\setlength{\tabcolsep}{5pt}
\resizebox{0.95\linewidth}{!}{
\begin{tabular}{@{}lrrrr@{}}
\toprule
Observer & Blender & Photorealistic & Time (s) & Cost (\$) \\
\midrule
Gemini 3.1 Pro & 44.8 & 38.5 & 58.5 & 0.094 \\
\rev{\MethodName{} (reconstruction)} & 45.7 & \textbf{47.7} & 176 & 0.045 \\
\MethodName{} & \textbf{56.9} & 44.0 & \textbf{0.34} & \textbf{0.0002} \\
\bottomrule
\end{tabular}
}
\end{table}

\FloatBarrier
\subsection{Planner Stress Tests}
\label{app:exp_planner}
\subsubsection{Construction and Automatic Evaluation}
\label{app:stress_construction}
\rev{These stress tests isolate placement: they contain scenes where correcting size in place necessarily fails, so only a planner that also moves the object can succeed.} Each scene contains three GSO~\citep{downs2022gso}/YCB~\citep{calli2015ycb} objects on a tabletop and a locked correction factor for one target. The planner may change only its depth and lateral position. Three rendered counterexamples define the tasks: enlarging by $4\times$ truncates at least 25\% of the target with at least 48 pixels beyond the frame; enlarging by $3\times$ reduces a neighbor's visibility by at least 0.35; shrinking to $1/6$ leaves a longest side of at most 28 pixels and area at most 0.3\% of the frame. Tight-table variants add only three centimeters around the union of the relevant object footprints. Scenes are deduplicated by geometry and capped by asset family within task type. The final set contains 116 scenes from 61 target families and 82 distinct assets: 59 normal-table and 57 tight-table scenes. The permitted camera field of view is 45--60 degrees; the retained horizontal fields of view span $45.4^\circ$--$55.8^\circ$ (median $46.9^\circ$). Table~\ref{tab:stress_composition} gives the six task--table combinations and accepted photorealistic counts.

\begin{table*}[!t]
\centering
\small
\captionsetup{font=small,skip=4pt}
\caption{\textbf{Planner stress-test composition.} Counts of Blender scenes by failure mode of in-place correction and table variant, and of accepted photorealistic counterparts. The prescribed factor is the linear physical-size correction of the target.}
\label{tab:stress_composition}
\resizebox{0.58\linewidth}{!}{%
\begin{tabular}{@{}llrrrr@{}}
\toprule
Failure of in-place fix & Factor & Normal & Tight & Blender & Photorealistic \\
\midrule
Truncation & $\times4$ & 20 & 20 & 40 & 35 \\
Occlusion & $\times3$ & 20 & 20 & 40 & 39 \\
Small target & $\times1/6$ & 19 & 17 & 36 & 35 \\
\midrule
Total & & 59 & 57 & 116 & 109 \\
\bottomrule
\end{tabular}
}
\end{table*}

Both planners receive the same image, annotations, image-estimated horizon, \rev{support mask (given to Gemini as a $64\times64$ grid), forward formulas, numeric placement requirements, and acceptance tiers; Appendix~\ref{app:planner_abstention} lists the requirements and our planner's configuration for this task.} Gemini-3.1-Pro-preview uses temperature 0.1, JSON output, and a 16,384-token output limit. Both plans are executed by Blender Cycles on CPU with 64 samples, no denoising, six maximum bounces, four transparent bounces, and $512\times512$ output. The renderer preserves target orientation and non-target geometry. Amodal masks come from alpha, visible masks from object-ID renders, and full-frame membership from mesh projection through an overscan camera.

\begin{table}[!htbp]
\captionsetup{font=small,skip=4pt}
\caption{\textbf{\rev{Blender layout checks.}} \rev{The pass rate requires all checks above the rule; full tabletop support is reported as an additional check.} \revb{Both planners are told the target's longest-side, boundary, and visibility requirements, the neighbor visibility requirements, and the support requirement; the remaining checks are not stated to either planner, and no scene fails on them alone.}}
\label{tab:app_planner_gates}
\centering\small
\setlength{\tabcolsep}{4pt}
\resizebox{0.9\linewidth}{!}{%
\begin{tabular}{@{}p{0.26\linewidth}p{0.33\linewidth}p{0.33\linewidth}@{}}
\toprule
Gate & Target & Each neighbor \\
\midrule
Longest side & 128--224 pixels & 96--192 pixels \\
Short side / area & At least 48 pixels / 1.5\% of frame & Same \\
Image boundary & Fully in frame; margin at least 12 pixels & Margin at least 12 pixels \\
Visibility & At least 0.90 & At least 0.90; visible fraction drops by at most 0.05 \\
Execution & Exact prescribed physical scale; orientation unchanged & Unmoved \\
\midrule
Additional support & Entire footprint on the table & --- \\
\bottomrule
\end{tabular}
}
\end{table}

\begin{figure*}[!t]
\centering
\includegraphics[width=\linewidth]{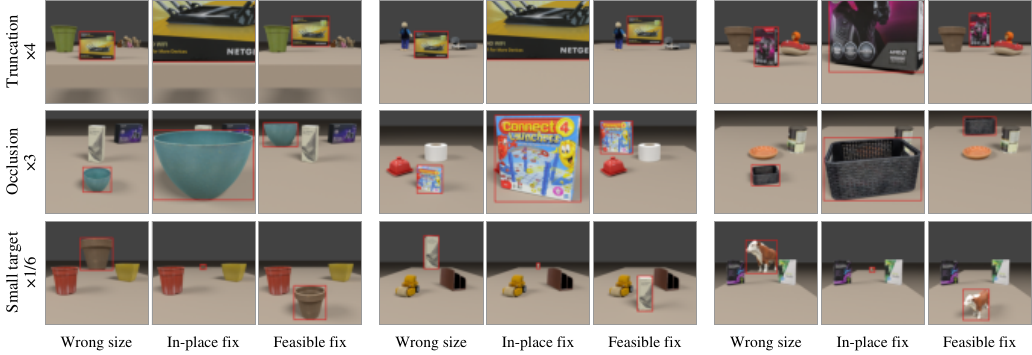}
\captionsetup{font=small,skip=4pt}
\caption{\textbf{How the stress tasks are constructed.} Three examples per failure mode; red boxes mark targets. Each triplet shows the wrong-size input, the prescribed correction applied in place, and the stored feasible reference. In-place correction truncates an enlarged target ($\times4$), hides a neighbor ($\times3$), or leaves a shrunken target too small ($\times1/6$). The reference uses the same physical-size change with adjusted placement and is not a planner input.}
\label{fig:stress_tasks}
\end{figure*}

Every scene also stores a feasible reference correction combining the prescribed physical-size change with a different depth and lateral position that satisfies the layout checks in Table~\ref{tab:app_planner_gates}. This reference demonstrates a valid solution to the constructed task but is never supplied to either planner. Figure~\ref{fig:stress_tasks} shows nine examples, three per failure mode.

\subsubsection{Subgroups}
\label{app:stress_development}
\rev{On the 116 Blender scenes, \MethodName{} passes 87.9\% versus 80.2\% for Gemini.} The advantage is concentrated in occlusion scenes: 35/40 pass for \MethodName{} and 26/40 for Gemini, a difference of 22.5 percentage points (95\% CI $[8.5,38.9]$). Both methods pass 34/40 truncation scenes and 33/36 small-target scenes. Results are 53/59 versus 44/59 on normal tables and 49/57 for both methods on tight tables. Adding full tabletop support as an extra gate yields 88/116 versus 80/116.

\subsubsection{Photorealistic Transfer}
\label{app:stress_photorealistic}
\begin{figure*}[!t]
\centering
\includegraphics[width=\linewidth]{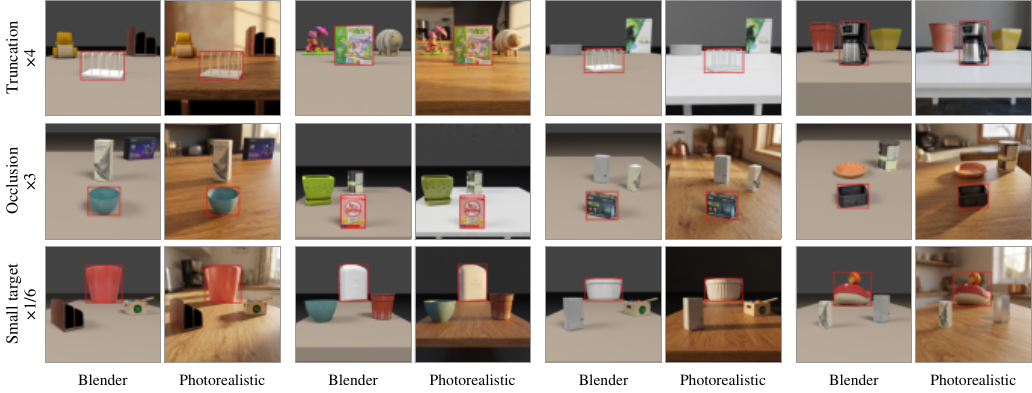}
\captionsetup{font=small,skip=4pt}
\caption{\textbf{Blender inputs and accepted photorealistic counterparts.} Four pairs per failure mode. The counterpart changes surface appearance, lighting, and background while passing the identity and geometry checks in Table~\ref{tab:app_stress_transfer_gates}. Red boxes identify targets. Both images retain the size error; the photorealistic image is not a corrected output.}
\label{fig:stress_pairs}
\end{figure*}

Gemini-3.1-Flash-Lite-Image~\citep{google2026flashliteimage} generates a photorealistic counterpart from each wrong-size Blender reference. We consider up to ten candidates and retain the first that passes the consistency gates in Table~\ref{tab:app_stress_transfer_gates}. The accepted set contains 109/116 scenes; the seven missing counterparts comprise five truncation scenes, one occlusion scene, and one small-target scene. Prompts use \rev{fixed} normal- and tight-table templates; generation is at 1024 pixels with a deterministic scene/environment/attempt seed. Figure~\ref{fig:stress_pairs} shows four accepted pairs per task.

\begin{table}[!htbp]
\centering\small
\captionsetup{font=small,skip=4pt}
\caption{\textbf{Acceptance checks for photorealistic counterparts.} These checks filter dataset inputs before planning; they are distinct from the output-level success gates. Pixel thresholds use the 512-coordinate grid.}
\label{tab:app_stress_transfer_gates}
\setlength{\tabcolsep}{6pt}
\resizebox{0.9\linewidth}{!}{%
\begin{tabular}{@{}lr@{}}
\toprule
Check against the Blender reference & Requirement \\
\midrule
Target identity (DINOv2~\revcitep{oquab2024dinov2} cosine similarity) & $\ge 0.60$ \\
Object center/contact drift & $\le 12$ pixels \\
Neighbor size error & $\le 0.075$ \\
Target size error & $\le 0.25$ \\
Median tight-table edge offset & $\le 12$ pixels \\
\bottomrule
\end{tabular}
}
\end{table}

The accepted images span eight environment templates: oak worktops (47), small oak tables (39), small white tables (10), marble counters (6), small birch tables (2), small walnut tables (2), wooden workbenches (2), and steel worktops (1). Small freestanding tables are used for tight-table scenes, while normal-table scenes mostly use wide worktops.

Let $\gamma$ denote the prescribed linear correction factor and $m$ the target's measured generation-induced size factor. We adjust the photorealistic request to $\gamma_{\mathrm{photo}}=\gamma/m$ rather than assuming image generation preserves target size exactly. SAM resegments the generated images using rendered boxes as prompts.

Both plans are executed on each photorealistic source by \rev{our editor}. Figure~\ref{fig:app_planner_more} shows additional photorealistic placement examples under the shared \rev{editor}.

\subsubsection{Blind Placement Preferences}
\label{app:stress_user_study}
\rev{Five raters compare the two planners' outputs on every scene, blinded to method and with randomized left/right order, choosing the placement that keeps the target visible, moderately sized, away from image boundaries, and from hiding neighbors, or a tie. Votes are scored $+1/0/-1$ (ours/tie/Gemini). Raters prefer \MethodName{}'s placements in both domains (Table~\ref{tab:app_stress_human}).}

\begin{table}[!htbp]
\centering\small
\captionsetup{font=small,skip=4pt}
\caption{\rev{\textbf{Blind placement preferences.} Five raters; scene-majority outcomes and mean preference on a $[-1,1]$ scale (positive favors \MethodName{}) with 95\% product-family bootstrap CIs.}}
\label{tab:app_stress_human}
\setlength{\tabcolsep}{4pt}
\resizebox{0.94\linewidth}{!}{%
\rev{\begin{tabular}{@{}lrrr@{}}
\toprule
Domain & Scenes & Ours / Gemini / tie & Preference [95\% CI] \\
\midrule
Blender & 116 & 30 / 15 / 71 & $+0.122\;[0.049,0.202]$ \\
Photorealistic & 109 & 40 / 8 / 61 & $+0.172\;[0.104,0.245]$ \\
\bottomrule
\end{tabular}}
}
\end{table}

\subsubsection{Complete Stress-Test Galleries}
\label{app:stress_examples}
\begin{figure*}[!tp]
\centering
\includegraphics[width=\linewidth,height=0.85\textheight,keepaspectratio]{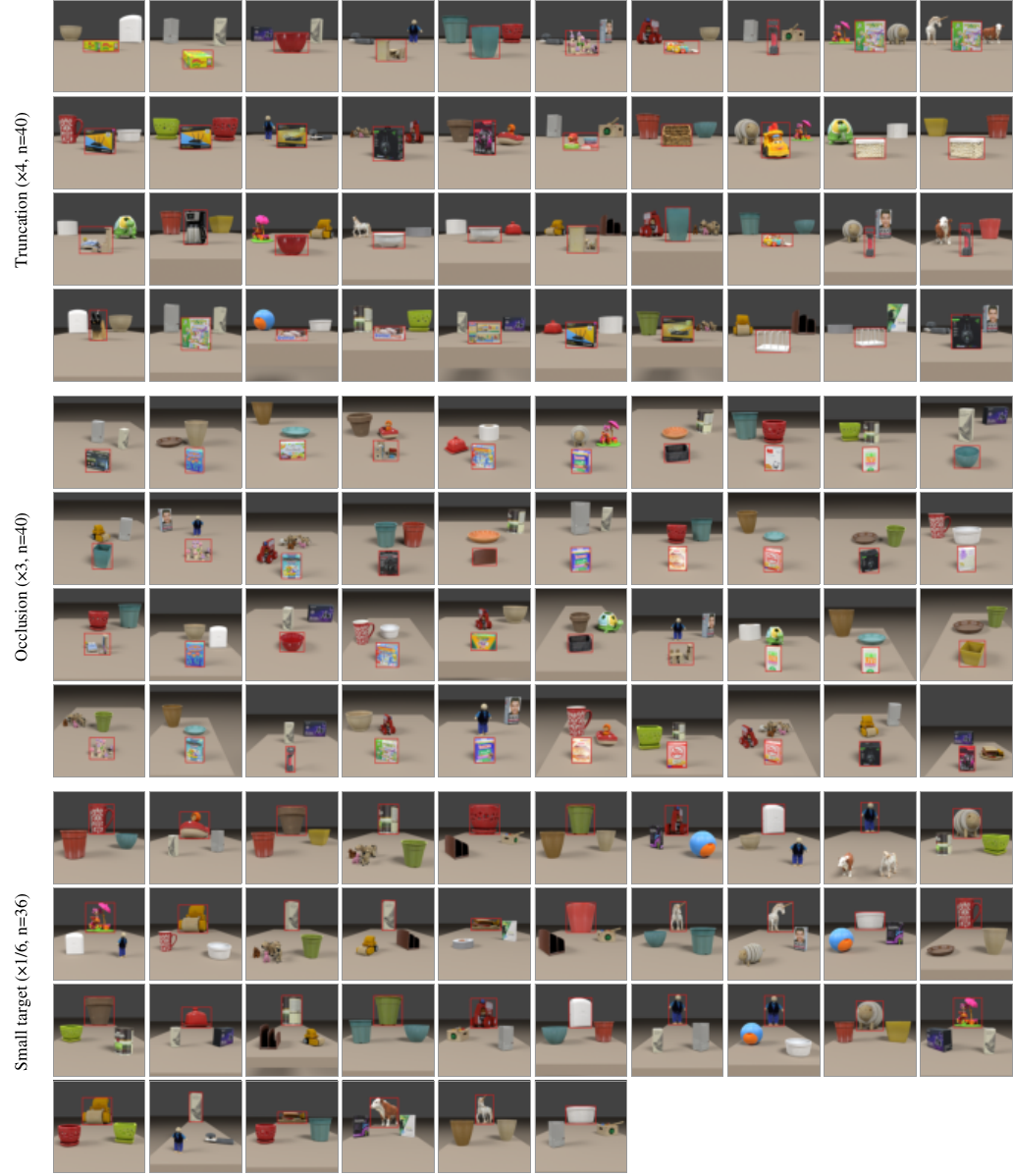}
\captionsetup{font=small,skip=4pt}
\caption{\textbf{All 116 Blender stress-test inputs.} Scenes are grouped by the failure caused by in-place correction; within each group, normal-table scenes precede tight-table scenes. Red boxes mark the wrong-size targets.}
\label{fig:stress_blender_all}
\end{figure*}

\begin{figure*}[!tp]
\centering
\includegraphics[width=\linewidth,height=0.85\textheight,keepaspectratio]{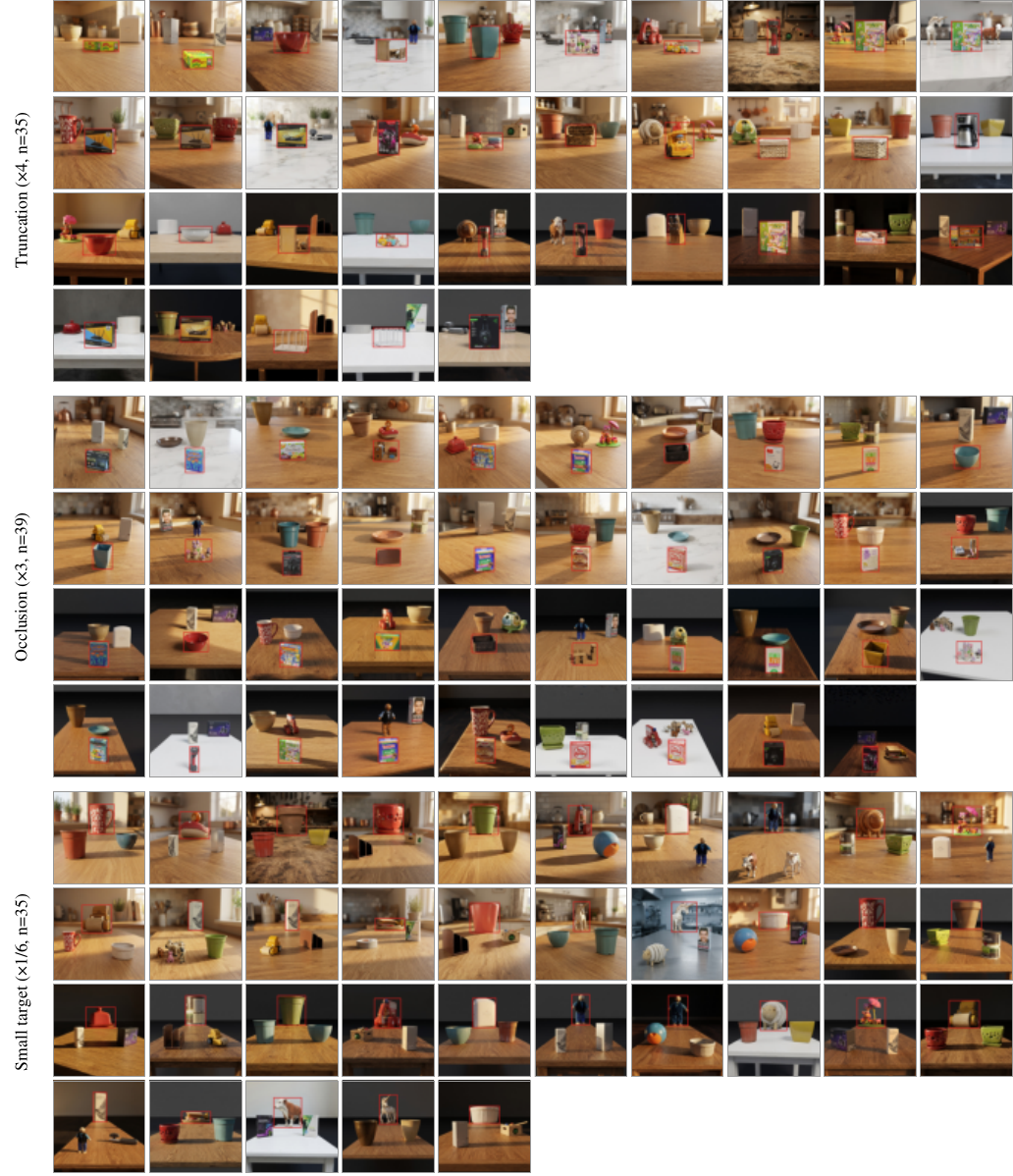}
\captionsetup{font=small,skip=4pt}
\caption{\textbf{All 109 accepted photorealistic stress-test inputs.} Grouping and relative order follow Figure~\ref{fig:stress_blender_all}, with seven unaccepted counterparts omitted. Red boxes mark the targets. The images are generated counterparts, not captured photographs.}
\label{fig:stress_photoreal_all}
\end{figure*}

\begin{figure*}[p]
\subsection{Photorealistic Planner Comparisons}
\label{app:qual_planner}
\begin{multicols}{2}
Figure~\ref{fig:app_planner_more} complements Fig.~\ref{fig:exp_planner_qual} with photorealistic stress scenes. Both planners receive the same fixed size correction, and their plans are executed by the same actuator. The displayed cases highlight placement differences after enlargement: the annotated Gemini outputs place targets beyond the support surface, against the image boundary, or partly outside the frame. These examples illustrate the support and framing criteria evaluated in Appendix~\ref{app:exp_planner}.
\end{multicols}
\centering
\captionsetup{font=small,skip=4pt,singlelinecheck=false}
\includegraphics[width=\linewidth,height=0.78\textheight,keepaspectratio]{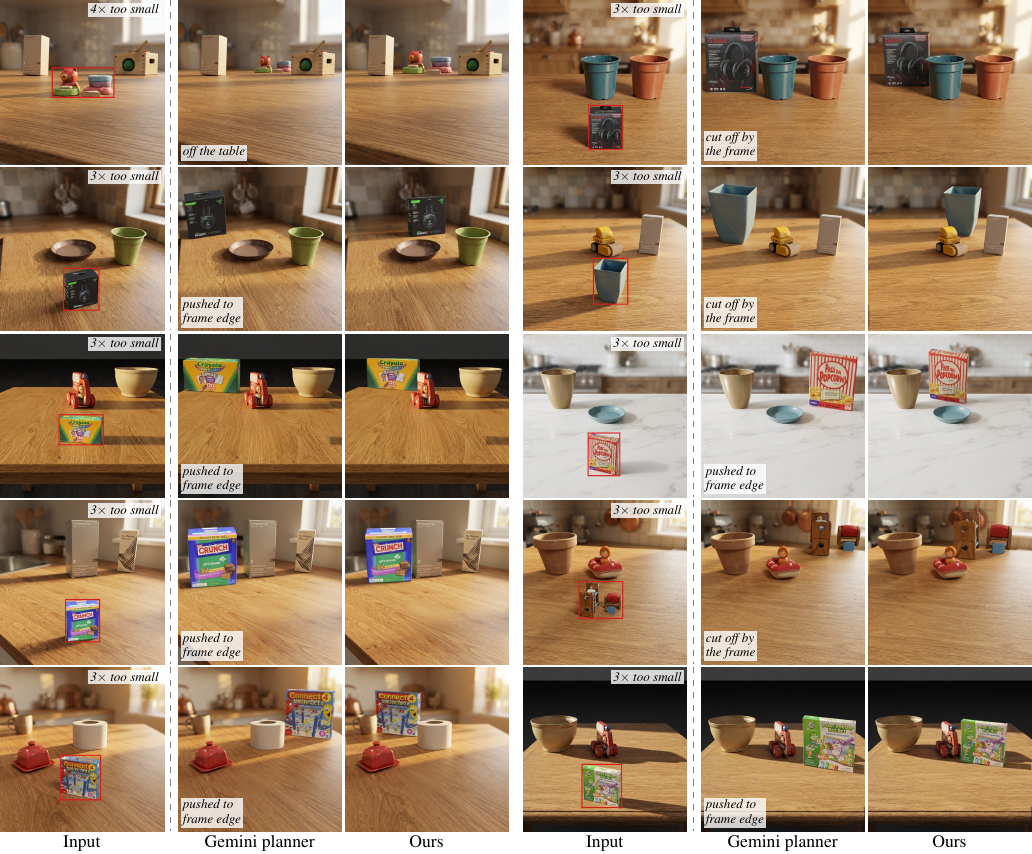}
\caption{\textbf{Additional planner comparisons on photorealistic stress scenes.} Ten examples are arranged in two groups of input, Gemini-planner output, and \MethodName{}-planner output. Red boxes identify the targets, and input annotations indicate the original size errors. Output annotations highlight support or frame-boundary problems in the displayed cases. The photorealistic images are generated counterparts of the rendered stress scenes, not independently captured photographs; aggregate automatic results appear in Table~\ref{tab:exp_planner} \revb{and human preferences in Table~\ref{tab:app_stress_human}}.}
\label{fig:app_planner_more}
\end{figure*}

Figures~\ref{fig:stress_blender_all} and~\ref{fig:stress_photoreal_all} show all 116 Blender inputs and all 109 accepted photorealistic inputs. Both galleries group scenes by failure mode and place normal-table scenes before tight-table scenes within each group. The photorealistic gallery omits only the seven rejected counterparts. Unlike the planner-output comparisons in Section~\ref{app:qual_planner}, these galleries document the evaluation inputs, not selected successful edits.

\FloatBarrier
\subsection{Controlled Editing and Training Ablations}
\label{app:exp_actuator}
\subsubsection{ObjectMover Protocol and Metrics}
Of 200 public requests, 12 are used for request-encoding development and 188 are reported, grouped into 175 scenes. The protocol was \rev{fixed} on the 12 development requests before the remaining destination photographs or archived outputs were downloaded, and \rev{our trained editor} was evaluated once on the 188 \rev{reported} requests with \rev{fixed} weights, inference settings, and encoding. Each box request specifies apparent scale $a=\sqrt{|B_{\mathrm{dst}}|/|B_{\mathrm{src}}|}$ and a destination mask centroid obtained by applying the same box-centered similarity transform. Coordinates are measured on a 512 grid and mapped to the $48\times48$ token grid. \MethodName{} uses the full 768 frame, seed 42, 40 steps, guidance one, and RoPE control for 20 steps, without the observer, planner, local crop, or paste-back.

\begin{figure*}[p]
\subsection{Prescribed Object Movement and Resizing}
\label{app:qual_objectmover}
\begin{multicols}{2}
Figure~\ref{fig:app_objectmover_more} extends Fig.~\ref{fig:exp_objectmover_qual} with same-request comparisons across varied targets and scene materials. The shared source and destination boxes allow direct inspection of requested placement and size, target identity, source-location cleanup, and changes to neighboring objects. The observer and planner are not part of this comparison.
\end{multicols}
\centering
\captionsetup{font=small,skip=4pt,singlelinecheck=false}
\includegraphics[width=\linewidth,height=0.78\textheight,keepaspectratio]{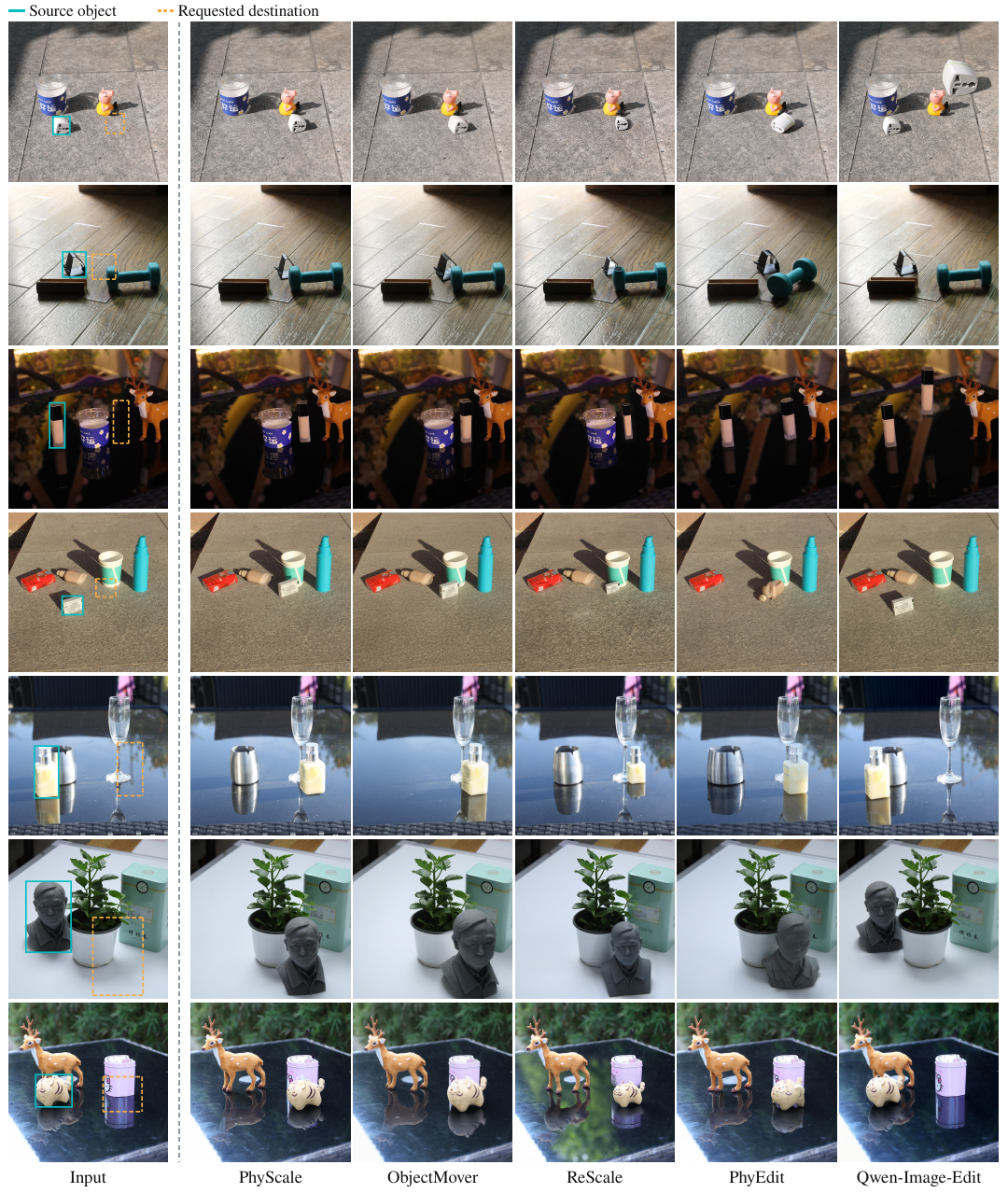}
\caption{\textbf{Additional ObjectMover actuator comparisons.} Seven prescribed edits, with columns showing the input, \MethodName{}, ObjectMover, the ReScale actuator, PhyEdit, and Qwen-Image-Edit. Solid cyan boxes mark the source objects and dashed orange boxes mark the requested destinations. All methods receive the same geometric request through their respective interfaces; ReScale is evaluated without its planner. Quantitative results and input adaptations are given in Table~\ref{tab:exp_actuator} and Appendix~\ref{app:exp_actuator}.}
\label{fig:app_objectmover_more}
\end{figure*}

Target DINO is the DINO-v1~\revcitep{caron2021dino} ViT-S/16 CLS cosine between tight target-box crops of output and destination photograph. Identity DINO instead compares the output target with the source-object crop. DreamSim~\revcitep{fu2023dreamsim} uses \rev{version 0.2.1 with its} default ensemble; PSNR is measured on the 512 grid. These are similarity metrics, not placement-success rates or naturalness ratings. ObjectMover uses its public output archive, whose sampling conditions are unknown. PhyEdit uses its Qwen-Image-Edit-2511 LoRA with 28 steps and Depth-Anything-3~\revcitep{lin2025depthanything3} input. The plain Qwen baseline receives a gray-background reference crop and textual coordinates, with 40 steps and guidance four; its performance applies to that interface. The released ReScale actuator receives the same ground-truth move/resize transform with its VLM planner removed; Gemini removes the source object from the conditioning background before editing. Figure~\ref{fig:exp_objectmover_qual} shows \rev{three} same-request \rev{comparisons}; its panels are illustrative rather than an additional evaluation sample. Figure~\ref{fig:app_objectmover_more} extends this comparison with seven further displayed requests.

\begin{table*}[!t]
\captionsetup{font=small,skip=4pt}
\centering
\small
\setlength{\tabcolsep}{2.5pt}
\caption{\textbf{Additional object-editing metrics on the same 188 requests.} Target and full-image PSNR compare outputs with the destination photograph. Qwen and PhyEdit use the adapted interfaces described in the text.}
\label{tab:app_actuator_metrics}
\resizebox{0.71\linewidth}{!}{%
\begin{tabular}{@{}lrrrrr@{}}
\toprule
Method & Target DINO $\uparrow$ & DreamSim $\downarrow$ & ID DINO $\uparrow$ & Target PSNR $\uparrow$ & Full PSNR $\uparrow$ \\
\midrule
\MethodName{} actuator & 0.8987 & 0.1050 & 0.8564 & 16.2659 & 27.0417 \\
ObjectMover (archive) & 0.8672 & 0.1402 & 0.8133 & 15.8262 & 24.8347 \\
PhyEdit (adapted) & 0.7912 & 0.2156 & 0.7383 & 14.3457 & 21.8593 \\
Qwen (crop + text) & 0.4101 & 0.6223 & 0.3790 & 10.5613 & 14.5137 \\
Unchanged source & 0.4309 & 0.6450 & 0.3193 & 11.1302 & 20.4608 \\
\bottomrule
\end{tabular}
}
\end{table*}

\subsubsection{Same-Backbone Controls}
The untrained RoPE control keeps the position channel and removes task adaptation. Its target DINO and DreamSim are 0.6338 and 0.3994; the main-table identity DINO is 0.561. Training on our 3D pairs raises target DINO by 0.265 (Table~\ref{tab:app_actuator_contrasts}). Section~\ref{app:exp_interface} reports a same-data \rev{retraining control that compares text-box and RoPE commands}.

The structured Qwen baseline adds blockwise depth ControlNet conditioning, a target-only hole dilated by eight pixels, a Gemini-cleaned background, and ReScale-style zoom crops. Its LoRA variant trains on 3,462 locally reconstructed pairs from the eight source datasets listed in ReScale, using rank 16, 3,000 optimizer steps, and effective batch eight. Its 24,000 sample exposures are close to \rev{the 24,680 exposures of our editor}, not an equal number of optimizer updates. The reconstructed pairs do not change object size. Structured inference uses 30 steps and seed 42.

\begin{table*}[!t]
\captionsetup{font=small,skip=4pt}
\caption{\textbf{Same-backbone structured controls.} The three paired DINO contrasts use the 175-scene grouping and their own Holm-adjusted family. \rev{Values for our editor come from this experiment's own scoring run.}}
\label{tab:app_structured_controls}
\centering\small
\setlength{\tabcolsep}{5pt}
\resizebox{0.66\linewidth}{!}{%
\begin{tabular}{@{}lrrr@{}}
\toprule
Qwen-Image-Edit-2511 configuration & Target DINO $\uparrow$ & DreamSim $\downarrow$ & Identity DINO $\uparrow$ \\
\midrule
Structured, untrained & 0.6957 & 0.3256 & 0.6530 \\
Structured, reconstructed-data LoRA & 0.6393 & 0.3908 & 0.5857 \\
\MethodName{} & 0.8987 & 0.1051 & 0.8564 \\
\bottomrule
\end{tabular}
}
\end{table*}

\rev{Our editor} minus structured-untrained target DINO is $0.203$ (95\% CI $[0.169,0.238]$); \rev{our editor} minus structured-LoRA is $0.259$ (95\% CI $[0.223,0.297]$). The reconstructed-data LoRA changes DINO by $-0.056$ (95\% CI $[-0.090,-0.023]$) relative to its untrained counterpart. These contrasts have Holm-adjusted $p$ between 0.0003 and 0.0006 in their specified family.

\begin{table*}[!t]
\captionsetup{font=small,skip=4pt}
\centering
\small
\setlength{\tabcolsep}{4pt}
\caption{\textbf{Target-DINO contrasts from the individual actuator experiments.} Higher target DINO is better. The mask contrast averages two-pixel erosion and dilation variants. Intervals come from the respective experiments.}
\label{tab:app_actuator_contrasts}
\resizebox{0.69\linewidth}{!}{%
\begin{tabular}{@{}lrrr@{}}
\toprule
Contrast & Difference & 95\% CI & Requests \\
\midrule
\rev{Ours} $-$ ObjectMover archive & $+0.031484$ & $[0.022327, 0.040418]$ & 188 \\
\rev{Ours} $-$ PhyEdit & $+0.107437$ & $[0.083411, 0.133518]$ & 188 \\
\rev{Ours} $-$ Qwen (crop + text) & $+0.488566$ & $[0.456006, 0.521376]$ & 188 \\
\rev{Ours $-$ untrained RoPE control} & $+0.264674$ & $[0.228537, 0.301586]$ & 188 \\
Mean of seeds 43, 44 $-$ archive & $+0.027493$ & $[0.018120, 0.036328]$ & 188 \\
Mean of mask perturbations $-$ reference & $-0.005470$ & $[-0.011716, -0.001038]$ & 188 \\
\bottomrule
\end{tabular}
}
\end{table*}

\subsubsection{Checkpoints and Supervision}
\label{app:exp_checkpoints}
\begin{table}[!htbp]
\captionsetup{font=small,skip=4pt}
\caption{\textbf{Checkpoint comparisons} on the 188 ObjectMover requests. Exposure counts are samples, not optimizer updates. \revb{Values come from a separate scoring run of the same outputs; DINO can differ from other tables by about 0.0002.}}
\label{tab:app_checkpoints}
\centering\small
\setlength{\tabcolsep}{4pt}
\resizebox{1\linewidth}{!}{%
\begin{tabular}{@{}lrrrr@{}}
\toprule
Variant & Rows & Exposures & Target DINO $\uparrow$ & DreamSim $\downarrow$ \\
\midrule
\rev{Ours, full training} & 2,468 & 24,680 & 0.8985 & 0.1050 \\
\rev{Ours, half the exposures} & 2,468 & 12,340 & 0.8955 & 0.1095 \\
RoPE only, untrained & \rev{0} & \rev{0} & 0.6338 & 0.3994 \\
\bottomrule
\end{tabular}
}
\end{table}

\rev{Halving the training exposures lowers target DINO by only 0.003, while removing training altogether lowers it by 0.265. The 3D-versus-2.5D comparison under matched training is reported in Appendix~\ref{app:rev_retrain}.}

\subsubsection{Textual versus Positional Commands}
\label{app:exp_interface}
\revb{The held-out renders cited alongside Table~\ref{tab:exp_actuator} in the main text are reported in Table~\ref{tab:app_interface_rendered}.}
\paragraph{Matched retraining.}
To compare command interfaces independently of backbone and data choice, we retrain a RoPE model and a text-box model from the same Qwen-Image-Edit-2511 checkpoint. Both use the same 2,468 rendered pairs, row order, 24,680 sample exposures, rank-16 LoRA, optimizer, and seed, on five B300 GPUs. RoPE uses the source-target positional transformation in Eq.~\eqref{eq:rope_warp}. The text-box model instead receives source and destination coordinates in its prompt: ``Move the \{target\} from box [x0,y0,x1,y1] to box [.\,.\,.] (0--1000 coordinates), resizing it to fit the new box. Keep .\,.\,. unchanged.'' These boxes come from the same prescribed similarity transform. Each model's auxiliary visual encoder receives the source image of its current training pair\rev{, as in all retrained models of Appendix~\ref{app:rev_retrain}}.

\paragraph{ObjectMover evaluation.}
We use the same 188 ObjectMover-A requests, scorer, inference stack, and geometric-association and failure-counting rules; paired intervals use 10,000 scene-group bootstrap samples. Table~\ref{tab:app_interface_control} reports the comparison with the main model as reference.

\begin{table*}[!t]
\centering\small
\captionsetup{font=small,skip=4pt}
\caption{\textbf{\rev{Command interfaces under matched retraining.}} Same-backbone, same-data retraining on 188 ObjectMover requests; \rev{our main model} is shown for reference. Geometric values are per-model medians.}
\label{tab:app_interface_control}
\setlength{\tabcolsep}{5pt}
\resizebox{0.65\linewidth}{!}{%
\begin{tabular}{@{}lrrr@{}}
\toprule
Metric & RoPE & Text boxes & \rev{Main model} \\
\midrule
Target DINO $\uparrow$ & 0.8953 & 0.4388 & 0.8987 \\
DreamSim $\downarrow$ & 0.108 & 0.620 & 0.105 \\
Identity DINO $\uparrow$ & 0.856 & 0.365 & 0.856 \\
Median destination error (pixels) $\downarrow$ & 2.4 & 216.5 & 2.4 \\
Median absolute log-scale error $\downarrow$ & 0.026 & 0.118 & 0.025 \\
\bottomrule
\end{tabular}
}
\end{table*}

\rev{With the same training pairs and budget, the text-box interface does not learn to execute the edits on these photographs: its target DINO of 0.439 is close to that of the unchanged source (0.431), whereas the RoPE interface reaches 0.895.} The RoPE-minus-text-box target-DINO difference is $+0.4565$ (95\% CI $[0.4248,0.4889]$). DreamSim, identity, destination, and scale contrasts also favor RoPE after Holm adjustment. \rev{The retrained RoPE model closely reproduces our main model (target DINO 0.8953 vs.\ 0.8987).}

\paragraph{Held-out rendered requests.}
To distinguish command following from transfer to photographs, we also evaluate 150 rendered requests with ground-truth destinations (Table~\ref{tab:app_interface_rendered}). RoPE-minus-text-box differences are $+0.400$ (95\% CI $[0.363,0.437]$) for target DINO and $+0.712$ (95\% CI $[0.666,0.753]$) for silhouette IoU. \rev{The text-box model places the object at the destination in only 67 of these 150 rendered requests, versus 149 for RoPE, so its gap on ObjectMover is not a transfer effect from renders to photographs.}

\begin{table*}[!t]
\centering\small
\captionsetup{font=small,skip=4pt}
\caption{\textbf{Rendered command-following control.} Results on 150 requests, including destination-placement failures in the reported counts\rev{; our main model is shown for reference}.}
\label{tab:app_interface_rendered}
\setlength{\tabcolsep}{7pt}
\resizebox{0.63\linewidth}{!}{%
\begin{tabular}{@{}lrrr@{}}
\toprule
Metric & RoPE & Text boxes & \rev{Main model} \\
\midrule
Target-box DINO $\uparrow$ & 0.921 & 0.522 & 0.921 \\
Silhouette IoU $\uparrow$ & 0.807 & 0.095 & 0.811 \\
Object placed at destination & 149/150 & 67/150 & 149/150 \\
\bottomrule
\end{tabular}
}
\end{table*}

\subsubsection{3D versus 2.5D Supervision}
\label{app:exp_supervision}
\rev{Under matched training (Appendix~\ref{app:rev_retrain}), 3D supervision yields higher target DINO than 2.5D on real photographs (0.879 vs.\ 0.864) and higher target-box DINO and silhouette IoU on held-out renders (0.902 vs.\ 0.871; 0.744 vs.\ 0.730). This subsection measures the scene effects that box-level metrics miss: shadows and perceived physical plausibility.} Both arms use 783 training rows and 7,830 exposures and differ only in whether the training destination is a full 3D re-render or an analytic 2.5D composite that warps and composites the source object. \rev{The shadow measurements and user study below use a separately trained matched pair of 3D and 2.5D models on the same 783 rows, with identical trainer, recipe, and budget.}

The 150 held-out requests are rendered with the \rev{same pipeline as our training pairs}, with camera and command ranges restricted to the 783-row training distribution and assets disjoint from all training manifests. Of the 150 requests, 108 change depth and 42 change only size; rendered overlap defines 40 target-in-front, 46 target-behind, and 64 unoccluded requests. All arms receive identical source pixels, commands, prompts, and seeds; intervals use 10,000 asset-pair cluster-bootstrap draws.

\paragraph{Shadow fidelity.}
Because target-box appearance metrics largely exclude ground shadows, we add pixel measurements against the true 3D render and a companion render with the moved object absent. Darkening relative to that companion defines shadow regions. We evaluate both the new target location and residual shadow at the old location. Table~\ref{tab:app_supervision_shadow} reports the errors and intervals. Shadow depth is relative to the true-render darkening (one denotes matching depth), not the fraction of correct shadows. The same qualitative conclusion holds on the 144 requests where all arms place the target correctly.

\begin{table*}[!t]
\centering\small
\captionsetup{font=small,skip=4pt,singlelinecheck=false}
\setlength{\tabcolsep}{3pt}
\caption{\textbf{Shadow measurements on 150 held-out renders.} Lower errors are better; shadow depth should approach one. Differences are 3D minus 2.5D; $p$ values are Holm-adjusted over the three shadow errors. \rev{Our main model (Main)} is shown for reference.}
\label{tab:app_supervision_shadow}
\resizebox{0.73\linewidth}{!}{%
\begin{tabular}{@{}lrrrlr@{}}
\toprule
Measure & 3D & 2.5D & \rev{Main} & Difference [95\% CI] & Holm $p$ \\
\midrule
Shadow luminance error ($n=147$) & 0.261 & 0.373 & 0.214 & $-0.112\;[-0.141,-0.086]$ & 0.0003 \\
New-location shadow error ($n=137$) & 0.384 & 0.564 & 0.316 & $-0.180\;[-0.229,-0.136]$ & 0.0003 \\
Old-location shadow residue ($n=132$) & 0.078 & 0.100 & 0.064 & $-0.022\;[-0.034,-0.008]$ & 0.0042 \\
Shadow depth (true render $=1$) & 0.61 & 0.06 & 0.88 & $+0.54\;[0.49,0.60]$ & \rev{n/a} \\
\bottomrule
\end{tabular}
}
\end{table*}

\paragraph{User study on physical plausibility.}
\label{app:exp_supervision_human}
\rev{Three raters compare the 3D and 2.5D outputs on 90 of the 150 rendered requests, sampled by occlusion class, blinded to method and with randomized left/right order. The primary question asks which output is more physically plausible, considering new contact and shadows, old-location remnants, and occlusion; a second asks which has more plausible shadows. Raters strongly prefer 3D supervision (Table~\ref{tab:app_supervision_human}).}

\begin{table}[!htbp]
\centering\small
\captionsetup{font=small,skip=4pt}
\caption{\rev{\textbf{User study comparing 3D and 2.5D supervision.} Three raters, 90 rendered requests. Preference for 3D with ties counted as one half (50\% is neutral) and 95\% asset-pair bootstrap CIs.}}
\label{tab:app_supervision_human}
\setlength{\tabcolsep}{5pt}
\resizebox{0.9\linewidth}{!}{
\rev{\begin{tabular}{@{}p{0.56\linewidth}rr@{}}
\toprule
Measure & Estimate & 95\% CI \\
\midrule
Physical plausibility (primary) & 83.5\% & [76.6, 89.8] \\
Shadow plausibility & 85.5\% & [78.8, 91.5] \\
\bottomrule
\end{tabular}}
}
\end{table}

\begin{figure*}[p]
\subsection{3D versus 2.5D Supervision}
\label{app:qual_supervision}
\begin{multicols}{2}
Figure~\ref{fig:app_supervision_more} expands Fig.~\ref{fig:exp_supervision_qual} with real-image comparisons of the two supervision types. In addition to the edited object's appearance, the panels expose effects outside its destination box, including remnants at the original location and changes to shadows and reflections. Appendix~\ref{app:exp_supervision} quantifies these effects on held-out renders.
\end{multicols}
\centering
\captionsetup{font=small,skip=4pt,singlelinecheck=false}
\includegraphics[width=\linewidth,height=0.78\textheight,keepaspectratio]{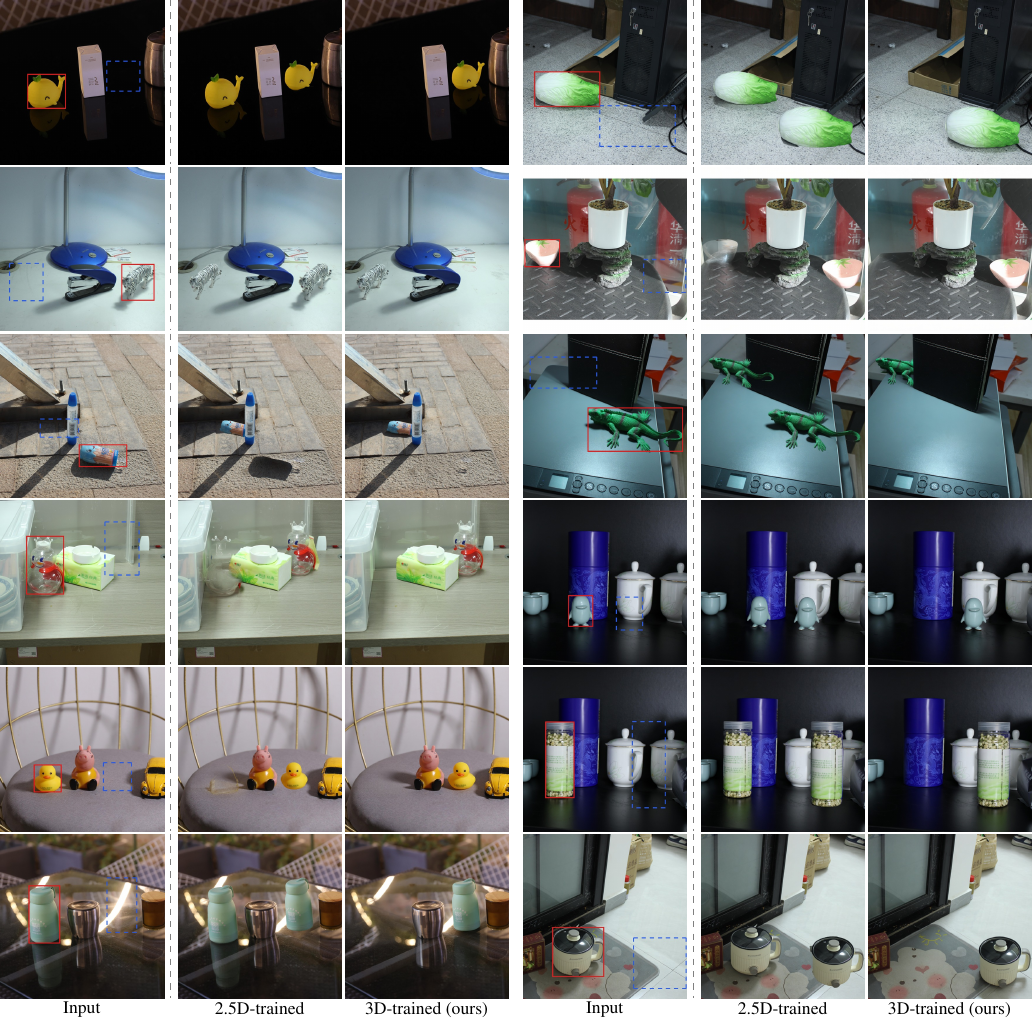}
\caption{\textbf{Additional visual comparisons of 2.5D and 3D supervision.} Twelve displayed requests are arranged in two groups of input, 2.5D-trained output, and 3D-trained output. Red source boxes and dashed blue destination boxes specify the edits. The examples show why inspecting only the object at its new location can miss residual copies, old shadows, or inconsistent surrounding appearance.}
\label{fig:app_supervision_more}
\end{figure*}

\rev{The 270 primary judgments comprise 206 preferences for 3D, 39 ties, and 25 for 2.5D, and the preference holds for target-in-front, target-behind, and unoccluded requests alike (84.7\%, 80.4\%, and 85.1\%). Figures~\ref{fig:exp_supervision_qual} and~\ref{fig:app_supervision_more} show real-image examples of the residual shadows and source-object remnants left by 2.5D training.}

\subsubsection{Training Dynamics}
\label{app:exp_training_curves}
\begin{figure*}[!t]
\centering
\includegraphics[width=0.85\linewidth]{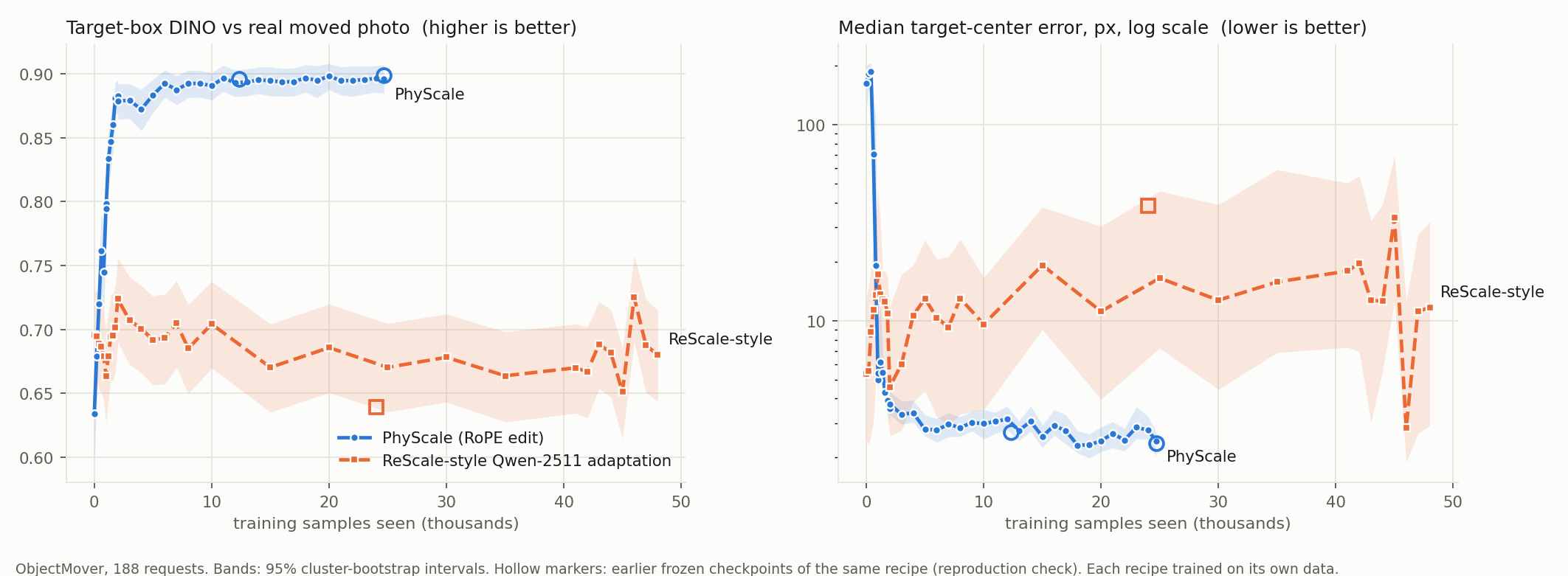}
\captionsetup{font=small,skip=4pt,singlelinecheck=false}
\caption{\textbf{Training dynamics on ObjectMover.} Left: target DINO against real destination photographs ($\uparrow$). Right: median target-center error in pixels on a logarithmic axis ($\downarrow$). The horizontal axis is training sample exposures in thousands. Shaded bands are 95\% cluster-bootstrap CIs over evaluation samples. The two recipes use different training data; the orange curve is the reconstructed ReScale-style adaptation, not the released ReScale actuator.}
\label{fig:app_training_curves}
\end{figure*}

Figure~\ref{fig:app_training_curves} evaluates checkpoints by sample exposures, the common unit across the two training scripts. \rev{Our recipe} learns from true 3D move/resize pairs; the ReScale-style Qwen adaptation learns from the reconstructed insertion data described above, whose objects do not change size. \rev{The comparison therefore shows how quickly each training recipe teaches the shared backbone to follow move/resize requests.} Both curves use the 188-request ObjectMover split.

\begin{table*}[!t]
\centering\small
\captionsetup{font=small,skip=4pt,singlelinecheck=false}
\setlength{\tabcolsep}{4pt}
\caption{\textbf{Selected training-curve checkpoints.} Sample exposures are not unique training pairs or optimizer updates. \rev{Our curve comes from a separate run of our training recipe; the main model is listed for reference.}}
\label{tab:app_training_curve_points}
\resizebox{0.74\linewidth}{!}{%
\begin{tabular}{@{}lrrlr@{}}
\toprule
Recipe / checkpoint & Exposures & Target DINO & 95\% CI & Center error (px) \\
\midrule
\rev{Ours}, untrained RoPE & 0 & 0.6340 & $[0.5959,0.6706]$ & 163.005 \\
\rev{Ours}, early & 1,000 & 0.7946 & $[0.7643,0.8230]$ & 5.410 \\
\rev{Ours}, early & 2,000 & 0.8788 & $[0.8642,0.8922]$ & 3.749 \\
\rev{Ours}, intermediate & 12,000 & 0.8928 & $[0.8821,0.9028]$ & 3.145 \\
\rev{Ours, separate run}, endpoint & 24,680 & 0.8962 & $[0.8849,0.9066]$ & 2.439 \\
\rev{Ours, main model} & 24,680 & 0.8987 & $[0.8882,0.9084]$ & 2.357 \\
\midrule
ReScale-style, untrained & 0 & 0.6957 & $[0.6605,0.7306]$ & 5.363 \\
ReScale-style, early & 2,000 & 0.7237 & $[0.6900,0.7560]$ & 4.589 \\
ReScale-style, highest tested DINO & 46,000 & 0.7250 & $[0.6904,0.7576]$ & 2.831 \\
ReScale-style, endpoint & 48,000 & 0.6801 & $[0.6437,0.7150]$ & 11.645 \\
\bottomrule
\end{tabular}
}
\end{table*}

\rev{Our recipe} reaches target DINO 0.8788 after 2,000 exposures and then remains near 0.89--0.90 over the measured range; \rev{this separate run ends at 0.8962, close to the main model (0.8987).} The endpoint difference against the reconstructed ReScale-style recipe is $+0.216$ (95\% CI $[0.18084,0.25282]$); \rev{even against that recipe's best checkpoint, the difference is $+0.171$ (95\% CI $[0.13887,0.20526]$), and extending its training does not close the gap.}

\subsubsection{Command and Geometric Diagnostics}
\label{app:exp_command_diagnostics}

\begin{figure}[!htbp]
\centering
\includegraphics[width=0.92\linewidth]{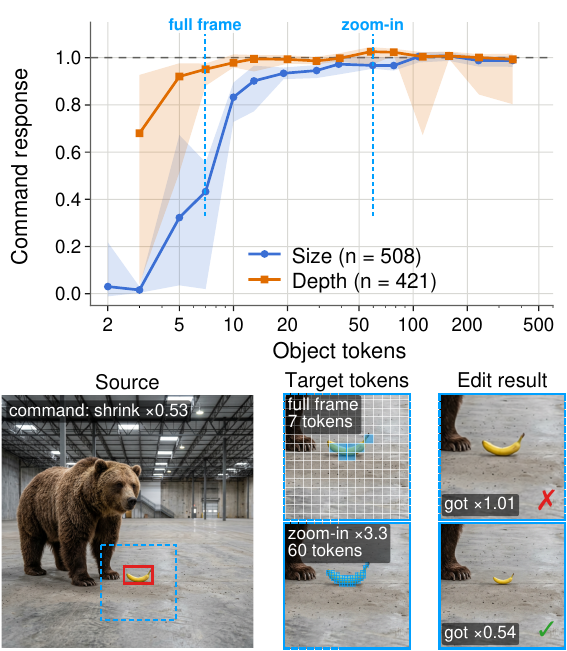}
\captionsetup{font=small,skip=4pt,singlelinecheck=false}
\caption{\textbf{Target-token support and local crop enlargement.} Top: size and depth curves report the fraction of the requested command executed, not a binary success rate; unit response follows the request. Bottom: a GenScale target requested at $0.53\times$ has seven command-carrying tokens in the full frame and remains at $1.01\times$. A $3.3\times$ enlarged crop provides 60 tokens and reaches $0.54\times$. Dashed lines on the curves mark these support levels.}
\label{fig:exp_token_response}
\end{figure}

The controlled size--depth response test contains 24 groups with nine command combinations. Estimated own-command responses are $0.910$ (95\% CI $[0.798,0.991]$) for size and $1.014$ (95\% CI $[1.005,1.030]$) for depth, where 1 means the requested change is executed exactly. Figure~\ref{fig:exp_token_response} reports command response, not binary success. Low token support attenuates size response; enlarged local crops increase the number of command-carrying tokens and recover the requested change in the illustrated example.

\rev{Table~\ref{tab:app_geometry} measures placement and scale on the 188 ObjectMover requests by automatically associating the edited object in each output with its destination. Our actuator places objects about twice as accurately as ObjectMover and matches the requested scale more closely.} Seed and mask-perturbation contrasts are in Table~\ref{tab:app_actuator_contrasts}.

\begin{table*}[!t]
\captionsetup{font=small,skip=4pt}
\centering
\small
\setlength{\tabcolsep}{3pt}
\caption{\textbf{Geometric accuracy on 188 ObjectMover requests.} Per-method medians; position is measured on a 512-coordinate grid.}
\label{tab:app_geometry}
\resizebox{0.49\linewidth}{!}{%
\begin{tabular}{@{}lrrr@{}}
\toprule
Method & Position (px) $\downarrow$ & $|\log\mathrm{scale}|$ $\downarrow$ & Mask IoU $\uparrow$ \\
\midrule
\MethodName{} actuator & \textbf{2.3572} & \textbf{0.0251} & \textbf{0.8919} \\
ObjectMover (archive) & 5.4315 & 0.0466 & 0.7739 \\
PhyEdit (adapted) & 11.6666 & 0.0513 & 0.6682 \\
Qwen (crop + text) & 141.0741 & 0.2629 & 0.0000 \\
Unchanged source & 229.4228 & 0.0613 & 0.0000 \\
\bottomrule
\end{tabular}
}
\end{table*}

\FloatBarrier
\subsection{User Study on Editing Quality}
\label{app:exp_human}
\rev{This study shows that our editor preserves object identity and looks natural on real photographs. Six raters complete a questionnaire over 50 ObjectMover sources. Each item shows the edit request and two unnamed outputs for the same source, ours and one opponent's; raters state a preference (tie allowed) and rate identity and naturalness on 0--4 scales. Table~\ref{tab:exp_human} reports pooled means, and Table~\ref{tab:app_human_paired} compares each opponent with our output on the same items (60 paired items over 30 sources per opponent). \revb{Completion is the share of ratings that mark the request as fully completed (the object is moved or resized as requested and no copy remains at its source), among the ratings that answer this question: 133 of 148 for \MethodName{}, 37 of 50 for ObjectMover, 15 of 48 for PhyEdit, and 1 of 50 for Qwen.} Raters score our outputs above PhyEdit and Qwen in both identity and naturalness, and on par with ObjectMover.}

\begin{table}[!htbp]
\centering\small
\captionsetup{font=small,skip=4pt,singlelinecheck=false}
\setlength{\tabcolsep}{3pt}
\caption{\textbf{Paired human editing results.} Identity and naturalness differences are ours minus opponent on 0--4 scales\rev{, with} 95\% source-bootstrap CIs; 60 paired items per opponent.}
\label{tab:app_human_paired}
\resizebox{1\linewidth}{!}{%
\begin{tabular}{@{}lll@{}}
\toprule
Opponent & Identity diff. & Naturalness diff. \\
\midrule
ObjectMover & $0.00\;[-0.48,0.29]$ & $-0.03\;[-0.40,0.28]$ \\
PhyEdit & $+1.65\;[0.96,2.29]$ & $+1.22\;[0.63,1.79]$ \\
Qwen & $+0.73\;[0.41,1.06]$ & $+2.32\;[1.80,2.82]$ \\
\bottomrule
\end{tabular}
}
\end{table}

\rev{ReScale was rated in a separate questionnaire with 60 identity and naturalness ratings (means 3.600 and 3.100); pairwise preferences are 32 for \MethodName{}, 11 for ReScale, and 17 ties, and 39 of its 50 outputs are rated fully completed (78.0\%). It therefore appears as a separately marked row in Table~\ref{tab:exp_human}.}

\FloatBarrier
\subsection{Pipeline Time and Cost}
\label{app:exp_efficiency}
\rev{The time and cost comparison uses 30 GenScale sources shared by both pipelines. Times are per-source sums of stage runtimes (observation, planning, editing, and restoration), and costs are estimated from measured compute time and API usage at the unit rates below. This complete-pipeline comparison is the basis of the one-twentieth cost ratio in the main text \revb{(a factor of 18.8)}; Appendix~\ref{app:exp_swap} separately projects costs for front-end and editor combinations on the 196-source swap subset. Our planner uses exact lazy token checking (Appendix~\ref{app:planner_complexity}), which selects exactly the same plans as exhaustive checking on all 27 sources with planning actions and reduces mean planning time from 189.7 to 18.1 seconds; Table~\ref{tab:app_pipeline_cost} lists both implementations, which differ only in planning time.}

\begin{table*}[!t]
\captionsetup{font=small,skip=4pt}
\caption{\textbf{\rev{Pipeline time and cost.}} Means over 30 sources. Prices are estimates at the \rev{listed} unit rates, not invoices.}
\label{tab:app_pipeline_cost}
\centering\small
\setlength{\tabcolsep}{6pt}
\resizebox{0.6\linewidth}{!}{%
\begin{tabular}{@{}lrr@{}}
\toprule
Pipeline & Summed stage time (s) & Estimated cost (\$) \\
\midrule
\MethodName{}, exhaustive token checks & 224.3 & 0.039 \\
\MethodName{}, exact lazy token checks & 52.8 & 0.037 \\
ReScale & 98.0 & 0.696 \\
\bottomrule
\end{tabular}
}
\end{table*}

The paired \MethodName{}-minus-ReScale time difference \rev{(lazy token checks)} is $-45.2$ seconds (95\% CI $[-68.1,-24.8]$) and the cost difference is $-\$0.659$ (95\% CI $[-0.86,-0.47]$). Compute prices are \$0.000542 per L40S second, \$0.001261 per H200 second, \$0.0000131 per CPU core-second, and \$0.00000222 per GiB-second. Gemini-3.1-Pro input/output rates are \$2/\$12 per million tokens at the time of measurement. ReScale's mean API cost is \$0.554 per source, including \$0.425 for diagnosis \revb{and \$0.029 for planning and layout grounding}.

\FloatBarrier
\clearpage

\clearpage

\Needspace{6\baselineskip}
\section{\texorpdfstring{\rev{Additional Evaluations}}{Additional Evaluations}}
\label{app:rev}

\subsection{\texorpdfstring{\rev{Independent Judges and Human Evaluation on GenScale}}{Independent Judges and Human Evaluation on GenScale}}
\label{app:rev_judges}
\rev{This subsection shows that \MethodName{}'s GenScale result holds under judges independent of ReScale's planner, without the knowledge base shared with the judge, and with blinded human raters. ReScale takes its diagnoses from the scores of the Gemini~3.1~Pro judge, the main GenScale judge, so they agree with that judge far more often than with judges from other model families (Appendix~\ref{app:rev_coupling}). We therefore score a stratified subset of 700 GenScale sources (100 per generator, 2,029 relations, fixed by hash order before scoring) with three judges, each scoring both corrections in a single batch: Gemini with the main five-vote protocol, and GPT-6.1 Terra~\revcitep{openai2026gpt61terra} and Claude Sonnet~5.5~\revcitep{anthropic2026sonnet55} with the unchanged GenScale judging prompt and one vote per relation, with no refused or lost judgment. Under the Gemini judge, whose scores supply ReScale's diagnoses, ReScale has a slightly lower error. Under both independent judges, \MethodName{} has the lower error and repairs more relations than ReScale (584 vs.\ 444 under GPT; 543 vs.\ 504 under Claude; Table~\ref{tab:app_rev_judges}, top). Both methods clearly improve on the source images, whose errors are 1.036 under GPT and 0.897 under Claude; \MethodName{} reduces them by 0.356 and 0.313. With relations weighted equally, as in Table~\ref{tab:exp_genscale}, \MethodName{} minus ReScale is $-0.092$ $[-0.143,-0.042]$ under GPT and $-0.025$ $[-0.075,+0.024]$ under Claude.}

\begin{table}[!htbp]
\captionsetup{font=small,skip=4pt}
\caption{\rev{\textbf{GenScale error under three judges.} Mean $|s-3|$ per source ($\downarrow$, sources weighted equally) on the same 700 sources (2,029 relations); unjudgeable relations receive error 2. Gemini uses the main five-vote protocol; GPT and Claude use one vote per relation. Top: GenScale judging prompt. Bottom: the same prompt with all knowledge-base dimensions and expected ratios removed. Diff.: \MethodName{} minus ReScale, with 95\% prompt-clustered bootstrap CIs.}}
\label{tab:app_rev_judges}
\centering\small
\setlength{\tabcolsep}{4pt}
\resizebox{1\linewidth}{!}{%
\rev{\begin{tabular}{@{}lrrr@{}}
\toprule
Judge & \MethodName{} & ReScale & Diff.\ [95\% CI] \\
\midrule
\multicolumn{4}{@{}l}{\emph{With knowledge-base dimensions}} \\
Gemini 3.1 Pro & 0.617 & \textbf{0.575} & $+0.042\;[-0.021,+0.103]$ \\
GPT-6.1 Terra & \textbf{0.680} & 0.733 & $-0.053\;[-0.109,+0.004]$ \\
Claude Sonnet 5.5 & \textbf{0.583} & 0.607 & $-0.024\;[-0.077,+0.029]$ \\
\midrule
\multicolumn{4}{@{}l}{\emph{Without knowledge-base dimensions}} \\
GPT-6.1 Terra & \textbf{0.543} & 0.596 & $-0.052\;[-0.106,+0.000]$ \\
Claude Sonnet 5.5 & \textbf{0.496} & 0.542 & $-0.046\;[-0.096,+0.003]$ \\
\bottomrule
\end{tabular}}
}
\end{table}

\rev{\textbf{Judging without the knowledge base.} \MethodName{} and the judging prompt use \revb{essentially} the same table of typical object dimensions \revb{(Appendix~\ref{app:observer})}. To remove this shared information, we delete every typical length and expected ratio from the prompt and ask the judges to rely on their own knowledge of real-world object sizes; the scoring rubric and output format are unchanged. \MethodName{} still reduces the source error by 0.283 (GPT) and 0.175 (Claude), with 95\% prompt-clustered bootstrap CIs of $[+0.230,+0.337]$ and $[+0.123,+0.227]$, and its error remains below ReScale's under both judges (Table~\ref{tab:app_rev_judges}, bottom). Without numeric references, both judges rate more source relations as plausible (source error 0.826 under GPT and 0.670 under Claude), which shrinks the available improvement for both methods alike.}

\rev{\textbf{Human evaluation.} Five raters, none of whom is an author, compare \MethodName{} and ReScale on 105 GenScale sources (15 per generator) sampled from those where the two outputs differ. Each item shows the source image, its prompt, and the two corrections in random left--right order without method names; raters choose the output with more plausible relative sizes (primary question), better content preservation, and higher naturalness, or answer ``same''. Votes are scored $+1/0/-1$ (ours/same/ReScale). Raters prefer \MethodName{} on all three questions (Table~\ref{tab:app_rev_human}). Counting the 381 sources on which the two methods return identical images as ties extends the comparison to all 2,800 sources, with preferences of $+0.19$, $+0.34$, and $+0.46$ (95\% CIs $[+0.11,+0.28]$, $[+0.26,+0.41]$, and $[+0.37,+0.55]$).}

\begin{table}[!htbp]
\captionsetup{font=small,skip=4pt}
\caption{\rev{\textbf{Blinded human comparison with ReScale.} Preference on a $[-1,1]$ scale (positive favors \MethodName{}) over 105 sources and five raters, with 95\% prompt-clustered bootstrap CIs and vote counts.}}
\label{tab:app_rev_human}
\centering\small
\setlength{\tabcolsep}{4pt}
\rev{\resizebox{1\linewidth}{!}{%
\begin{tabular}{@{}lrr@{}}
\toprule
Question & Preference [95\% CI] & Ours / ReScale / same \\
\midrule
Relative size & $+0.223\;[+0.120,+0.320]$ & 195 / 78 / 252 \\
Content preservation & $+0.390\;[+0.297,+0.480]$ & 235 / 30 / 260 \\
Naturalness & $+0.531\;[+0.410,+0.644]$ & 323 / 44 / 158 \\
\bottomrule
\end{tabular}
}}
\end{table}

\subsection{\texorpdfstring{\rev{Agreement Between ReScale's Diagnoses and the Judges}}{Agreement Between ReScale's Diagnoses and the Judges}}
\label{app:rev_coupling}
\rev{ReScale takes its diagnoses directly from the GenScale judge: its pipeline reads the scores that the Gemini~3.1~Pro judge assigns to the source image, and the constraints of its edit plans reproduce these scores and their rationales for all 7,521 diagnosed relations of the 2,800 sources. We compare the verdict implied by each ReScale diagnosis (keep, enlarge, or shrink the target) with each judge's verdict on the source image in a separate evaluation run, over the 1,899 relations of the 700-source subset that carry a ReScale diagnosis (Table~\ref{tab:app_rev_coupling}). ReScale's diagnoses coincide with the Gemini judge on 88.5\% of relations, well above the 74--77\% agreement between any two judges, and with the independent GPT and Claude judges on 74.9\% and 78.0\%. The comparison under the Gemini judge therefore favors ReScale's front end; under the two independent judges, \MethodName{} has the lower error (Appendix~\ref{app:rev_judges}).}

\begin{table}[t]
\centering
\caption{\rev{\textbf{Verdict agreement on source images (\%).} 1,899 relations; verdicts are keep, enlarge, or shrink.}}
\label{tab:app_rev_coupling}
\small
\begin{tabular}{lccc}
\toprule
& \rev{Gemini} & \rev{GPT} & \rev{Claude} \\
\midrule
\rev{ReScale diagnosis} & \rev{\textbf{88.5}} & \rev{74.9} & \rev{78.0} \\
\rev{Gemini judge} & \rev{---} & \rev{73.9} & \rev{76.7} \\
\rev{GPT judge} & \rev{73.9} & \rev{---} & \rev{75.7} \\
\bottomrule
\end{tabular}
\end{table}

\subsection{\texorpdfstring{\rev{Sources Not Used During Development}}{Sources Not Used During Development}}
\label{app:rev_heldout}
\rev{This subsection shows that the GenScale comparison does not depend on sources seen during development. We tuned planner thresholds on 230 GenScale sources. Removing them leaves 2,570 sources on which \MethodName{} minus ReScale is $-0.001$, so the overall comparison in Table~\ref{tab:exp_genscale} is unchanged. \revb{Among the 700 sources scored by the independent judges (Appendix~\ref{app:rev_judges}), 355 appear in none of our development or pilot experiments. On these sources, with relations weighted equally, \MethodName{} minus ReScale is $-0.116$ (95\% CI $[-0.190,-0.040]$) under GPT and $-0.035$ (95\% CI $[-0.105,+0.037]$) under Claude.}}

\rev{\textbf{Edited and unedited sources.} \MethodName{} edits 2,091 of the 2,800 sources (75\%) and ReScale 1,874 (67\%). On the sources that \MethodName{} edits, its error is lower than ReScale's (0.625 vs.\ 0.671; difference $-0.046$, 95\% prompt-clustered bootstrap CI $[-0.091,-0.001]$), as it is on the 1,546 sources that both methods edit (Table~\ref{tab:app_rev_edited}). The overall parity in Table~\ref{tab:exp_genscale} therefore does not come from processing sources selectively: the 709 sources that \MethodName{} delivers unchanged keep their source error, whereas ReScale lowers it. Among these 709 sources, the planner finds no correction necessary in 232; in 305, fewer than two prompted objects are available as separate, measurable instances; in 137, the geometric fit of an object fails, for example because its contact point lies above the estimated horizon; and 35 are unchanged for other reasons.}

\begin{table*}[!tp]
\captionsetup{font=small,skip=4pt}
\caption{\rev{\textbf{GenScale error by edit decision.} Mean $|s-3|$ over relations ($\downarrow$) under the main protocol, grouped by whether each method changes the source image.}}
\label{tab:app_rev_edited}
\centering\small
\setlength{\tabcolsep}{4pt}
\resizebox{0.62\linewidth}{!}{%
\rev{\begin{tabular}{@{}lrrrrr@{}}
\toprule
Sources & Count & Relations & Source & \MethodName{} & ReScale \\
\midrule
Edited by \MethodName{} & 2,091 & 6,688 & 0.947 & \textbf{0.625} & 0.671 \\
\quad edited by both & 1,546 & 5,253 & 0.990 & \textbf{0.608} & 0.638 \\
\quad edited by \MethodName{} only & 545 & 1,435 & 0.790 & \textbf{0.685} & 0.790 \\
Unchanged by \MethodName{} & 709 & 1,313 & 0.718 & 0.718 & \textbf{0.538} \\
\quad edited by ReScale only & 328 & 736 & 0.986 & 0.985 & \textbf{0.664} \\
\quad unchanged by both & 381 & 577 & 0.376 & 0.378 & 0.376 \\
\midrule
All & 2,800 & 8,001 & 0.910 & \textbf{0.640} & 0.649 \\
\bottomrule
\end{tabular}}
}
\end{table*}

\subsection{\texorpdfstring{\rev{Preservation and General-Purpose Editor Baselines}}{Preservation and General-Purpose Editor Baselines}}
\label{app:rev_preservation}
\rev{This subsection shows that \MethodName{} changes only what it corrects and that general-purpose editors do not solve the task. \textbf{Non-target objects.} We detect and segment every prompt object in the source and corrected images with Grounding DINO~\revcitep{liu2023groundingdino} and SAM~\revcitep{kirillov2023sam}, without any VLM, and measure each object's size change as the change in the square root of its mask area. \revb{On the sources that \MethodName{} edits, objects that receive no size command and do not touch an edited object (1,175 objects in 885 sources) change by a median of 0.05\% (ReScale, on its own edited sources: 0.15\%); only 3 of these objects change by more than 10\%.} Measured by the longest box side, the median change over all non-target objects is 0.00\%. Objects that touch an edited object change in visible area mainly because the resized neighbor covers or reveals part of them.}

\rev{\textbf{Background.} On the 1,496 sources that both methods edit \revb{and whose prompted objects are all segmented in the source image}, we remove the union of all object masks from the source and both outputs, dilated by 24 pixels, so both methods are compared on the same background pixels (75.5\% of the image on average). \MethodName{} alters this background less on every metric and for all seven generators (Table~\ref{tab:app_rev_background}). Half of our background pixels (51.0\%) remain bit-identical to the source, versus 2.3\% for ReScale, because our editor modifies only a local crop and leaves the rest of the image untouched.}

\begin{table}[!htbp]
\captionsetup{font=small,skip=4pt}
\caption{\rev{\textbf{Background preservation.} Shared background of 1,496 sources edited by both methods. Changed pixels: any channel altered by more than 32/255. Differences are \MethodName{} minus ReScale with 95\% prompt-clustered bootstrap CIs.}}
\label{tab:app_rev_background}
\centering\small
\setlength{\tabcolsep}{4pt}
\rev{\resizebox{1\linewidth}{!}{%
\begin{tabular}{@{}lrrr@{}}
\toprule
Metric & \MethodName{} & ReScale & Difference [95\% CI] \\
\midrule
PSNR (dB) $\uparrow$ & \textbf{39.57} & 32.39 & $+7.17\;[+6.64,+7.71]$ \\
SSIM~\revcitep{wang2004ssim} $\uparrow$ & \textbf{0.960} & 0.898 & $+0.062\;[+0.058,+0.066]$ \\
LPIPS~\revcitep{zhang2018lpips} $\downarrow$ & \textbf{0.048} & 0.079 & $-0.031\;[-0.035,-0.027]$ \\
Changed pixels (\%) $\downarrow$ & \textbf{0.65} & 1.42 & $-0.77\;[-1.07,-0.52]$ \\
\bottomrule
\end{tabular}
}}
\end{table}

\rev{\textbf{General-purpose editors.} We give GPT-Image-2~\revcitep{openai2026gptimage2} and the base Qwen-Image-Edit-2511~\revcitep{qwen2025edit2511} the object list with knowledge-base dimensions and ask them to make the relative sizes realistic while keeping everything else unchanged, without telling them which object is wrong. We also write our planner's commands as text instructions for GPT-Image-2. All arms use the same 140 sources (20 per generator)\revb{; each editor is scored together with our outputs in one judging batch} (one vote per relation). Our complete system attains the lowest error (Table~\ref{tab:app_rev_editors}): the base Qwen editor leaves the error unchanged (0.880 vs.\ 0.881 for the source), and GPT-Image-2 redraws the whole image, changing pixels about five times as much as our editor (mean absolute difference 18.8 vs.\ 3.9 levels). Both text-prompted editors have higher error than \MethodName{}, with 95\% prompt-clustered bootstrap CIs of the difference excluding zero ($[+0.060,+0.364]$ for GPT-Image-2 and $[+0.157,+0.440]$ for Qwen).}

\begin{table}[!htbp]
\captionsetup{font=small,skip=4pt}
\caption{\rev{\textbf{General-purpose editors on 140 GenScale sources.} Mean $|s-3|$ ($\downarrow$); \revb{the two GPT-Image-2 rows share one judging batch and the Qwen row uses another, each with its own source and \MethodName{} reference}. Difference: editor minus \MethodName{}.}}
\label{tab:app_rev_editors}
\centering\small
\setlength{\tabcolsep}{3pt}
\resizebox{1\linewidth}{!}{%
\rev{\begin{tabular}{@{}lrrrr@{}}
\toprule
Editor & Source & Editor & \MethodName{} & Difference \\
\midrule
GPT-Image-2, text & 0.892 & 0.781 & \textbf{0.567} & $+0.214$ \\
GPT-Image-2, our plan as text & 0.892 & 0.664 & \textbf{0.567} & $+0.098$ \\
Qwen-Image-Edit-2511, text & 0.881 & 0.880 & \textbf{0.583} & $+0.296$ \\
\bottomrule
\end{tabular}}
}
\end{table}

\subsection{\texorpdfstring{\rev{RoPEMover as the Pipeline Editor}}{RoPEMover as the Pipeline Editor}}
\label{app:rev_ropemover}
\rev{This subsection shows that our editor, not the planner alone, is what turns plans into correct sizes. We keep our observer and planner fixed and replace only the editor with the official RoPEMover~\revcitep{oztas2026ropemover} code and weights, using its default settings, on 197 GenScale development sources \revb{(a subset of the 230 tuning sources of Appendix~\ref{app:rev_heldout})} with 316 planned actions. RoPEMover receives the planned source mask, displacement, apparent scale, and depth change; in the main configuration it also receives the same enlarged editing crop as our editor. An action counts as realized when the object is found at the planned location and its measured apparent scale (square root of the mask-area ratio, from Grounding DINO and SAM) lies within 20\% of the planned scale; actions that cannot be measured count as not realized. Our editor realizes 77\% of the planned size changes and follows the planned scale with slope 0.92 (Table~\ref{tab:app_rev_ropemover}); under the GenScale judge, the same plans executed by our editor lower the error from 0.921 to 0.505, versus 0.820 with RoPEMover. On the ObjectMover requests of Table~\ref{tab:exp_actuator}, RoPEMover uses its released drag parameters. Given edit parameters identical to ours, RoPEMover reaches 0.866 target DINO. The advantage is not a sampling artifact: across three RoPEMover seeds and three seeds of our editor, all nine seed pairings favor \MethodName{} on these 188 requests (target DINO $+0.026$ to $+0.038$; mean $+0.032$, 95\% CI [$+0.023$, $+0.042$]), and on the 150 held-out size--depth renders of Table~\ref{tab:app_interface_rendered} our editor reaches 0.921 target-box DINO versus 0.768 for RoPEMover.}

\begin{table}[!htbp]
\captionsetup{font=small,skip=4pt}
\caption{\rev{\textbf{Executing \MethodName{} plans with different editors.} 316 planned actions on 197 sources. Realized: within 20\% of the planned apparent scale at the planned location (95\% source-bootstrap CI). Slope: measured versus planned scale (1 = as planned, 0 = unchanged). Error: GenScale $|s-3|$ before restoration (source 0.921).}}
\label{tab:app_rev_ropemover}
\centering\small
\setlength{\tabcolsep}{3pt}
\resizebox{0.9\linewidth}{!}{%
\rev{\begin{tabular}{@{}lrrr@{}}
\toprule
Editor & Realized (\%) $\uparrow$ & Slope & Error $\downarrow$ \\
\midrule
RoPEMover, full image & 7.3 $[4.3,10.5]$ & 0.07 & 0.909 \\
RoPEMover, our crop & 16.8 $[12.7,21.1]$ & 0.19 & 0.820 \\
\MethodName{} & \textbf{77.2} $[72.6,81.7]$ & \textbf{0.92} & \textbf{0.505} \\
\bottomrule
\end{tabular}}
}
\end{table}

\subsection{\texorpdfstring{\rev{Matched Retraining}}{Matched Retraining}}
\label{app:rev_retrain}
\rev{This subsection shows that accurate resizing comes from the size--depth training commands and that 3D supervision outperforms 2.5D under matched training. To separate the training-command distribution from data quantity and composition, we retrain the editor with one recipe (rank-16 LoRA, learning rate $10^{-4}$, $768\times768$, 7,830 sample exposures) and the same trainer as the positional-control comparison (Appendix~\ref{app:exp_interface}). Each comparison in Table~\ref{tab:app_rev_retrain} changes one factor: data quantity (full set vs.\ size--depth subset), command distribution (subset vs.\ translation-only), data composition (subset vs.\ small 3D set), and supervision (small 3D vs.\ small 2.5D set). The translation-only set re-renders the same 783 scenes with identical assets, layouts, cameras, and source frames, replacing every size--depth command with a lateral translation. Its real-photo relocation fidelity is unchanged, but held-out target DINO drops from 0.909 to 0.826 and silhouette IoU from 0.771 to 0.629, so accurate resizing comes from the size--depth commands. This model is the matched counterpart of a relocation-only RoPE editor: backbone, interface, scenes, and budget are identical, and only the size--depth commands are removed. With matched trainers and data, 3D supervision also exceeds 2.5D on both real photographs and held-out renders.}

\begin{table}[!htbp]
\captionsetup{font=small,skip=4pt}
\caption{\rev{\textbf{Matched retraining.} ObjectMover: target DINO on 188 real requests. Held-out: target-box DINO and silhouette IoU on 150 size--depth renders. All models share the trainer, recipe, and evaluation environment ($\uparrow$ for all columns).}}
\label{tab:app_rev_retrain}
\centering\small
\setlength{\tabcolsep}{3pt}
\resizebox{1\linewidth}{!}{%
\rev{\begin{tabular}{@{}lrrrr@{}}
\toprule
Training set & Pairs & ObjectMover & Held-out DINO & Held-out IoU \\
\midrule
Full set & 2,468 & \textbf{0.895} & \textbf{0.923} & \textbf{0.808} \\
Size--depth subset & 783 & 0.893 & 0.909 & 0.771 \\
Translation-only & 783 & 0.890 & 0.826 & 0.629 \\
Small 3D set & 783 & 0.879 & 0.902 & 0.744 \\
Small 2.5D set & 783 & 0.864 & 0.871 & 0.730 \\
\bottomrule
\end{tabular}}
}
\end{table}

\subsection{\texorpdfstring{\rev{Reordering Planner Priorities for Spatial Relations}}{Reordering Planner Priorities for Spatial Relations}}
\label{app:rev_relation}
\rev{\MethodName{} largely preserves scene layout by default: on the 2,800 GenScale sources, 65.5\% of edited objects are resized in place, and on the output images of edited sources 97.8\% of object pairs keep their left--right order and 94.4\% their front--back order (pairs at least 8 pixels apart at $512\times512$; centroid and ground-contact order). Because the planner's priorities are explicit rules, spatial-relation preservation can be promoted without retraining and without changing any size command. On 197 GenScale development sources (176 edited, 316 actions), we count a left--right flip when the centroid order of an object pair changes and a front--back flip when their ground-contact order changes, for pairs initially at least 8 pixels apart at $512\times512$. Placing relation preservation directly after physical validity reduces flips from 10 to 1 of 504 left--right pairs and from 18 to 1 of 322 front--back pairs, with no physical violations (Table~\ref{tab:app_rev_relation}); both remaining flips come from the single action that cannot satisfy the physical constraints otherwise. Only 20 of the 176 edited sources receive a different plan.}

\begin{table}[!htbp]
\captionsetup{font=small,skip=4pt}
\caption{\rev{\textbf{Planner priority orders.} Order flips among object pairs; Violations and Physical count actions (of 316) that violate any planner constraint and a physical constraint, respectively. Size commands are identical in all rows.}}
\label{tab:app_rev_relation}
\centering\small
\setlength{\tabcolsep}{3pt}
\resizebox{1\linewidth}{!}{%
\rev{\begin{tabular}{@{}lrrrr@{}}
\toprule
Priority order & L--R flips & F--B flips & Violations & Physical \\
\midrule
In place only & 0/504 & 0/322 & 96 & 54 \\
Default & 10/504 & 18/322 & 5 & 0 \\
Physics, then relations & 1/504 & 1/322 & 14 & 0 \\
Relations first & 0/504 & 0/322 & 15 & 1 \\
\bottomrule
\end{tabular}}
}
\end{table}

\subsection{\texorpdfstring{\rev{In-Place Resizing versus Full Placement}}{In-Place Resizing versus Full Placement}}
\label{app:rev_inplace}
\rev{Full placement planning matters where resizing in place would break the scene. On the same 197 GenScale development sources (316 edit actions), restricting the planner to in-place resizing leaves 96 actions that violate its geometric constraints (support, collision, overlap, framing, target-token support, or editor range); full planning resolves 91 of them by changing the object's depth or lateral position (Table~\ref{tab:app_rev_relation}, first two rows). We render the full and in-place plans on the 21 sources whose plans change under either relation-preserving order in Table~\ref{tab:app_rev_relation} and score them with the Gemini judge (five votes per relation). On these sources, the mean error falls from 1.214 for the source images to 0.627 with full planning, compared with 0.730 with in-place resizing.}

\subsection{\texorpdfstring{\rev{Robustness to the Size Prior}}{Robustness to the Size Prior}}
\label{app:rev_sizeprior}
\rev{This subsection shows that \MethodName{} does not hinge on a hand-built size table and initiates edits only for deviations beyond a tolerance.}

\rev{\textbf{The size table can come from a language model.} We give only the 100 category names to Claude Sonnet~5.5 and GPT-6.1 Terra and ask for typical dimensions, without any numbers or images. For 80 (Claude) and 85 (GPT) of the 100 categories, the longest dimension lies within 10\% of \rev{the knowledge base}, and the category rankings agree almost perfectly (Spearman 0.98). Re-running the planner on \rev{the 2,119 non-development GenScale sources that pass instance binding and observation, with the language-model values in place of the knowledge base,} reverses the correction direction (enlarge vs.\ shrink) for only 0.9\% (Claude) and 0.5\% (GPT) of the edited objects. Randomly perturbing every category length of \rev{the knowledge base} by up to $\pm10\%$ or $\pm20\%$ reverses the direction for 0.3\% and 0.5\% of edited objects. Because the table is an explicit input, users can also replace or override any entry directly.}

\rev{\textbf{Edits target clear deviations.} The planner edits an object only when its observed size deviates from the scene's shared scale by more than a factor of 1.30; objects within this tolerance are edited in only 2.0\% of cases, and objects beyond it are left unedited in only 4.2\%. Table~\ref{tab:app_rev_kb_gate} groups the GenScale relations \rev{whose two objects are both measured by the observer} by the observed size ratio relative to the knowledge-base ratio; 1.2$\times$ corresponds to the judge's tolerance for a plausible score. Relations that receive no command have a median deviation of $1.10\times$. \revb{Outside the judge's plausible band (ratio between 0.8 and 1.2)}, 97.7\% of the 4,579 relations receive a size command, and the commanded change points in the required direction in 99.4\% of cases\revb{; the symmetric bins of Table~\ref{tab:app_rev_kb_gate} place 152 further relations (ratio between 0.80 and 0.83) in the $1.2$--$1.5\times$ row}. Commands on relations within $1.2\times$ arise because the objects share one scene scale: an object resized to fix one relation also changes its other relations.}

\begin{table}[!htbp]
\captionsetup{font=small,skip=4pt}
\caption{\rev{\textbf{Edit rate by observed deviation.} \rev{The 6,166 GenScale relations whose two objects are both measured by the observer}, grouped by the observed size ratio relative to the knowledge-base ratio (the larger of the ratio and its inverse). Edited: at least one object receives a size command.}}
\label{tab:app_rev_kb_gate}
\centering\small
\setlength{\tabcolsep}{4pt}
\rev{\resizebox{0.68\linewidth}{!}{%
\begin{tabular}{@{}lrr@{}}
\toprule
Observed deviation & Relations & Edited (\%) \\
\midrule
$\le 1.2\times$ & 1,435 & 39.7 \\
$1.2$--$1.5\times$ & 1,527 & 95.7 \\
$1.5$--$2\times$ & 1,341 & 98.1 \\
$2$--$3\times$ & 1,110 & 98.7 \\
$> 3\times$ & 753 & 97.9 \\
\bottomrule
\end{tabular}
}}
\end{table}

\subsection{\texorpdfstring{\rev{Apparent-Size Baseline}}{Apparent-Size Baseline}}
\label{app:rev_2d}
\rev{We compare our observer with a reading that uses only 2D apparent size (the change in the target's mask size) on controlled Blender renders with known interventions: 12 object pairs seen from four viewpoints, in which the target's physical size, its depth, or both change by a factor of $e^{0.4}\approx1.49$ while the reference object is unchanged. \revb{The design has 48 pair--viewpoint groups; after removing configurations that leave the frame, 57 depth-only cases and 55 joint size--depth cases remain.} A method reports a size change when its estimated relative size change exceeds a factor of 1.2. When the target is moved along the ground with its physical size unchanged (57 cases), the 2D reading reports a size change in all 57 cases and our observer in 7. When size and distance change together so that the apparent size is preserved within 5\% (46 cases), our observer recovers the physical size change in 39 cases and the 2D reading in none (Table~\ref{tab:app_rev_2d}).}

\begin{table}[!htbp]
\captionsetup{font=small,skip=4pt}
\caption{\rev{\textbf{Separating size from distance on controlled renders.} Depth only: the target moves along the ground without resizing, so any reported size change is a false alarm (our observer abstains on one of these cases). Size and depth: size and depth change together with apparent size preserved within 5\%.}}
\label{tab:app_rev_2d}
\centering\small
\setlength{\tabcolsep}{4pt}
\rev{\resizebox{1\linewidth}{!}{%
\begin{tabular}{@{}lrrr@{}}
\toprule
Intervention & Cases & 2D reading & Our observer \\
\midrule
Depth only: false size alarms $\downarrow$ & 57 & 57 & \textbf{7} \\
Size and depth: change recovered $\uparrow$ & 46 & 0 & \textbf{39} \\
\bottomrule
\end{tabular}
}}
\end{table}

\subsection{\texorpdfstring{\rev{Sensitivity to the Assumed Camera}}{Sensitivity to the Assumed Camera}}
\label{app:rev_camera}
\rev{The observer and planner assume one camera for every image: a normalized focal length $f/W=2.474$ (a $23^\circ$ horizontal field of view) and a fixed camera height, with the tilt set by the estimated horizon (Appendix~\ref{app:observer}). We measure how much this assumption matters by re-running both stages on the 2,119 non-development GenScale sources that pass instance binding and observation, changing the assumed camera consistently in the observer and the planner while keeping the masks, horizon estimates, and poses fixed. With the deployed camera, the replay reproduces the stored readings and commands of all 2,119 sources.}

\rev{\textbf{Camera height.} Halving or doubling the assumed camera height shifts the log-size residuals of all objects in an image by the same constant. The relative readings between objects therefore change by less than $10^{-4}$, every pairwise verdict is unchanged, and no correction direction reverses; the only effect is that some objects leave the scale interval searched by the solver (130 and 11 sources then abstain).}

\rev{\textbf{Field of view.} Table~\ref{tab:app_rev_camera} varies the assumed horizontal field of view from $15^\circ$ to $60^\circ$. A pairwise verdict is keep, enlarge, or shrink at the judge's tolerance of $1.2\times$. Across this range, the correction direction (enlarge vs.\ shrink) reverses for at most 0.64\% of the objects edited under both cameras, comparable to substituting the size table (Appendix~\ref{app:rev_sizeprior}), and 89--98\% of object pairs keep their verdict. Pairs that involve flat objects are the most sensitive. At wider assumed fields of view, more objects fall outside the solver's scale interval, so the observer abstains on more sources.}

\begin{table}[!htbp]
\captionsetup{font=small,skip=4pt}
\caption{\rev{\textbf{Sensitivity to the assumed field of view.} 2,119 GenScale sources; all rows are compared with the deployed camera ($23^\circ$). Reading change: median absolute change of the relative log-size reading between two objects in an image. Beyond tolerance: object pairs whose relative reading changes by more than $1.2\times$. Reversed: edited objects whose correction direction flips.}}
\label{tab:app_rev_camera}
\centering\small
\setlength{\tabcolsep}{2pt}
\resizebox{1\linewidth}{!}{%
\rev{\begin{tabular}{@{}rrrrrrr@{}}
\toprule
\makecell[r]{Assumed\\FOV} & $f/W$ & \makecell[r]{Reading\\change} & \makecell[r]{Beyond\\tolerance\\(\%)} & \makecell[r]{Verdict\\unchanged\\(\%)} & \makecell[r]{Reversed\\(\%)} & \makecell[r]{Sources\\abstained} \\
\midrule
$15^\circ$ & 3.798 & 0.014 & 0.2 & 97.7 & 0.06 & 3 \\
$30^\circ$ & 1.866 & 0.013 & 0.1 & 97.4 & 0.10 & 39 \\
$45^\circ$ & 1.207 & 0.042 & 2.4 & 92.8 & 0.25 & 165 \\
$60^\circ$ & 0.866 & 0.071 & 10.6 & 88.6 & 0.64 & 337 \\
\bottomrule
\end{tabular}}
}
\end{table}

\subsection{\texorpdfstring{\revb{Editor Comparison on Sources Outside Development}}{Editor Comparison on Sources Outside Development}}
\label{app:rev_heldout_swap}
\revb{This subsection repeats the editor comparison of Table~\ref{tab:exp_swap} on sources that appear in none of our development or pilot experiments. We select 196 such GenScale sources (28 per generator, by a fixed hash order; 603 relations) and execute ReScale's plans with our editor. All images are scored in one batch by the Gemini judge and by the independent GPT judge, with one vote per relation. Replacing ReScale's editor with ours lowers the error under both judges (Table~\ref{tab:app_rev_heldout_swap}). Under the independent GPT judge, the complete \MethodName{} system also has a lower error than ReScale on these sources (0.637 vs.\ 0.699).}

\begin{table}[!htbp]
\captionsetup{font=small,skip=4pt}
\caption{\revb{\textbf{Editor comparison on 196 sources outside development.} Mean $|s-3|$ per source ($\downarrow$). Both edited columns use ReScale's front end. Difference: ReScale's editor minus ours, with 95\% prompt-clustered bootstrap CIs.}}
\label{tab:app_rev_heldout_swap}
\centering\small
\setlength{\tabcolsep}{2pt}
\resizebox{1\linewidth}{!}{%
\revb{\begin{tabular}{@{}lrrrr@{}}
\toprule
Judge & Source & \makecell[r]{ReScale\\editor} & \makecell[r]{Our\\editor} & Difference [95\% CI] \\
\midrule
GPT-6.1 Terra & 1.026 & 0.699 & \textbf{0.626} & $+0.073\;[+0.010,+0.143]$ \\
Gemini 3.1 Pro & 0.869 & 0.554 & \textbf{0.458} & $+0.096\;[+0.019,+0.178]$ \\
\bottomrule
\end{tabular}}
}
\end{table}

\end{document}